\documentclass[11pt]{article}

\usepackage{latexsym,mathrsfs}
\usepackage{amsmath,amssymb} 
\usepackage{amsthm,enumerate,verbatim}
\usepackage{amsfonts}
\usepackage{graphicx}
\usepackage{algorithm}
\usepackage{algorithmic}
\usepackage{url}
\usepackage[dvipsnames]{xcolor}
\usepackage{pifont}
\usepackage{mathtools}
\usepackage{makecell}
\usepackage{rotating}
\usepackage{tabularx}
\usepackage{tablefootnote}
\usepackage{multirow}
\usepackage{lscape} 
\usepackage[normalem]{ulem}
\usepackage{hhline}

\let\oldFootnote\footnote
\newcommand\nextToken\relax

\renewcommand\footnote[1]{%
	\oldFootnote{#1}\futurelet\nextToken\isFootnote}

\newcommand\isFootnote{%
	\ifx\footnote\nextToken\textsuperscript{,}\fi}

\newtheorem{definition}{Definition}
\newtheorem*{definition*}{Definition}

\newtheorem{remark}{Remark}

\DeclareMathOperator{\logdet}{logdet} 
\DeclareMathOperator{\tr}{tr} 
\DeclareMathOperator{\diag}{diag} 
\DeclareMathOperator{\dist}{dist}

\title{Beyond Linear Subspace Clustering: A Comparative Study of Nonlinear Manifold Clustering Algorithms}

\date{}

\author{Maryam Abdolali$^{*}$ \qquad Nicolas Gillis\thanks{Emails: \{maryam.abdolali, nicolas.gillis\}@umons.ac.be. The authors acknowledge the support by the European Research Council (ERC starting grant no 679515), and by the Fonds de la Recherche Scientifique - FNRS and the Fonds Wetenschappelijk Onderzoek - Vlaanderen (FWO) under EOS Project no O005318F-RG47.} \\   
	Department of Mathematics and Operational Research \\ 
	Facult\'e Polytechnique, Universit\'e de Mons \\ 
	Rue de Houdain 9, 7000 Mons, Belgium
}

\definecolor{brightpink}{rgb}{1.0, 0.0, 0.5}
	
	\newcommand{\ngi}[1]{{{\color{brightpink} #1}}}

\begin{document}

\maketitle

	\begin{abstract}
	Subspace clustering is an important unsupervised clustering approach. It is based on the assumption that the high-dimensional data points are approximately distributed around several low-dimensional linear subspaces. The majority of the prominent subspace clustering algorithms rely on the representation of the data points as linear combinations of other data points, which is known as a self-expressive representation.  
	To overcome the restrictive linearity assumption, 
	numerous nonlinear approaches were proposed to extend successful subspace clustering approaches to data on a union of nonlinear manifolds. 
	In this comparative study, we provide a comprehensive overview of nonlinear subspace clustering approaches
	proposed in the last decade. We introduce
	a new taxonomy to classify the state-of-the-art approaches into three categories, namely 
	locality preserving, 
	kernel based, and 
	neural network based. 
	The major representative algorithms within each category are extensively compared on carefully designed synthetic and real-world data sets. 
	The detailed analysis of these approaches
	unfolds potential research directions and unsolved challenges in this field.
\end{abstract}


\section{Introduction} \label{intro}
Understanding and processing high-dimensional data is a key component of numerous applications in many domains including machine learning, signal processing and computer vision.  However, analyzing high-dimensional data using classical data mining algorithms is not only challenging due to computational costs but it can also easily lead to the well-known problem of the \emph{curse of dimensionality}~\cite{bellman1966dynamic}. Luckily, in most of applications, the data often have fewer degrees of freedom than the ambient dimension and they can be approximately represented by a small number of features; see~\cite{van2009dimensional, udell2019big}. 
This led to the development of a vast variety of algorithms for the long-standing problem of extracting latent structures from high-dimensional data; see for example~\cite{vidal2016principal} and references therein.  

The main focus of the majority of these algorithms is to fit a \emph{single} low-dimensional linear subspace to the data, with principal component analysis (PCA) being the most well-known pioneer algorithm in this area~\cite{jolliffe1986principal}. 
However, the data often belongs to multiple categories with different intrinsic structures, and modeling the data using only \emph{one} low-dimensional subspace might be too restrictive~\cite{vidal2005generalized}. 
In fact, in many applications, the data is better represented by \emph{multiple} subspaces. Representing the data using a union of multiple subspaces gave rise to \emph{linear subspace clustering}~\cite{costeira1995multi, bradley2000k}; see \cite{vidal2011subspace} for a survey paper. 

\begin{definition}[Linear subspace clustering (linear SC)]\label{sc_def}
	Let $X \in \mathbb{R}^{d \times n}$ be the input high-dimensional data, with $d$-dimensional data points as its columns. Suppose the data points are distributed around $c$ unknown linear subspaces $S_1, S_2,\dots, S_c$ with intrinsic dimensions $d_1, d_2,\dots, d_c$, respectively, such that $d_i \ll d$ for all $i \in \{1,\dots,c\}$. The problem of subspace clustering is defined as segmenting the data points based on their corresponding subspaces and estimating the parameters of each subspace.
\end{definition}
In the literature, linear SC is usually referred to as SC but, to avoid any confusion, we refer to it as linear SC in this survey. Linear SC is a clustering framework where the data points are grouped together based on the underlying subspaces they belong to. In other words, the similarity between the data points is measured by how well they fit within a low-dimensional linear subspace. 
Modeling data with multiple subspaces has many applications in image and signal processing. In fact, data points are often collected from multiple classes/categories, and extracting latent low-dimensional structures within each class/category results in more meaningful and more compressed representations. 
For example, given a collection of facial images taken with variations in the lightening, cluserting these images according to the person they represent can be modeled as a linear SC problem. In fact, under the Lambertian surface assumption, facial images of a single individual under different illumination conditions lie on a roughly nine-dimensional subspace which has far less effective dimensions than the original ambient space (usually an image consists of thousands of pixels)~\cite{basri2003lambertian}. Hence, facial images of various people under different lightening conditions can be approximated by multiple low-dimensional subspaces. Other notable examples of data that can be approximated by multiple low-dimensional subspaces include segmenting trajectories of moving objects in a video~\cite{tomasi1992shape}, clustering images of hand-written digits~\cite{hastie1997metrics}, and partitioning sequences of video frames into semantic parts~\cite{vidal2016principal}. 

Driven by this wide range of applications, many efficient algorithms for linear SC have been developed. 
These algorithms are divided into four categories~\cite{vidal2011subspace,elhamifar2013sparse}: 
(i)~iterative, 
(ii)~statistical, 
(iii)~algebraic, and 
(iv)~spectral clustering based approaches. 
Let us briefly describe these four categories:  
\begin{itemize}
	
	\item[(i)] Iterative approaches were among the first ones for linear SC~\cite{bradley2000k,ho2003clustering,tseng2000nearest}. Inspired by iterative refinement in centroid-based clustering, such as k-means, two steps are carried out alternatively: assigning each data point to the closest  subspace, and updating the parameters of each subspace based on the points assigned to it (e.g., using the truncated singular value decomposition). 
	However, iterative approaches are sensitive to initialization and parameters such as the dimensions and the number of subspaces (these are usually not available in most applications). 
	
	\item[(ii)] Statistical approaches~\cite{tipping1999mixtures,gruber2004multibody}, which suffer from the same drawbacks as iterative ones, usually assume that the distribution of the data within each subspace is Gaussian, and hence the linear SC problem is mapped into the iterative problem of fitting a mixture of Gaussian distribution to the data using expectation maximization. 
	
	\item[(iii)] Some approaches use algebraic matrix factorization to perform the data segmentation~\cite{costeira1998multibody,kanatani2001motion}. 
	These approaches are not only sensitive to noise but are also based on the assumption that the underlying subspaces are independent. In related algebraic-geometric approaches~\cite{vidal2005generalized,tsakiris2017algebraic}, the association of the data points to the subspaces are revealed by fitting a polynomial to the data. These approaches are computationally expensive and sensitive to noise.

	\item[(iv)] Spectral clustering based algorithms are the most popular and successful linear SC approaches. They have received significant attention over the last decade. These approaches are based on recent advances in sparse and low-rank representation~\cite{huang2007sparse,wright2008robust,zhang2015survey}. They first learn a directed weighted graph representing the interaction between the data points,
	and then apply  spectral clustering to segment this graph; see Section~\ref{sc} for more details. 
\end{itemize}

Even though linear SC approaches have achieved impressive performances in several applications, 
the global \emph{linearity} assumption is somewhat strong. Assuming that the data points lie close to multiple \emph{linear} subspaces is limiting and might be violated in some applications. In order to extend the applicability of linear SC in real-world problems, there has been numerous attempts to generalize linear SC for data lying on the union of nonlinear subspaces or manifolds.  This is the main topic of this survey paper.

\paragraph{Our contributions} 

The main contributions of this paper are summarized as follows:
\begin{itemize}
	
	\item We propose a new taxonomy to classify nonlinear SC algorithms,  divided  them into three categories: 
	(1)~locality preserving, 
	(2)~kernel based, and 
	(3)~neural network based 
	
	\item We provide a comprehensive overview of these nonlinear SC algorithms. 
	
	\item We numerically analyze and compare the representative algorithms in each category. This survey can be considered as a guiding tutorial for understanding and developing nonlinear SC algorithms. 
	
\end{itemize}

\paragraph{Outline of the survey}

The rest of the paper is organized as follows. 
Since spectral clustering based linear SC algorithms are at the core of the majority of existing nonlinear SC methods, we briefly introduce and review the most important ones in Section~\ref{sc}. 
Section~\ref{nsc} provides a detailed overview of nonlinear SC methods, based on our new taxonomy that divides them into three categories.  
Section~\ref{sec:compcost} discusses the computational cost of these methods, while Section~\ref{eval} provides a numerical comparison of the main representative methods on synthetic and real-world data sets. 
In Section~\ref{future}, we discuss the existing challenges in dealing with nonlinear structures in data, and future research directions. 
We conclude the paper in Section~\ref{conc}. 

\section{Linear subspace clustering based on spectral clustering} \label{sc} 

\begin{figure}[ht!]
	\begin{center}
		\includegraphics[width=1\textwidth]{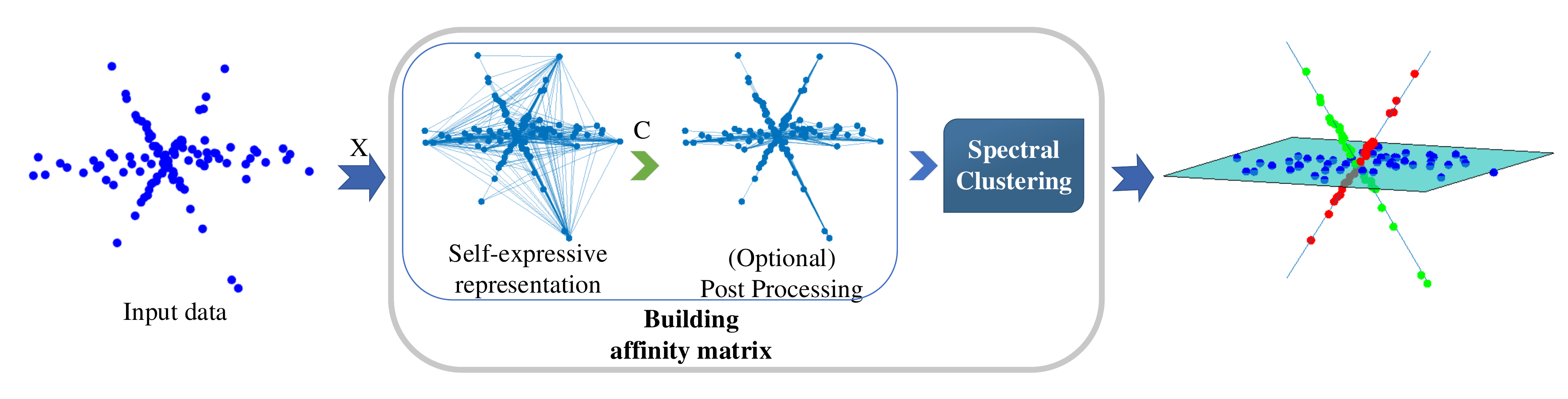}   
		\caption{The overall procedure of linear SC approaches based on spectral clustering. In the first step of these approaches, an affinity graph representing the input data is constructed, which is sometimes post-processed to improve the quality of the representation (e.g., using sparsification).   
			In the second step, a spectral clustering algorithm is applied to  obtain the clusters.} 
		\label{overal}
	\end{center}
\end{figure}

Among the wide variety of linear SC approaches, the current state-of-the-art algorithms are based on spectral clustering. 
As illustrated in Figure~\ref{overal}, these algorithms follow a two-step strategy: 
\begin{itemize} 
	
	\item In the first step, an affinity matrix is constructed: each vertex in the corresponding graph corresponds to a  data point while edges connect similar data points.

	\item In the second step, spectral clustering is applied on the affinity matrix to segment the data points. 
	
\end{itemize}  

In the next two sections, we describe these two steps in more details. 

\subsection{Building the affinity matrix} 

The first and undoubtedly main step of linear SC algorithms based on spectral clustering is to build an affinity matrix. The affinity matrix should reveal the \emph{pairwise} similarities  between the data points, that is, the data points from the same cluster/subspace should be highly connected with high similarities, as opposed to points from different subspaces. 
The construction of this affinity matrix is typically
carried out using a self-expressive representation; this is discussed in Section~\ref{sec:selfexp}. 
It is sometimes followed by a post-processing step to polish the obtained pairwise similarities;  this is discussed in Section~\ref{sec:post_proc}. 

\begin{remark} 
	Building the affinity matrix is not  only useful for  clustering. 
	It can be used in other contexts where it is useful to  model the input data as a graph by capturing pairwise relationships between the data points~\cite{cheng2009learning}. 
	In other words, this first step of spectral based SC approaches is also used in other  applications that model and analyze the structure within the data~\cite{qiao2018data}.
\end{remark} 

\subsubsection{Self-expressive representation} \label{sec:selfexp}

Constructing a pairwise affinity matrix that reflects the multiple subspace structure of the data points is the main challenge in the linear SC algorithms based on spectral clustering. The affinity matrix is often constructed by exploiting the \emph{self-expressiveness} property of the data points. This property, which is known as \emph{collaborative representation} in the sparse representation literature~\cite{zhang2011sparse}, is based on the fact that each data point can be represented as a linear combination of other data points in the same subspace. Mathematically,  for all $i\in \{1,\dots,n\}$, 
\begin{align} \label{selfexpress}
	X(:,i) = XC(:,i) \quad \text{with} \quad C(i,i)=0, 
\end{align}
where $C\in \mathbb{R}^{n \times n}$ is the coefficient matrix, $C(i,i)$ is the $i$-th diagonal entry of $C$, 
and $X(:,i)$ is the $i$-th column of $X$. 
The condition $C(i,i)=0 \ \text{for all } i\in\{1,\dots.,n\}$ eliminates the trivial solution of expressing each point by itself. 
However, there are typically infinitely many solutions to~\eqref{selfexpress} and many of them might not be \emph{subspace preserving}, that is, all data point might not be expressed using a linear combination of data points belonging only to the \emph{same} subspace. Let us formally define this concept. 

\begin{definition}[Subspace preserving representation]\label{sp_def}
	Given the data matrix $X \in \mathbb{R}^{d \times n}$ drawn from the union of $c$ subspaces $S_1, S_2,\dots, S_c$, let $C \in \mathbb{R}^{n \times n}$ be the coefficient matrix corresponding to the output of a linear SC algorithm based on self-expressiveness; see~\eqref{selfexpress}.  The matrix $C$ is subspace preserving if for any data point $X(:,i) \in S_\ell$, $C(i,j)=0$ for all $j$ such that $X(:,j) \not \in S_\ell$. In other words, for all $i \in \{1,\dots,n\}$, the nonzero elements in $C(:,i)$ only correspond to  data points from the same subspace as $X(:,i)$.
\end{definition}

To fulfill the goal of linear SC and reveal the association of each data point to the underlying subspace, the coefficient matrix should be (approximately) subspace preserving. Hence, to ensure that the obtained coefficient matrix is subspace preserving, several regularizations on the coefficient matrix are used in the literature, including sparsity~\cite{elhamifar2013sparse} and low-rankness~\cite{liu2012robust}. In general, the linear SC problem based on self-expressiveness is formulated as follows:
\begin{align} 
	\min_{C \in \mathbb{R}^{n \times n}, E \in \mathbb{R}^{d \times n}} & \; f(C) + \lambda g(E) \label{generalsc} 
	\\  \text{ such that }  & \; X = XC + E \text{ and } 
	\diag(C)=0, \nonumber
\end{align} 
where $f$ is a regularization function for the coefficient matrix (see below), 
$E \in \mathbb{R}^{d \times n}$ models the noise, 
$g$ is a regularization function for modeling the noise 
(for example the $\ell_1$ norm for sparse gross noise, or the Frobenius norm for Gaussian noise), 
and $\diag(C)$ is the vector containing the diagonal entries of $C$. 
The conditions under which the obtained coefficient matrix is subspace preserving strongly depends on the regularization function $f$. 
Table~\ref{coefreg} summarizes the most common regularization functions and the theoretical conditions for the corresponding coefficient matrix to be subspace preserving. 
In this table $||\cdot||_0$, $||\cdot||_1$, $||\cdot||_*$ and $||\cdot||_F^2$ indicate the $\ell_0$ norm (the number of nonzero entries), the component-wise $\ell_1$ norm (the sum of the absolute value of the matrix entries), 
the nuclear norm (the sum of singular values), 
and the squared Frobenius norm (the sum of the squares of the matrix entries), respectively. 

\begin{table}[h!]
	\caption{Major linear SC models based on spectral clustering. They differ in the regularization function $f$ used for the  coefficient matrix $C$ in~\eqref{generalsc}, and guarantee to provide a subspace preserving representation under different conditions. } 
	\label{coefreg}  
	\begin{tabularx}{\textwidth}{|c|c|X|} \hline
		\thead{Method} & \thead{\begin{tabular}{@{}c@{}}Regularization \\ Function $f$ \end{tabular}} & \thead{\thead{\begin{tabular}{@{}c@{}}Available subspace 
					\\preserving guarantees \end{tabular}}} \\
		\hline
		\begin{tabular}{@{}c@{}}Sparse Subspace Clustering \\ (SSC)~\cite{elhamifar2013sparse}\end{tabular} 
		& $||C||_1$ & independent, disjoint and intersecting subspaces in noiseless and noisy cases \\ \hline
		\begin{tabular}{@{}c@{}}$\ell_0$-induced Sparse Subspace \\ Clustering ($\ell_0$-SSC)~\cite{yang2016ell}\end{tabular} 
		& $||C||_0$ & all arrangements of noiseless distinct subspaces\\
		\hline
		\begin{tabular}{@{}c@{}}Low Rank Representation \\ (LRR)~\cite{liu2012robust} \end{tabular}  
		& $||C||_*$ & noiseless independent   subspaces\\ \hline
		\begin{tabular}{@{}c@{}}Least Square Regression \\ (LSR)~\cite{lu2012robust} \end{tabular} 
		& $||C||_F^2$ & noiseless independent subspaces \\ \hline
		\begin{tabular}{@{}c@{}}Multi-Subspace Representation \\ (MSR)~\cite{luo2011multi} \end{tabular}
		& $||C||_1 + \lambda ||C||_*$ & Not available \\ \hline
		\begin{tabular}{@{}c@{}}Subspace Segmentation via QP \\ (SSQP)~\cite{wang2011efficient} \end{tabular}
		& $||C^\top C||_1$ & noiseless orthogonal subspaces\\ \hline
		\begin{tabular}{@{}c@{}}Correlation Adaptive Subspace  \\ Segmentation (CASS)~\cite{lu2013correlation} \end{tabular}
		& $\sum_j ||X \diag\big(C(:,j)\big)||_*$ & noiseless independent subspaces \\ \hline
		\begin{tabular}{@{}c@{}}Elastic Net Subspace  \\ Clustering (ENSC)~\cite{you2016oracle} \end{tabular}
		& $||C||_1 + \lambda||C||_F^2$ & independent and disjoint subspaces for the noiseless case with strong dependency on the value of the parameter $\lambda$\\
		\hline
	\end{tabularx} 
\end{table}

\paragraph{Effect of regularization on subspace preserving representations}

The difficulty of linear SC depends on several factors, these include 
the arrangement of the subspaces, 
the separation between subspaces, 
the distribution of the data points within each subspace, 
the number of points per subspace, and 
the noise level. 
There are three main arrangements of subspaces which play a key role in identifying the subspace recovery conditions: independent, disjoint, and intersecting (or overlapping) subspaces. These arrangements are defined as follows: 
\begin{definition}[Independent subspaces] A collection of $c$ subspaces $S_1,\dots,S_c$ are said to be independent if 
	$\dim(\oplus_{i=1}^c S_i) = \sum_{i=1}^c \ \dim(S_i)$, 
	where $\oplus$ denotes the direct sum between subspaces, 
	and $\dim(S)$ is the dimension of $S$. 
\end{definition}
\begin{definition}[Disjoint subspaces] A collection of $c$ subspaces are disjoint if 
	$\dim(S_i \oplus S_j)=\dim(S_i)+\dim(S_j)$ and $S_i \cap S_j = \{0\}$ for all $i \neq j$. 
\end{definition}
\begin{definition}[Intersecting subspaces] A collection of $c$ subspaces are intersecting/overlapping if  $1 \leq \dim(S_i \cap S_j) < \min \{\dim(S_i),\dim(S_j)\}$ 
	for some $i \neq j$. 
\end{definition}

Independent subspaces (orthogonal subspaces being a special case) are the easiest to separate, and hence most  regularizations are guaranteed to be subspace preserving in this case, at least in noiseless conditions; see Table~\ref{coefreg}.    
An example is two distinct lines in a plane.    

Clustering data from disjoint subspaces is a more challenging  scenario.  
An example are three distinct lines in a plane.  
Sparsity regularization based on the $\ell_0$ and $\ell_1$ norms  is the only one proven to be subspace preserving, under some conditions; see Table~\ref{coefreg}. 
These conditions depend on the separation between the subspaces and the distribution of the data points  within each subspace. With sufficiently separated subspaces and well-spread data points (not skewed towards a specific direction), sparsity regularization is guaranteed to provide subspace preserving coefficients in noiseless and noisy cases. 
For detailed theoretical discussions on this topic, we refer the interested reader 
to~\cite{elhamifar2013sparse,wang2016noisy}. 

Intersecting/overlapping subspaces is the most general subspace arrangement for which there is no particular assumption on the subspaces, and any two subspaces can have a nontrivial 
intersection~\cite{wang2016noisy,soltanolkotabi2012geometric}. 
Note that data points belonging to the intersection of two subspaces lead to non-unique membership assignments. 
Similar to disjoint subspaces, sparsity is the only regularization that is proven to be effective for clustering intersecting subspaces. Two distinct two-dimensional planes in three dimensions is an example of intersecting subspaces.

The two most widely used algorithms for self-expressive based linear SC are the following: 
\begin{itemize}
	\item  Sparse Subspace Clustering 
	(SSC)~\cite{elhamifar2013sparse} uses the component-wise $\ell_1$ norm, that is, $||C||_1 = \sum_{i,j} |C_{i,j}|$, as a  convex surrogate of the $\ell_0$ norm to enhance the sparsity of $C$, 
	
	\item Low-Rank Representation (LRR)~\cite{liu2012robust} uses the nuclear norm, that is, $||C||_* = \sum_{i} \sigma_i(C)$ where $\sigma_i(C)$ is the $i$th singular value of $C$,  
	to promote~$C$ to have low rank.  
\end{itemize}   
SSC is the pioneer work in this context, and has strong theoretical guarantees in noisy and noiseless cases for independent, disjoint and intersecting subspaces \cite{elhamifar2013sparse, soltanolkotabi2012geometric, soltanolkotabi2014robust, you2015geometric}. 
LRR has shown competing results with SSC. However, its behavior in noisy and disjoint subspaces is still not well understood theoretically\footnote{Although the nuclear norm is a convex surrogate of the rank, it has different subspace preserving properties; 
	see~\cite[Example~1]{lu2012robust} for a numerical example. This is in contrast to the strong theoretical guarantees of the $\ell_0$ norm regularizations and its corresponding convex surrogate, the $\ell_1$ norm; see Table~\ref{coefreg}.}~\cite{wang2013provable}. 

\begin{remark}[Oversegmentation] 
	Accurate SC not only depends on subspace preserving representations but also on the connectivity of data points within the same subspace. Even though sparsity regularization has strong theoretical guarantees compared to other regularizations, it not only emphasizes the sparsity of between-cluster connections, but also the sparsity of the inner-cluster connections. Hence, it is possible that the data points
	from the same subspace form multiple connected components~\cite{nasihatkon2011graph}, which leads to the creation of more clusters than necessary, that is, oversegmentation. 
	The connectivity issue is less apparent in other regularizations as they inherently promote denser representations.
\end{remark}

\subsubsection{Post-Processing} \label{sec:post_proc}

In order to enhance the {quality} of the coefficient matrix $C$, several post-processing strategies exist. The goal of these approaches is to decrease the number of wrong between-subspaces connections and/or strengthening the within-subspaces connections. Even though there is no generally accepted strategy in the literature, a few common post-processing methods are as follows:

\begin{enumerate}
	
	\item Normalizing the columns of the coefficient matrix as $C(:,j) \leftarrow C(:,j)/||C(:,j)||_\infty$ where $||\cdot||_\infty$ is the infinity norm~\cite{elhamifar2013sparse}. This can be helpful when some data points have drastically different norms compared to other data points.
	
	\item Applying a hard thresholding operator on each column of the coefficient matrix by keeping only the $k$ largest entries 
	{in absolute value}~\cite{peng2016constructing}. 
	This post-processing is based on the property of
	Intra-subspace Projection Dominance (IPD) which states that the entries
	corresponding to data points in the same subspace are larger (in absolute value) than the ones corresponding to data points in different subspaces.
	
	\item Performing an ad-hoc post-processing strategy using multiple steps. First, a percentage of the top entries in each column of the coefficient matrix are preserved. 
	Next the shape interaction matrix
	method~\cite{costeira1995multi} is applied to remove the noise; it consists in a low-rank approximation of $C$ based on the skinny singular value decomposition.  
	Finally each entry is raised to some power larger than one (the value 4 is the often used) in order to intensify the dominant connections. This strategy depends on several parameters with no explicit theoretical justification. 
	However, it is widely used in nonlinear SC approaches based on neural networks~\cite{ji2014efficient,ji2017deep}. 
	
	\item Finding \emph{good} neighbors which correspond to the key connections in the coefficient matrix~\cite{yang2019subspace}. This approach is based on three parameters: 
	$k >  q  > p$. 
	First, only the $k$ largest coefficients are kept for each data point. Then, only a subset of these $k$ connections is preserved based on the notion of good neighbors. 
	A good neighbor is defined as a data point with at
	least $p$ (usually $p=1$) common data points inside the $k$ largest  connections. 
	The $q$ good neighbors of each data point with the largest coefficients form the final selected subset.
	A disadvantage of this approach is that it depends on parameters which are difficult to tune. (Note that using $q = k$ and $p=0$ reduces to the second post-processing approach described above.)
\end{enumerate}

\subsection{Spectral clustering using the coefficient matrix}

After obtaining the coefficient matrix $C$ using \eqref{generalsc}, the next step is to infer the clusters using spectral clustering~\cite{shi2000normalized,ng2002spectral}; 
see~\cite{von2007tutorial} for a tutorial. 
In fact, the entries in the coefficient matrix $C$ can be interpreted as the links between the data points: $C(i,j) \neq 0$ means that the data point $j$ is used for expressing data point $i$, and hence it is likely that the data points $i$ and $j$ belong to the same subspace. Therefore, the coefficient matrix $C$ corresponds to a directed graph structure $G=(V,E)$ where the nodes of the graph ($V$) are the data points and the weights of the edges ($E$) are determined by the entries in $C$. Using this interpretation, the problem of linear SC is mapped into the problem of segmenting the graph $G$. The symmetric adjacency matrix $A$ is constructed as $|C|+|C|^T$ and then the celebrated algorithm of spectral clustering is applied on the affinity matrix $A$ to partition the graph.

An important advantage of spectral based approaches is that they do not need to know the dimensions ($\{d_i\}_{i=1}^c$) of the subspaces. Moreover, by utilizing spectral clustering, they can estimate the number of clusters by analyzing the spectrum of the Laplacian matrix corresponding to the adjacency matrix $A$~\cite{von2007tutorial}. Furthermore, robustness of spectral clustering to small perturbations is advantageous for \emph{correct} clustering especially with unavoidable slight violation of subspace preservation in the coefficient matrix for real-world noisy cases~\cite{wang2016noisy}. \\

Linear SC based on exploiting the self-expressive property initiated an extensive amount of research in various directions. Several extensions of self-expressive based SC approaches have been proposed in the past decade to overcome notable challenges such as scalability~\cite{matsushima2019selective,
	chen2020stochastic,
	abdolali2019scalable,you2018scalable}, improving robustness~\cite{heckel2015robust,lu2013correntropy}, multi-view data clustering~\cite{gao2015multi,kang2020partition}, and the ability to deal with missing data~\cite{lane2019classifying}. However, a crucial limit to linear SC is the linearity assumption. In the next section, we review the approaches that were proposed to segment data points that are drawn from nonlinear manifolds. 

\section{Nonlinear subspace clustering} \label{nsc}


Linear SC algorithms often fail in dealing with data belonging to several nonlinear manifolds.
This is expected since these approaches only consider the \emph{global} linear relationship between data points.  
For nonlinear manifolds, the \emph{local} relationship between data points plays a more important role, as we will explain later in this section. But first, let us define the problem of nonlinear subspace clustering~\cite{elhamifar2011sparse}. 

\begin{definition}[Nonlinear subspace clustering (nonlinear SC)]\label{mc_def}
	Let $X \in \mathbb{R}^{d \times n}$ be the input  data. Suppose the data points lie on $c$ manifolds $\mathcal{M}_1, \mathcal{M}_2,\dots, \mathcal{M}_c$ with intrinsic dimensions\footnote{
		A manifold has dimension $d$ if every data point has a neighborhood homeomorphic to the Euclidean space in $\mathbb{R}^d$. 
		For example, a circle is  a one-dimensional manifold as the neighborhood of each point is locally a segment~\cite{daverman2009embeddings}.
	} 
	$d_1, d_2,\dots, d_c$ such that $d_i \ll d$ for all $i \in \{1,\dots,c\}$. The problem of nonlinear SC is defined as segmenting the data points based on their corresponding manifolds and obtaining a low-dimensional embedding of the data points within each manifold.
\end{definition}

\begin{figure}[h!]
	\begin{center}
		\includegraphics[width=0.6\textwidth]{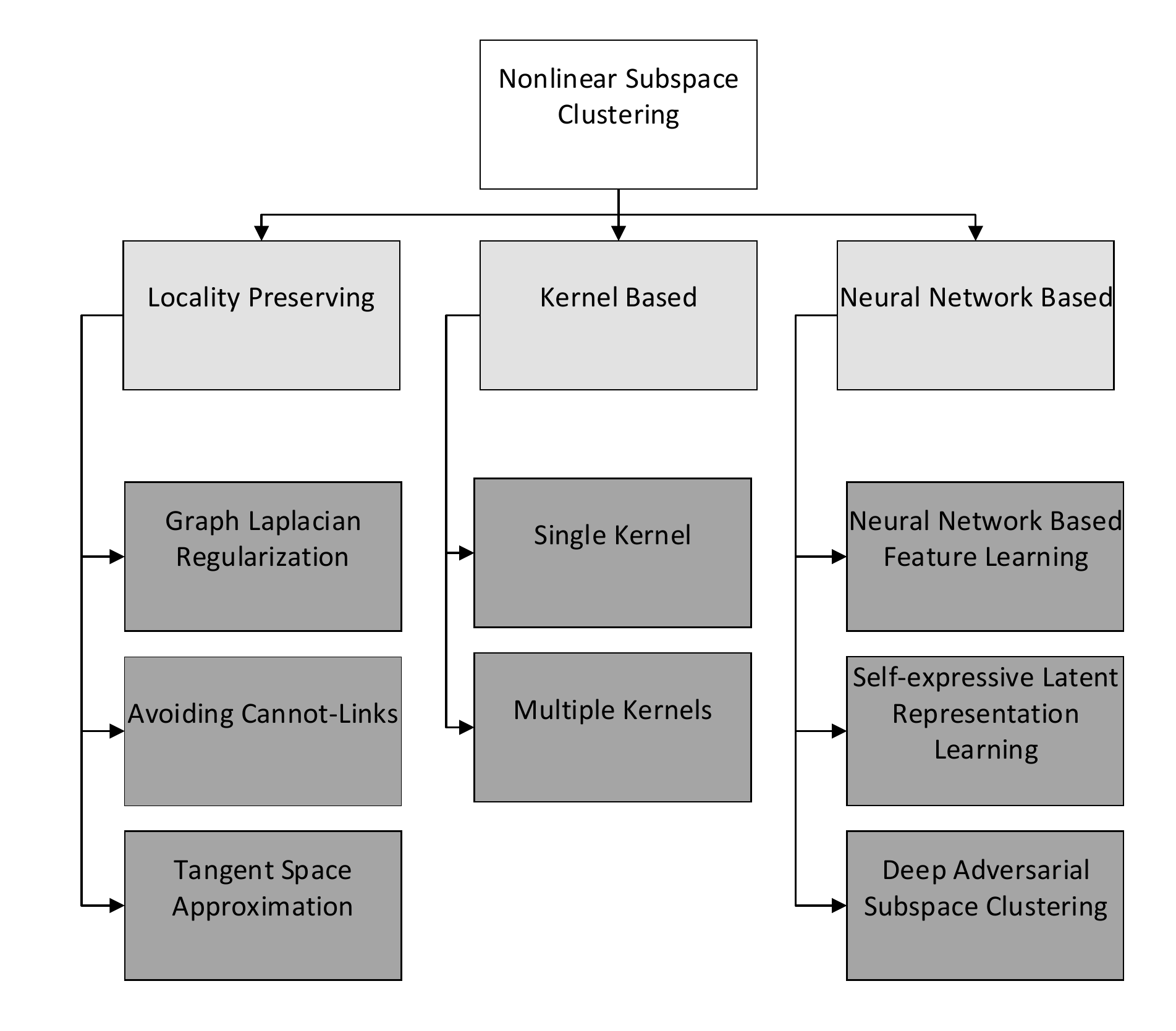}   
		\caption{Categorization of nonlinear SC approaches.	
		} 
		\label{tax}
	\end{center}
\end{figure}

Nonlinear SC is also referred to as  \emph{manifold clustering} in the literature~\cite{elhamifar2011sparse}. We mainly use the nonlinear SC term to emphasize that this survey focuses on the nonlinear approaches that extend the concepts in linear SC for data on nonlinear manifolds. 
Nonlinear SC approaches can be divided into three main categories: 
\begin{enumerate}
	
	\item locality preserving: they exploit the local geometric structure of the manifold to bridge the gap between \emph{thinking globally}, that is, using the global information from the whole data set, and \emph{fitting locally}, that is, using spatial local information around each point. 
	
	\item kernel based: they use implicit but predefined nonlinear mappings to transfer the data into a space where the linearity assumption is more likely to be satisfied.
	
	\item neural network based: they use the recent advances in structured neural networks to learn a nonlinear mapping of the data that respects the union of linear SC structure.  
	
\end{enumerate} 
Figure~\ref{tax} illustrates these three categories, along with several subcategories. 
In the following three sections, the major approaches within each category are presented. 

\subsection{Locality preserving nonlinear subspace clustering} \label{local}

A major disadvantage of linear SC approaches is that they are not \emph{faithful} to the data structure in the high-dimensional space. In particular, the self-expressiveness property does not guarantee that the nearby points in the ambient space have similar representations in the latent coefficient space~\cite{wang2010locality}, that is, a small value of $\|X(:,i) - X(:,j)\|$ does not imply that  $\| C(:,i)-W(:,j)\|$ is small. 
However, preserving the local configuration of the nearest neighbors of each point in the original ambient space is essential in being faithful to the data structure in nonlinear manifolds~\cite{roweis2000nonlinear}. In order to reflect the local structures, there are three major categories: 
graph Laplacian regularization, 
avoiding cannot links, 
and 
tangent space approximation. 

In fact, the root of these approaches comes from the single manifold learning algorithms based on preserving locality relationships among data points~\cite{saul2003think,roweis2000nonlinear,belkin2001laplacian}. Inspired by these classical algorithms, two assumptions are needed for these approaches to work (although  they are not always stated explicitly):  
\begin{itemize}
	
	\item[(i)] the underlying manifolds are \emph{smooth},  and
	
	\item[(ii)] the manifolds are \emph{well-sampled}, 
	that is, there are sufficiently many data points sampled in the neighborhood of each data point.

\end{itemize}  
Subsequently, these two assumptions imply that that each data point and the nearby points on the same manifold lie approximately on a \emph{local linear patch}\footnote{In single manifold learning, it is proven that under the smoothness and well-sampled assumptions, for a \emph{d-dimensional} manifold, each data point along with its $2d$ neighbors define an approximately linear patch of the manifold~\cite{saul2003think}.}. Hence, the nonlinear structure of the manifolds can be captured by the \emph{local linear} reconstruction of the data points. 
Let us now present the approaches within each of the three categories in the following sections.

\subsubsection{Graph Laplacian regularization} \label{sec:graphLap}

The methods in this category exploit the local structure of the data to construct the matrix $C$. To do so, a \emph{pairwise} local similarity matrix $S \in \mathbb{R}^{n \times n}$ is first constructed from the data points. The matrix $S$ is symmetric, 
and its $(i,j)$-th entry measures how \emph{similar} or \emph{close} the data points $X(:,i)$ and $X(:,j)$ are in the ambient space. 
The entries of $S$ are computed in different ways, including the following: 
\begin{enumerate}
	
	\item Binary k-nearest neighbors (K-NN): $S(i,j)$ is equal to one if $i$ is among the $k$ nearest neighbor of $j$, or vice versa~\cite{hu2014smooth}, that is, 
	\[
	S(i,j)=
	\begin{cases}
		1 \qquad \text{if} \ X(:,j) \in \mathcal{N}_k(X(:,i)) \ \text{ or} \ X(:,i) \in \mathcal{N}_k(X(:,j)), \\
		0 \qquad \text{otherwise},
	\end{cases}
	\]
	where $\mathcal{N}_k(x)$ computes the $k$ nearest samples of the data point $x$. 
	
	\item Continuous similarity functions: 
	the value $S(i,j)$ is computed using 		
	various similarity functions, such as Gaussain kernels~\cite{yang2014data,lu2013graph,zhang2020joint} given by  
	\[
	S(i,j)=e^{\frac{-||X(:,i)-X(:,j)||_2^2}{\sigma^2}},
	\]
	where $\sigma$ is a parameter. 
	Other data dependent similarity functions such as the geodesic distance for data on Grassmann manifolds can also be used~\cite{wang2016laplacian}. 
	
	\item Weighted K-NN is a combiantion of the two strategies described above~\cite{liu2014enhancing}: 
	\[
	S(i,j)=
	\begin{cases}
		\text{sim}(X(:,i),X(:,j)) \qquad X(:,j) \in \mathcal{N}_k(X(:,i)) \ \text{ or} \ X(:,i) \in \mathcal{N}_k(X(:,j)), \\
		0 \qquad \text{otherwise},
	\end{cases}
	\]
	where $\text{sim}\big(x,y\big)$ is a similarity function, such as Gaussian kernels, between two input vectors $x$ and $y$.
\end{enumerate}
For a recent comprehensive overview of similarity and neighborhood construction algorithms, we refer the interested  reader to~\cite{pourbahrami2020survey}. \\ 

Once the matrix $S$ is constructed, the following  regularizer for $C$ is constructed using this local similarity information: 
\begin{align} \label{local_cons}
	\mathcal{L}(C)=\frac{1}{2}\sum_{i,j}||C(:,i)-C(:,j)||_2^2 \ S(i,j).
\end{align}
Let the Laplacian matrix $L \in \mathbb{R}^{n \times n}$ corresponding to $S$ be defined as $L = D - S$ where $D$ is the diagonal matrix with $D_{ii}=\sum_j S(i,j)$ for all $i$. The function~\eqref{local_cons} can be rewritten as:
\begin{align} \label{lap_reg}
	\mathcal{L}(C) = \frac{1}{2}\sum_{i,j}||C(:,i)-C(:,j)||_F^2 \ S(i,j) &= \sum_i C(:,i)^\top C(:,i)D_{ii} - \sum_{i,j} C(:,i)^\top C(:,j) S(i,j) \\
	& = \tr(CDC^\top)-\tr(CSC^\top)=\tr(CLC^\top),  \nonumber
\end{align}
where $\tr(\cdot)$ denotes the trace of a matrix. 
This regularizer $\mathcal{L}(C)$ is known as the Graph Laplacian or manifold regularization. 
It is used within existing global linear SC optimization problems, that is, it is added in the objective function of the optimization problem~\eqref{generalsc} with a proper penalty parameter.  
Manifold regularization promotes a \emph{grouping effect}, which is defined as follows.  
\begin{definition}
	[Grouping Effect~\cite{hu2014smooth}] Given the input data matrix $X \in \mathbb{R}^{d \times n}$, let $C \in \mathbb{R}^{n \times n}$ be the obtained coefficient matrix corresponding to $X$ by an SC approach. The matrix $C$ has the grouping effect if $||C(:,i)-C(:,j)||_2$ goes to zero as $X(:,i)$ and $X(:,j)$ get closer, that is, as 
	$||X(:,i)-X(:,j)||_2$ goes to zero.   	 
\end{definition}
Intuitively, the grouping effect encourages locally similar points (that is, pairs of points whose corresponding value in $S$ is large) 
to have similar coefficient representations.

The grouping effect of Graph Laplacian regularizers generates graphs that better represent the structure of the data and hence are better connected~\cite{hu2014smooth}. This might benefit SC approaches based on self-expressiveness as they use spectral clustering for the final step~\cite{von2007tutorial}.
Note that the Graph Laplacian regularization is based on the assumption that the representation coefficient vector corresponding to each sample, that is, $C(:,i)$ for each $i=1,...,n$, is changing \emph{smoothly} on each manifold~\cite{belkin2006manifold}. 
Hence, nearby data points in the ambient space have similar coefficient vectors. 
This regularization also implicitly assumes that the data points from different clusters are not likely to be near each other and that sufficiently many data points are sampled from each manifold.

Graph Laplacian regularization can be used exclusively with the reconstruction error, $||X-XC||_F^2$, as in the Smooth Representation (SMR) algorithm~\cite{hu2014smooth}. 
SMR solves the following optimizatin problem: 
\[
\min_C \; \lambda||X-XC||_F^2 + \tr(CLC^\top).
\]
Note that SMR does not enforce $\diag(C) = 0$ explicitly, because the regularization $\tr(CLC^\top)$ prevents the identity matrix to be an optimal solution, for $\lambda$ sufficiently large, since the identity matrix $C=I$ satisfies $\|C(:,i) - C(:,j)\|_2 = 1$ for all $i \neq j$.

It is proven that Graph Laplacian regularization of the coefficient matrix leads to subspace preserving coefficients for noiseless {linear} independent subspaces~\cite{hu2014smooth}. Graph Laplacian regularization can be used in combination with other coefficient regularizations such as the $\ell_1$ norm~\cite{yang2014data}, or the nuclear norm~\cite{liu2014enhancing, yin2015dual}. 
In particular, using the $\ell_1$ norm along with Graph Laplacian regularization leads to the algorithm referred to as Laplacian Regularized $\ell_1$-SSC (LR$\ell_1$-SSC)  which solves the following optimization problem~\cite{yang2014data}: 
\[
\min_C \; ||C||_1 + \frac{\lambda_1}{2} ||X-XC||_F^2 +\lambda_2 \tr(CLC^\top) \quad \text{such that} \ \diag(C)=0.
\] 
However, as we will see in the numerical experiments in Section~\ref{eval}, promoting the smoothness of the representation over the manifold might not be sufficient to recover the complex nonlinear structures. Moreover, these methods are rather sensitive to the choice of the similarity function used to construct the pairwise local similarity matrix $S$.

\subsubsection{Avoiding cannot-links}

Instead of encouraging locally nearby  points to have similar coefficient representations, 
as presented in the previous section, 
one could encourage the entries of $C$ corresponding to \emph{faraway} points to have small/zero values. 
In other words, data points should not use faraway points in their representation, which we refer to as avoiding cannot-links. 
The name ``cannot-links" is borrowed from the graph learning and spectral clustering literature to indicate the pairwise constraints on the links that are encouraged/enforced to be avoided~\cite{lu2008constrained,xiang2008learning}.

Avoiding cannot-links attracted less attention compared to the Graph Laplacian regularization methods. Generally, the cannot-links refer to graph links/coefficients corresponding to distant points that are specified based on a dissimilarity criterion. There are two strategies to integrate pairwise cannot-links constraints: 
\begin{enumerate}
	\item Explicit elimination of faraway samples: A simple approach to prevent cannot-links is to explicitly add them as proper constraints in the optimization problem; for example, in \cite{zhuang2016locality}, Zhuang et al.\ use the following constraint: for all $j$,  
	\[
	C(i,j)=0 \; 
	\quad 
	\text{ for all }  X(:,i) \notin \mathcal{N}_k\big( X(:,j) \big). 
	\]
	A similar approach was proposed for linear SSC~\cite{han2015locality}. Instead of additional explicit constraints on the entries of the coefficient matrix, the self-expressiveness term is modified such that each sample is represented by the locally nearby data points. 
	
	\item Implicit penalizing of distant pairwise connections: Cannot-links information can be added as a 
	weighted penalty term. 
	For example, in~\cite{zheng2013low}, the following regularization term was proposed:
	\[ 
	\sum_{i,j} ||X(:,i)-X(:,j)||^2_2 \ |C(i,j)|.
	\]
	This term is similar to \eqref{local_cons}. However, for any two faraway data points $X(:,i)$ and $X(:,j)$, the corresponding $C(i,j)$ is encouraged to be small while this is not the case for graph Laplacian regularization (for which $S(i,j)$ will be close to zero). In other words, in graph Laplacian regularization, 		
	the nearby samples are encouraged to have similar coefficient representation but there is no explicit penalty on the coefficient representations corresponding to faraway data points. 
	{In fact, graph Laplacian regularization tend to produce dense coefficient matrices (see the numerical results in Section~\ref{synthetic_num}), hence do not have the disadvantage of oversegmentation that avoiding cannot-links may have~\cite{abdolali2019scalable}.} 
	
	A similar idea was presented 
	in~\cite{zhong2020nonnegative}, together with non-negativity constraint on the coefficient matrix $C$. 
	Another penalty term to penalize ``non-local" connectivities was proposed in~\cite{chen2019locality}, namely   
	$\sum_{i,j} e^{\dist(X(:,i),X(:,j))^2} C_{i,j}^2$,  
	where $\dist$ is a normalized 
	distance metric. 
\end{enumerate}
Avoiding non-local data points in self-expression might be intuitively similar to encouraging spatially close connections, as with the Graph Laplacian regularization. 
However, they lack the grouping effect of Graph Laplacian regularizations. 

\subsubsection{Tangent space approximation}

The approaches in this category rely on the estimation of the local tangent space for each data point, by fitting an affine subspace in the neighborhood of each data point. 
A simple strategy to approximate such tangent spaces is to consider a fixed predefined number of nearest neighbors for each sample, and approximate the tangent space by fitting an affine subspace to the data point and its neighbors. 
The crucial question is how to define the neighborhood of each data point in order to balance two goals: 
large enough to capture the local geometry of the manifolds, and small enough to avoid data points from other manifolds/clusters so as to preserve the curvature information of the underlying manifold. 

Despite the fact that choosing a \emph{fixed} value for neighborhood size with no prior knowledge is challenging, Deng et al.~\cite{deng2020low} proposed a multi-manifold embedding approach
based on the angles between approximated tangent spaces using a fixed neighborhood size. In particular, they assumed that if two data points belong to the same manifold, the angle between their corresponding tangent space is small whereas the angle is large if they belong to different manifolds. 
In contrast to estimating tangent space using fixed neighborhood size, a different strategy is to allow the size of the neighborhood to be arbitrary. 
This can be achieved via 
sparse representations in order to estimate the neighboring samples and simultaneously calculate the contribution weight of each neighborhood sample. These weights would be helpful to reflect the clustering information as well. 
In this section, we focus on this particularly effective strategy.

\paragraph{Sparsity based tangent estimation for manifold clustering}

In order to obtain the neighborhood of each point, Elhamifar and Vidal~\cite{elhamifar2011sparse} assumed that the tangent space of the manifold it belongs to can be approximated by the low-dimensional affine subspace constructed using a few neighboring points from the same manifold. 
Based on this assumption, they proposed an approach dubbed Sparse Manifold Clustering and Embedding (SMCE). 
It relies on two main principles, \emph{sparsity} and \emph{locality}, to estimate the tangent spaces. 

\begin{figure}[ht!]
	\begin{center}
		\includegraphics[width=0.7\textwidth]{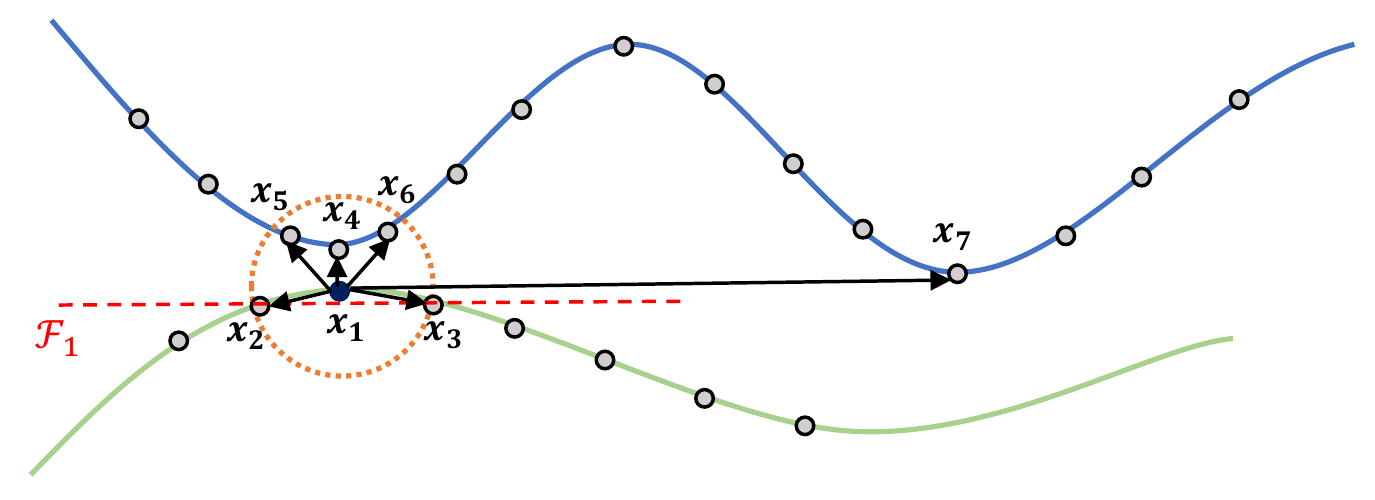}   
		\caption{Illustration of the necessity of sparsity and locality in manifold clustering. Local affine directions through sample $x_1$, that is, $x_2-x_1,\dots,x_7-x_1$ are shown with arrows. 
			The dotted circle is the smallest ball centered at $x_1$ and containing $x_2$ and $x_3$ from the same manifold. 
			Sparsity helps to avoid approximating the tangent space by close samples such as $x_4, x_5$ and $x_6$. 
			Locality excludes the use of $x_7$ which belongs to the other cluster. 
			The tangent space computed by SMCE is $\mathcal{F}_1$. 
		} 
		\label{proximity}
	\end{center}
\end{figure}

For an illustration of the main idea behind SMCE, consider the example in Figure~\ref{proximity}, showing two clusters of samples (gray circles) and their underlying manifolds. Suppose we want to approximate the tangent space for the sample $x_1$. Based on the principle of sparsity, the sparsest affine representation for the sample $x_1$ should contain two other distinct samples and among all possible combinations of two samples, 
$x_1$ is close to the affine span of $\{x_2,x_7\}$ and   $\{x_2,x_3\}$. 
To avoid representing $x_1$ using the affine subspace containing $x_7$, which belongs to the other manifold, the locality principle comes into effect to favor the affine span of $x_2$ and $x_3$. However, it should be noted that locality alone is not helpful to avoid representing $x_1$ using the points $\{x_4,x_5,x_6\}$ which are spatially close to $x_1$.  
This example illustrates the necessity of both sparsity and locality in the approximation of local tangent spaces  using the samples from the same manifold.

More precisely, SMCE works as follows. 
To construct the tangent space to $X(:,i)$, the normalized sample differences is first constructed as follows:
\[ 
U_i = 
\left[
\frac{X(:,1)\text{$-$}X(:,i)}{||X(:,1)\text{$-$}X(:,i)||_2}, 
\dots, 
\frac{X(:,i\text{$-$}1)\text{$-$}X(:,i)}{||X(:,i\text{$-$}1)\text{$-$}X(:,i)||_2}, 
\frac{X(:,i\text{$+$}1)\text{$-$}X(:,i)}{||X(:,i\text{$+$}1)\text{$-$}X(:,i)||_2}, 
\dots, 
\frac{X(:,n)\text{$-$}X(:,i)}{||X(:,n)\text{$-$}X(:,i)||_2} 
\right] . 
\] 
Based on these directions, the affine subspace $\mathcal{F}_i$ is represented using  
\[
\mathcal{F}_i = \big\{  X(:,i)+U_ic_i 
\, | \, e^\top c_i=1, c_i \in \mathbb{R}^{n-1} \big\},
\]
where $e$ is the all-one vector.
The goal is to estimate parameters of the local subspace $\mathcal{F}_i$ such that it has minimal distance to the unknown target tangent space. 
Hence, the  coefficients $c_i$ 
in the linear combination are calculated such that the distance between $X(:,i)$ and the affine subspace $\mathcal{F}_i$ is minimized: 
\[
\min_{c_i} ||X(:,i) - \big( X(:,i)+U_ic_i\big)||
\;  =  \; \min_{c_i} \ ||U_ic_i|| . 
\]
However, this formulation does not guarantee that the data points are expressed using locally nearby points from the same manifold. To estimate the tangent space using few nearby samples, SMCE optimizes the following problem for each point $X(:,i)$:
\begin{align}
	\min_{c_i} ||Q_ic_i||_1+\frac{\lambda}{2}||U_ic_i||_2^2 \quad \text{such that} \ e^\top c_i = 1,
\end{align}
where $Q_i \in \mathbb{R}^{(n-1) \times (n-1)}$ is called the proximity inducing matrix. The matrix $Q_i$ is a positive-definite diagonal matrix. The entries of $Q_i$ are defined by normalized $\ell_2$ distance between each data point $X(:,j)$ and $X(:,i)$ for $j\neq i$, that is, 
\[
Q_i(j,j)=\frac{||X(:,j)-X(:,i)||_2}{\sum_{t\neq i}||X(:,t)-X(:,i)||_2}.
\] 
Hence the diagonal entries of $Q_i$ that correspond to closer samples to $X(:,i)$ have smaller values, 
allowing the corresponding entries of $c_i$ to be larger. 
In contrast, the diagonal entries of $Q_i$ which correspond to farther samples have larger values, hence favoring smaller values of $c_i$. 
The nonzero entries of the vector $c_i$ indicate the points that are estimated to be on the same manifold as $X(:,i)$. Finally, 
the coefficient matrix $C$ is constructed as follows: $C(i,i)=0$ for all~$i$, and 
\begin{equation} \label{smce_graph}
	C(i,j) =  
	\frac{\frac{c_i(j)}{||X(:,j)-X(:,i)||_2}}{\sum_{t\neq i} \frac{c_i(t)}{||X(:,t)-X(:,i)||_2}} 
	\   \text{ for } \ j \neq i. 
\end{equation}
Applying spectral clustering on the corresponding symmetrized affinity matrix of $C$ reveals the final manifold clustering assignments.

A very similar approach to SMCE was presented in \cite{zhang2013semi}. However, 
the authors enforced locality and sparsity  in separate steps. 
First a sparse representation of the data points using the affine directions is computed  (without the proximity inducing matrix), 
and then the obtained sparse representation is weighted using the spatial locality information.

\subsection{Kernel based non-linear subspace clustering} \label{kernel}

\begin{figure}[ht!]
	\begin{center}
		\includegraphics[width=0.6\textwidth]{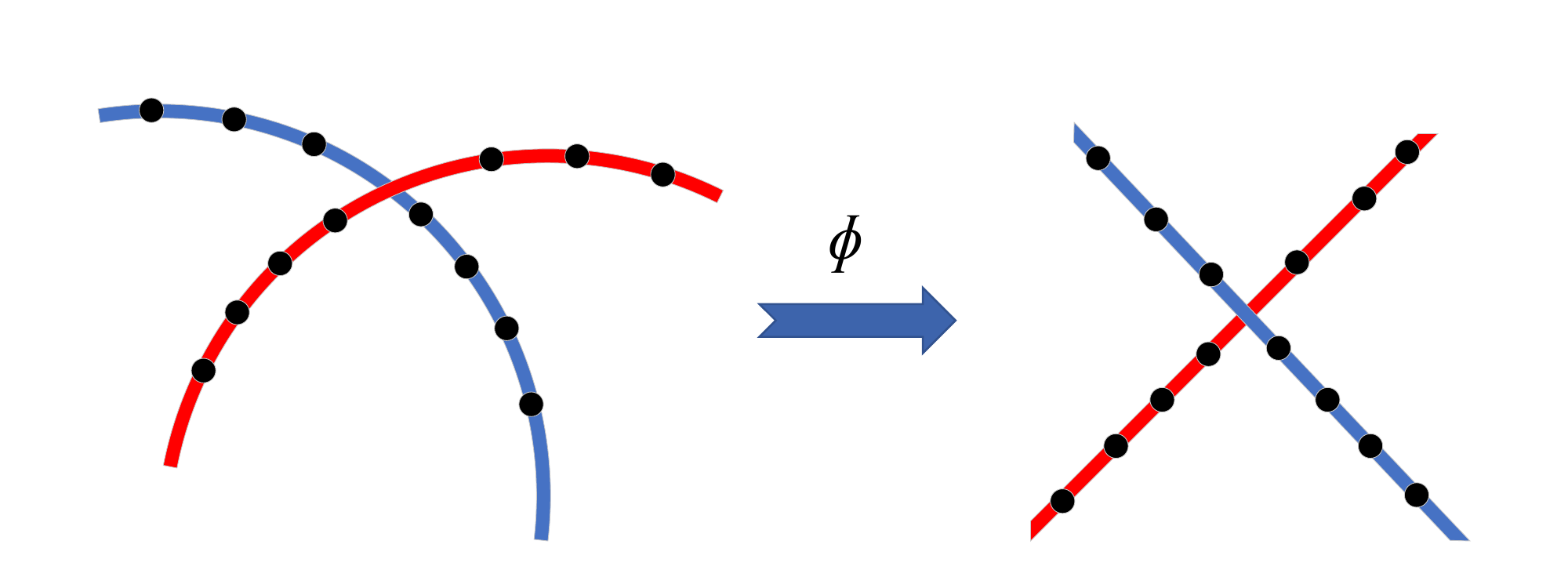}   
		\caption{Using implicit nonlinear transformations $\phi$ in kernel based approaches, the data is mapped to a space which might be more suitable for linear subspace clustering.} 
		\label{kernel_fig}
	\end{center}
\end{figure}

Kernel methods are commonly used to identify nonlinear structures and relationships among data points. They map the data from the input space $\mathcal{X}$ to the reproducing kernel Hilbert space $\mathcal{H}$ where the points might be better represented for a specific task. 
In nonlinear SC, the goal is to map the data points through a nonlinear transformation such that the linearity assumption is better satisfied; 
see Figure~\ref{kernel_fig} for a simple illustration. 
However, instead of explicitly using a nonlinear function, Kernel methods rely on the kernel trick to implicitly apply the nonlinear transformation. 
The kernel trick represents the data through pairwise similarity functions.  It avoids the often computationally expensive nonlinear transformations (feature maps), that is, the explicit computation of the coordinates of the  transformed data points in the feature space. 
Using this trick, only the inner products between all pairs of transformed data points in the feature space are computed.

Let $\phi: \mathcal{X} \rightarrow \mathcal{H}$ be the explicit nonlinear feature map. The kernel function $k : \mathcal{X} \times \mathcal{X} \rightarrow \mathbb{R}$ for $x,y \in \mathbb{R}^{d}$ is defined as:
$k(x,y)=\big< \phi(x),\phi(y) \big>$, where $\big<. , . \big>$ is the inner product. 
The matrix $\mathcal{K}_{XX} \in \mathbb{R}^{n \times n}$ is a positive semidefinite matrix, known as the Gram matrix, whose entries are defined as:
\begin{equation} \label{eq:kernel}
	\mathcal{K}_{XX}(i,j)=k(X(:,i),X(:,j))=\big<\phi(X(:,i)), \phi(X(:,j))\big>.
\end{equation} 
Using the kernel function, the explicit representation for the nonlinear transformation $\phi$ is no longer needed. 

In this section, we provide an overview of existing nonlinear SC based on kernel methods. These approaches can be divided into two categories:
single kernel and multiple kernels, 
and are presented in the next two sections.

\subsubsection{Using a single kernel} 

Patel et al.~\cite{patel2014kernel} proposed a kernelized SC approach by using the kernel trick in the (affine) sparse SC formulation. 
Their algorithm, dubbed Kernelized Sparse Subspace Clustering (KSSC), solves the following optimization problem 
\begin{align} \label{nonlinear_ssc}
	\min_C & \; ||C||_1+\frac{\lambda}{2} ||\Phi(X)-\Phi(X)C||_F^2 \quad 
	\text{such that} \quad \diag(C)=0, \ C^\top e=e, 
\end{align}
where $\Phi(X)=[\phi(X(:,1)),\phi(X(:,2)),\dots,\phi(X(:,n))]$. The kernel trick can be used to avoid computing $\Phi(X)$ explicitly, using 
\begin{align}	 
	||\Phi(X)-\Phi(X)C||_F^2  
	& = \tr\left( 
	\Phi(X)^\top \Phi(X) 
	\right)
	- 2 \tr\left(
	\Phi(X)^\top \Phi(X)C  
	\right)
	+  \tr\left(
	C^\top \Phi(X)^\top  \Phi(X)C 
	\right) \nonumber \\ 
	& = 
	\tr\left( 
	\mathcal{K}_{XX}-2\mathcal{K}_{XX}C+C^\top\mathcal{K}_{XX}C 
	\right),  \label{eq:kerneltrick} 
\end{align}
where the kernel $\mathcal{K}_{XX}$ is given in~\eqref{eq:kernel}. 
As for SSC, the authors used sparse regularization for the coefficient matrix. 
Finally, \eqref{nonlinear_ssc} can be reformulated as follows  
\begin{align} \label{kssc}
	\min_C &  \; ||C||_1+\frac{\lambda}{2} \  \tr\left(\mathcal{K}_{XX}-2\mathcal{K}_{XX}C+C^\top\mathcal{K}_{XX}C\right) \quad 
	\text{such that}   \quad \diag(C)=0, \ C^\top e=e. 
\end{align}

In order to reduce the computational cost of calculating the coefficient matrix for high-dimensional data, the same 
authors~\cite{patel2015latent} extended KSSC
to combine dimensionality reduction and kernelized SC. To this end, their approach learns the implicit mapping of the data onto a high-dimensional feature space using the kernel method, and then projecting onto a low-dimensional space via a linear projection. 

The ease of use of the kernel trick in the self-expressive based linear SC 
initiated the development of other nonlinear SC algorithms with additional regularizations, motivated by different purposes. 
These algorithms include for example the kernelized multi-view SC approaches from~\cite{zhang2020one, xie2020robust}. 
Recently Kang et al.~\cite{kang2020structure} proposed additional 
structure preserving regularization for KSSC in order to minimize the inconsistency between inner products of the transferred data, $\Phi(X)$, and inner products for reconstructed transferred data, $\Phi(X)C$. \\ 

However, the traditional challenges of kernel methods are still present for kernel based SC. The two main challenges are: 
\begin{enumerate}
	
	\item Using the Frobenius norm to compute the representation error facilitates the use of the kernel trick; see~\eqref{eq:kerneltrick}. 
	However, the Frobenius norm implicitly assumes that the noise follows a Gaussian distribution. 
	How can the kernel trick be applied on non-Gaussian noise models including, e.g., gross corruptions and  outliers? 
	
	\item With no prior knowledge, how can we select an appropriate kernel function and its parameters?
	
\end{enumerate}
In the rest of this section, we review the few approaches that tried to tackle these challenging limitations.

\paragraph{Robust kernelized nonlinear SC}

To improve the robustness, 
Xiao et al.~\cite{xiao2015robust} adopted the $\ell_{2,1}$ norm to replace the Frobenius norm in the kernelized LRR algorithm. 
The $\ell_{2,1}$ norm of the matrix $E \in \mathbb{R}^{d \times n}$ is defined as  \[
||E||_{2,1} = \sum_{i=1}^n ||E(:,i)||_2 = \sum_{i=1}^n \sqrt{E(:,i)^\top E(:,i)}. 
\] 
This norm, which enhances column-wise sparsity, models sample-specific outliers assuming that a fraction of the data points are corrupted. Based on this norm, \emph{robust} nonlinear LRR solves the following optimization problem: 
\begin{align} \label{l21_kernel}
	\min_C \; ||C||_*+\lambda ||\Phi(X)-\Phi(X)C||_{2,1}.
\end{align} 
Defining $P=I-C \in \mathbb{R}^{n \times n}$, \eqref{l21_kernel} can be reformulated as 
\begin{align}
	\min_{P,C} \ &||C||_* + \lambda \sum_{i=1}^n \sqrt{P(:,i)^\top \Phi(X)^\top \Phi(X) P(:,i)},  
\end{align}
which allows to use the kernel trick. 
To the best of our knowledge, no kernelized SC algorithm is proposed for other robust noise measurement norms such as the component-wise $\ell_1$ norm.

\paragraph{Adaptive kernel learning}

All the aforementioned methods (and in fact, the majority of kernel based algorithms) make use of standard predefined kernel functions, such as Gaussian RBF kernels $k(x,y)= \exp\big(-\frac{||x-y||_2^2}{2\sigma^2}\big)$, 
and polynomial kernels $k(x,y)=\big(x^\top y+a\big)^b$. However, due to lack of a criterion for measuring the \emph{quality} of a kernel function, selecting a proper kernel function is challenging. In the context of SC, a \emph{good} kernel should result in a union of linear subspaces in the implicit embedded space. 
Following this intuition, 
Ji et al.~\cite{ji2017adaptive} presented a kernel SC approach where they explicitly enforced the feature map $\Phi(X)$ to be of low rank and self-expressive:
\begin{align}
	\min_{C} \ &||\Phi(X)||_* + \lambda ||C||_1 
	\quad 
	\text{ such that }  
	\quad 
	\Phi(X) = \Phi(X) C, \quad \diag(C)=0, \quad C^\top e=e.\nonumber
\end{align}
However, this formulation has two limitations: 
(i)~applying the kernel trick on the first term of the objective function is not straightforward, 
and 
(ii)~the lack of constraints on the kernel leads to trivial solutions such as mapping all data points to zero. 
In order to overcome these issues, the authors exploited the symmetric positive definiteness of the kernel gram matrix $\mathcal{K}$, which implies that $\mathcal{K}=B^\top B$ for some square matrix $B$, while 
\begin{align} \label{kernel_trick}
	||\Phi(X)||_* = ||B||_*  \quad \text{for all} \  B \ \text{such that} \ \mathcal{K}=B^\top B.
\end{align}
Moreover, they restricted the unknown kernel matrix to be close (but not identical) to a predefined kernel matrix (such as Gaussian RBF or polynomial kernels) to avoid trivial solutions: 
\begin{align} \label{adaptive_kernel}
	\min_{C,B,\Phi} \ & ||B||_* + \lambda_1 ||C||_1 + \frac{\lambda_2}{2} ||\Phi(X) - \Phi(X) C||_F^2 + \lambda_3 ||K_G - B^\top B||_F^2,\\
	\text{ such that }  & \quad C^\top e=e, \quad \diag(C)=0, \quad \Phi(X)^\top \Phi(X)=B^\top B, \nonumber
\end{align}
where $K_G \in \mathbb{R}^{n \times n}$ is the predefined kernel matrix. Using the kernel trick~\eqref{eq:kerneltrick} and substituting $\Phi(X)^\top \Phi(X)$ by $B^\top B$, they solved this problem using ADMM. 
This formulation ensures that the transformed data points in the latent feature space are low-rank while an \emph{adaptive} kernel function is used. 
The optimization problem \eqref{adaptive_kernel} was further extended in \cite{xue2020robust} by regularizing the mapped features $\Phi(X)$ using the \emph{weighted Schatten p-norm} as a tighter nonconvex surrogate for the rank and using correntropy for more robust measurement of the difference between the predefined kernel $K_G$ and the estimated kernel $BB^\top$. The same idea was used in the robust nonlinear multi-view SC approach in \cite{zhang2019robust}.

However, these approaches suffer from two limitations: (i)~they still need a predefined kernel matrix, $K_G$, 
and 
(ii)~assuming that a ``good" kernel leads to a low-rank embedding of the data points is rather strong for data on multiple manifolds. In particular, there is no guarantee that learning a low-rank kernel corresponds to an implicit nonlinear transformation that preserves sufficient separability between the data points from different manifolds. In other words, a low-rank kernel (even with the additional self-expressiveness structure) does not necessary lead to a well separated embedding of the data points.

\subsubsection{Using multiple kernels} \label{multiple_kernel} 

Even though kernel methods are considered principled approaches to provide non-linearity in linear models, 
they rely on selecting and tuning a kernel. 
In many (unsupervised) applications, with no prior knowledge, this is a challenging problem. Multiple Kernel Learning (MKL) methods overshadow this issue, 
by learning a consensus kernel $\mathcal{K}$ from a set of predefined candidate kernels $\{\mathcal{K}^{(i)}_G\}_{i=1}^k$, where $k$ is the number of given kernels. 
The sought kernel $\mathcal{K}$ is regularized towards the predefined candidate kernels via a penalty function $h\left(\mathcal{K},\{\mathcal{K}_G^{(i)}\}_{i=1}^k,w \right)$ which is added as a penalty term in the objective function. The vector $w \in \mathbb{R}^k_+$ contains the weights/importance of the predefined kernels.
This regularization function is typically defined in two ways: 

\begin{enumerate}
	\item Centroid-based MKL~\cite{kang2018self} encourages the consensus kernel $\mathcal{K}$ to be close to every predefined kernel by minimizing  
	\begin{align} \label{adaptive_MKL1}
		h\left(\mathcal{K},\{\mathcal{K}_G^{(i)}\}_{i=1}^k,w \right) 
		= 
		\sum_{i=1}^k w(i) \left\|\mathcal{K}-K^{(i)}_G \right\|_F^2. 
	\end{align}	
	\item Linear combination MKL~\cite{ren2020multiple} encourages the consensus kernel to be approximately a convex combination of the predefined kernels:
	\begin{align} \label{adaptive_MKL2}
		h\left(\mathcal{K},\{\mathcal{K}_G^{(i)}\}_{i=1}^k,w \right)  
		=  
		\left\| \mathcal{K}-\sum_{i=1}^k w(i)\mathcal{K}_G^{(i)} \right\|_F^2 \quad \text{ where } \ w\geq 0, \ \sum_{i=1}^k w(i)=1.
	\end{align}
\end{enumerate}

The main difference between MKL based approaches is the enforced property for the ideal consensus kernel, using different regularizations. 
This relates to the criteria used to define the ``goodness" of the consensus kernel, 
and the two most widely used ones promote 
\begin{itemize}
	\item a block-diagonal coefficient matrix, or 		
	\item a low-rank consensus kernel.
\end{itemize}	 
Let us discuss these two strategies in the next two paragraphs.

\paragraph{MKL encouraging a block-diagonal representation} 
A criterion for quantifying the quality of the unknown consensus kernel is based on the assumption that it should encourage a block diagonal representation for the coefficient matrix (up to permutation), where each block corresponds to a different manifold.
In fact, the coefficient matrix $C$ is block diagonal, up to permutation, if and only if the subspace preserving property holds (see Definition~\ref{sp_def}).

A common approach to promote a block-diagonal representation is to use a proper regularizer for the coefficient matrix. 
Let us define the Laplacian matrix $L$ associated to $C$ as $L=\diag(Ae)-A$ where $A = |C|+|C|^\top$ is the affinity matrix corresponding to $C$.   
Then, minimizing the  sum of the $c$ smallest eigenvalues of $L$, that is, $\sum\nolimits_{i=1}^{c}\sigma_i(L)$ where $\sigma_i(L)$ is the $i$-th smallest eigenvalue of $L$, promotes $C$ to be block diagonal. In fact, the multiplicity of the zero eigenvalue of the Laplacian matrix is equal to the number of connected components in the  graph corresponding to the matrix $C$~\cite{von2007tutorial}. 
Using the Ky Fan theorem, 
Lu et al~\cite{lu2018subspace} write this nonconvex regularizer into a convex counterpart. 

As one of the first approaches presented for MKL based nonlinear SC, Kang et al.~\cite{kang2018self} proposed the Self-weighted
Multiple Kernel Learning (SMKL) algorithm. SMKL combines the multi-kernel learning formulation in \eqref{adaptive_MKL1} with the block-diagonal regularizer in the kernelized LSR formulation~\cite{lu2012robust}\footnote{LSR is a linear SC algorithm that regularizes the coefficient matrix $C$ with the Frobenius norm; see Table~\ref{coefreg}.}: 
\begin{align} \label{sc_MKL1} 
	\min_{C,\mathcal{K}}  \ &\underbrace{||\Phi(X)-\Phi(X)C||_F^2 + \lambda_1 ||C||_F^2}_{\text{kernel based LSR self-expressiveness}} + \underbrace{\lambda_2 \sum\nolimits_{i=1}^k w(i)||\mathcal{K}-K^{(i)}_G||_F^2}_{\text{Multiple kernel learning }}+\underbrace{\lambda_3\sum\nolimits_{i=1}^{c}\sigma_i(L)}_{\text{block-diagonal}} , \\ \nonumber
	& \text{ such that } 
	\ C \geq 0, 
	\ C = C^\top, 
	\ \mathcal{K} = \Phi(X)^\top \Phi(X) 
	\ \text{and} 
	\ L=\diag(Ce)-C, 
\end{align} 

where the kernel trick can be easily applied on the first term in the objective function; see \eqref{eq:kerneltrick}.  
An iterative alternative optimization approach was used to solve \eqref{sc_MKL1}. 
In each iteration, the entries of the weight vector $w$ are updated adaptively during the optimization process using 
\[
w(i)= \frac{1}{2||\mathcal{K}-K^{(i)}_G||_F^2}.
\]
However, in our implementations, we noticed that it might prevent convergence, because the parameter $w$ in~\eqref{sc_MKL1} is modified at each iteration. 
Moreover, with no constraint on the summation of the weights, they tend to have very small values. 
\begin{remark}[Nonnegativity] 
	The nonnegativity and symmetry constraints in~ \eqref{sc_MKL1} make the coefficient matrix $C$ a valid affinity matrix~\cite{lu2018subspace}, with   
	$L=\diag(Ce)-C$ being the corresponding Laplacian matrix. 
	However, with the nonnegativity constraint on the entries of the matrix $C$, the optimal solution may not be subspace preserving for the ``extreme points" within the subspaces (since they are not contained within the conical hull of points from the same subspace). 
	As discussed for affine SC in \cite{li2018geometric}, 
	the spectral clustering step is however likely to generate correct clusters. This is justified by the ``collaborative effect" phenomenon which states that the coefficient representation corresponding to non-extreme points can ``pull" the extreme points toward the data points corresponding to the same subspace, and hence make the spectral clustering robust to the wrong connections. However, this depends on the strength of the within subspace connectivity of non-extreme data points and the number of the extreme data points. 
\end{remark}

Similar formulations were presented in \cite{yang2019joint,zheng2020robust,ren2020simultaneous}. 
In these approaches, the  update  of $w$ was performed using variations of the  Correntropy Induced Metric (CIM) for measuring the distance between entries of the kernels, that is, $w(i)=CIM(\mathcal{K},K_G^{(i)})$. The CIM between two matrices $X\in \mathbb{R}^{N \times N}$ and $Y\in \mathbb{R}^{N \times N}$ is defined as $CIM(X,Y)= 1-\frac{1}{N^2}\sum_{i=1}^N\sum_{j=1}^N G(X(i,j)-Y(i,j))$ where $G(x)=\exp(-x^2/2\sigma^2)$, and is known to be a more robust distance metric compared to the Frobenius norm~\cite{lu2013correntropy}, and hence might lead to a more accurate weighting assignment.
In particular, \cite{yang2019joint} used the mere block-diagonal regularizer and eliminated the Frobenius norm regularizer for the coefficient matrix. This is motivated by the fact that block-diagonal regularizer is proven to be subspace preserving for independent subspaces (similar to the Frobenius norm)~\cite{8259470}. 
The same idea was used for multi-view nonlinear SC based on MKL in \cite{zheng2020robust}. 
Ren et al.~\cite{ren2020simultaneous} introduced an additional regularization to learn the affinity and coefficient matrices  simultaneously.

\paragraph{MKL encouraging a low-rank consensus kernel}

Although the data might be nonlinear in the original ambient space, the linear subspace structure should be present in the implicit embedding space. Hence a proper kernel should implicitly lead to a low-rank embedded data.
Based on this intuition, 
Kang et al.~\cite{kang2019low}, proposed Low-rank Kernel learning for
Graph matrix (LKG) that learns a \emph{low-rank consensus kernel} from a weighted linear combination of the given kernels  by solving the following optimization problem: 
\begin{align} \label{sc_MKL2}
	\min_{C,\mathcal{K},w}  \ &\underbrace{\frac{1}{2}||\Phi(X)-\Phi(X)C||_F^2 + \lambda_1 f(C)}_{\text{kernel based self-expressiveness}} + \underbrace{\lambda_2 ||\mathcal{K}-\sum\nolimits_{i=1}^k w(i)\mathcal{K}_G^{(i)}||_F^2}_{\text{Multiple kernel learning }}+ \ \underbrace{\lambda_3 \ ||\mathcal{K}||_*}_{\text{low-rank kernel}} ,\\ \nonumber
	& \text{ such that } \ C \geq 0, \ \mathcal{K} = \Phi(X)^\top \Phi(X) \geq 0, \ w\geq 0, \ \sum_{i=1}^k w(i)=1.
\end{align}
However, \cite{ren2020multiple} explained that minimizing $||\mathcal{K}||_*$ does not necessarily lead to a low-rank transformed data in the implicit feature space, that is, a small value for $||\Phi(X)||_*$. In fact, we have 
\[
||\mathcal{K}||_* = \tr \sqrt{\mathcal{K}^\top\mathcal{K}}=\tr \big(\Phi(X)^\top \Phi(X)\big) =  ||\Phi(X)||_F^2.
\]
Hence, using the observation in \eqref{kernel_trick}, 
and defining an auxiliary square matrix $B$ such that $\mathcal{K}=B^\top B$, 
they enforced the low-rankness of $\Phi(X)$  by minimizing the nuclear norm of $B$~\cite{ren2020multiple}.

\subsection{Neural network based nonlinear SC} \label{deep}

Neural networks have emerged as a very powerful representation learning framework and have drawn substantial attention due to many successful applications in different fields. The past four years have witnessed an increasing number of papers which tried to merge the representational power of neural networks with classical linear SC algorithms to deal with data from nonlinear manifolds. In a nutshell, the main motivation of neural networks based SC approaches is to overcome the fundamental limitation of kernel based alternatives, by  \emph{learning the proper embedding of the data from the data}.  Geometrically, as for Kernel-based approaches, this proper embedding should ideally have a union of linear subspaces structure where classic linear self-expressiveness leads to subspace preserving representations. 
These algorithms can be partitioned into three main categories (see also Table~\ref{tax}, page~\pageref{tax}):  
\begin{enumerate}
	\item \textbf{Neural network based feature learning} that  extract nonlinear features from the data using neural networks as a preprocessing step for linear SC algorithms,  
	
	\item \textbf{Self-expressive latent representation learning} that enforce explicitly the neural network to learn self-expressive latent representation either by changing the architecture or the objective function, and 
	
	\item \textbf{Deep adversarial SC} that uses generative adversarial networks to perform nonlinear SC. 
	
\end{enumerate}
In the next three sections, we describe these three categories.

\subsubsection{Neural network based feature learning} 

Early attempts for employing neural network for nonlinear SC were mainly limited to providing better, that is, \emph{more discriminative}, features (representations) for the data points. 
In fact, several SC papers in the literature (including linear  and nonlinear  methodologies) still rely on manually-designed/handcrafted features, such as Local Binary Patterns (LBP)~\cite{ojala1994performance}, Histogram of Oriented Gradients (HOG)~\cite{dalal2005histograms}, and Scale-Invariant Feature Transform (SIFT)~\cite{lowe1999object}, 
for satisfactory results on real-world data sets. 
However, there is no theoretical justification on how any of these handcrafted features might affect the geometric structure of the data lying on multiple (nonlinear) subspaces. From the perspective of unsupervised feature learning, features extracted by neural networks have an important advantage over traditional handcrafted features: they are specifically \emph{learned from the data}.  

The focus of neural network based approaches for SC is to \emph{learn} proper representations/features from the data points for the specific task of SC. 
The main shared characteristic of these approaches is that they treat feature learning and clustering as \emph{separate tasks}, which is suboptimal. They are divided into two main subcategories: (i)~transforming the unsupervised problem into a (self)-supervised problem, and 
(ii)~using the connectivity information from the output of a classical linear SC algorithm as prior knowledge. 
Let us discuss, in the next paragraphs, the approaches within each category in detail.

\paragraph{Self-supervised subspace clustering}

The success of \emph{supervised} feature learning in neural networks inspired a few works to map the unsupervised SC task to a (self-)supervised problem. To do so, 
the clustering output of a \emph{linear} SC algorithm can be used as labels to provide supervision~\cite{sekmen2017unsupervised, zhou2018iterative}.  
However, these labels are highly unreliable as they are obtained by applying algorithms based on the linearity assumption. 
There are two strategies to exploit the information from these \emph{noisy} labels to train a network: 
\begin{itemize}
	\item Use the concept of ``confident learning": 
	a confidence weight is assigned to each data point quantifying its  likeliness of having the correct label~\cite{northcutt2019confident}. 
	The highly confident samples and their corresponding labels can be used to train a neural network in a supervised fashion. 
	Sekmen et al.~\cite{sekmen2017unsupervised} used the criterion of the distance of the samples to the obtained subspaces from a linear SC algorithm to choose highly confident data points for the supervision.  
	However, there are serious drawbacks to this approach. The weighting procedure based on the criterion of distance to the estimated subspaces using linear algorithms is not reliable for nonlinear data. Moreover, selecting the highly confident labels depends on the value of a threshold which is not easy to set with no prior knowledge. However, we would like to emphasize that confident learning is an emerging topic in robust deep learning given noisy labels with significant theoretical and numerical advances over the past couple of years~\cite{jiang2020beyond,ma2020normalized}. 
	However, how to select the ``confident samples'' and use them for nonlinear manifold clustering task requires more in-depth study.
	
	\item Increase the number of clusters incrementally: To reduce the effect of wrong labels, Zhou et al.~\cite{zhou2018iterative} suggested an iterative approach. Specifically, at the first iteration, the linear SSC algorithm is applied on the (raw) input data and the spectral clustering is applied on the coefficient matrix to segment the data in \emph{two} clusters. A neural network is trained with the obtained labels (clusters) for a binary classification problem. In the following iterations, the features from the previously trained network are used as input data to the SSC and the number of clusters to segment the corresponding graph (in the spectral clustering step) is increased by one. The network parameters are updated using the obtained labels as a supervised classification problem. This iterative process continues until the number of clusters reaches the desired predefined number. However, there is no theoretical evidence or justification that this strategy can improve the robustness to the erroneous output of linear SC algorithms.
\end{itemize}

\paragraph{Linear based connectivity as prior knowledge}

Another strategy to leverage linear SC algorithms as a prior information is to use the coefficient matrix to guide the geometric structure of the embedded data points in the latent space of neural networks. 

As the pioneer algorithm, Peng et al.~\cite{peng2018structured,peng2016deep} proposed a structured autoencoder, dubbed \mbox{$\text{StructAE}$}, where self-expressiveness of the latent representations is incorporated in the loss function of the network. StructAE relies on a previously computed coefficient matrix $C$ by a classical linear SC algorithm. 
The authors suggested the use of SSC and LSR, leading to StructAE-L1 and StructAE-L2 respectively\footnote{Recall that SSC and LSR use the $\ell_1$ and $\ell_2$ norms for regularization of the columns of the coefficient matrix, respectively; see Section~\ref{sec:selfexp}.}. Chen et al.~\cite{chen2018subspace} used LRR instead to construct the matrix $C$. 
Given $C$, consider an autoencoder with $M$ fully-connected hidden layers, where the first $M/2$ layers indicate the encoder and the last $M/2$ layers indicate the decoder. We denote $\Theta$ the parameters of the network, that is, the weights and biases of each layer. Let us also denote the compact latent representation of the encoder output as $Z_\Theta(:,i)$ for the input data $X(:,i)$, and, the reconstructed data, which is the output of the decoder, by $\tilde{X}_\Theta$. They both depend on the parameters $\Theta$.

The optimization problem is defined as follows 
\begin{align} \label{structAE}
	\min_{\Theta} \ &\frac{1}{2} ||X-\tilde{X}_\Theta||_F^2 + \frac{\lambda_1}{2}||Z_\Theta-Z_\Theta C||_F^2 + \frac{\lambda_2}{2}||\Theta||_F^2, 
\end{align}
where the first term is the reconstruction error of the autoencoder, 
the second term enforces the latent representation $Z$ to respect the multiple subspace structure encoded by the matrix $C$, and 
the third term is regularizing the parameters of the network to avoid overfitting. 
The coefficient matrix $C$ contains prior information on the \emph{global structure} of the data, because it is obtained by minimizing $||X-XC||_F^2$, 
via a linear SC approach,  containing the information of all data points using self-expressiveness. 
The clustering membership is obtained by clustering the latent representation $Z$ using algorithms such as k-means or by applying a linear SC algorithm such as SSC and LSR.
The main motivation of StructAE is to integrate the individual and global structures together, and encourage the autoencoder to learn features/latent representation respecting the geometric structure provided by the coefficient matrix.
However, the matrix $C$ is obtained by a \emph{linear} SC based on a global linearity assumption. Hence 
the performance tends to significantly depend on how well the linearity assumption is satisfied.

\subsubsection{Self-expressive latent representation learning}

Separating the two modules of representation/feature learning and (subspace) clustering assignment typically leads to suboptimal results. To overcome this limitation, a group of approaches explicitly include the self-expressiveness term in the network loss function. These approaches are based on the intuition that minimizing self-expressiveness in the embedded space encourages the neural network to learn the \emph{appropriate} embedding for the task of SC. 
In general, the optimization problem of these networks has the following form: 
\begin{align}  \label{nn_selfexpressive}  
	\min_{C, \Theta} & \; \frac{1}{2} ||Z_\Theta-Z_\Theta C||_F^2 + \lambda_1 f(C) + \lambda_2 h(X,Z_\Theta,C)
	\quad \text{ such that }  
	\quad 
	Z_\Theta = \Phi_e(X,\Theta) 
	\text{ and } 
	\diag(C)=0, 
\end{align}  
where $\Phi_e$ and $\Theta$ denote the network embedding and parameters, respectively.  
The embedding of the network, which is denoted by $Z_\Theta$, is encouraged to have a union of linear subspaces structure by explicitly minmizing the self-expressive representation term.
As in the rest of the paper, $f(C)$ is the regularization on the coefficient matrix. 
The extra regularization $h(X,Z_\Theta,C)$ plays a critical role in removing trivial solutions and specifying the properties of the nonlinear mapping and the embedding space. In particular, looking for self-expressiveness in the transformed space is not sufficient. There exists infinite nonlinear transformations that can lead to small/zero self-expressive error, that is, 
$||Z_\Theta-Z_\Theta C||_F \approx 0$. Hence, the main difference between these approaches lies on the regularization function $h$ that reduces the possible nonlinear mappings space and forms the geometric structure of the data embedding. There are mainly four choices for $h$ in the literature: 
\begin{enumerate}
	
	\item Instant normalization~\cite{peng2020deep} promotes the norm of the embedded data points in the latent space to be close to 1 using the regularization $\frac{1}{2}\sum_{i=1}^n ||Z_\Theta(:,i)^\top Z_\Theta(:,i)-1||_2^2$. 
	This avoids an arbitrary
	scaling factor in the embedding space. 
	
	\item Locality preservation~\cite{zhu2018nonlinear} captures the local structure of data through Graph Laplacian regularization, similarly as done in Section~\ref{sec:graphLap}. This regularization is defined as $\frac{1}{2} \tr(CLC^\top)$,
	where $L$ is the Laplacian matrix defined from a similarity matrix over the data points $X$. 
	
	\item Autoencoder reconstruction loss~\cite{ji2017deep} is based on an autoencoder architecture and uses a decoder network to reconstruct the original data from the self-expressive representation, that is, $Z_\Theta C$. More precisely, the regularization is defined as $||X-\tilde{X}_\Theta||_F^2$ where $\tilde{X}_\Theta=\Phi_d(ZC,\Theta)$, and $\Phi_d$ represents the decoding mapping which depends on the parameters $\Theta$. 
	This regularization that  minimizes the reconstruction error of the input matrix ensures that the self-expressive based embedding preserves sufficient information to recover the original data. 
	
	\item Restricted transformations~\cite{maggu2020deeply} is based on learning stacked linear transformations through multiple layers that are connected via non-negativity constraints inspired by the rectified linear unit (ReLu) that sets negative values to zero after each layer~\cite{hahnloser2001permitted, glorot2011deep}. 
	Specifically, Maggu et al.~\cite{maggu2020deeply} 
	proposed a three layered transformation for SC, solving the following optimization problem: 
	\begin{align}
		\min_{\Theta = \{T_i\}_{i=1}^3, C}  \ & \ \underbrace{||T_3T_2T_1X-Z_\Theta||_F^2}_{\text{deep transformed data: $Z_\Theta \approx \Phi_e(X,\{T_i\}_{i=1}^3$)}}  + \underbrace{\lambda_1 ||Z_\Theta-Z_\Theta C||_F^2 + f(C)}_{\text{self-expressiveness in latent space}} \\ & + \underbrace{\lambda_2 \sum_{i=1}^3 \big(||T_i||^2_F - \logdet T_i\big),}_{\text{avoid trivial transformations}} \quad
		\text{ such that } \ \underbrace{ T_2T_1X \geq 0 \ \text{and} \ T_1X \geq 0}_{\text{nonnegative representations}}, \nonumber
	\end{align}
	where $\{T_i\}_{i=1}^3$ are the linear transformations such that the dimension of each layer output is reduced. 
	To avoid trivial or degenerate solutions ($T_i \rightarrow 0 \ \text{or} \ T_i \rightarrow \infty$), the linear transformation corresponding to each layer is penalized by minimizing the Frobenius norm minus the logarithm of the determinant of the transformation matrices. 
	Note that the non-negativity constraints are inspired by the ReLU activation but they are inherently different. 
	
\end{enumerate}

Undoubtedly, the deep SC approaches based on autoencoder regularization are the most popular. A large number of algorithms have been proposed using this regularization. In the rest of this section, we focus on these approaches.

\paragraph{Autoencoder regularized deep subspace clustering}

The autoencoder based feature extraction for SC was first proposed as the Cascade Subspace Clustering (CSC) algorithm 
in~\cite{peng2017cascade}. In a nutshell, CSC first learns compact features by pretraining a simple autoencoder, 
and then discards the decoder part. Afterwards, it fine-tunes the parameters of the encoder with a novel loss function based on the  \emph{invariance of distribution} property. The invariance of distribution is based on the assumption that the conditional distribution of any data point given the cluster centers is invariant to different distance metrics. The conditional distribution for each data point is a $c$-dimensional vector providing a probability that is related to the closeness between the given data point and the centroids of the $c$ clusters, so that the data point has a higher probability to be assigned to a closer cluster.  
The closeness is measured by different metrics  such as the Euclidean and cosine similarity metrics. The invariance of distribution uses the KL-divergence to minimize the discrepancy among different distributions defined by different closeness measures. After updating the encoder parameters and the cluster centroids in the fine-tuning step, each data point is assigned to the cluster with the closest centroid. 
Nevertheless, it is not clear how minimizing the discrepancy among different distributions benefits the SC or affects the geometric structure of the data from multiple manifolds. 
Moreover, CSC only uses the autoencoder network for the initialization of the parameters of the encoder.

\begin{figure}[ht!]
	\begin{center}
		\includegraphics[width=0.8\textwidth]{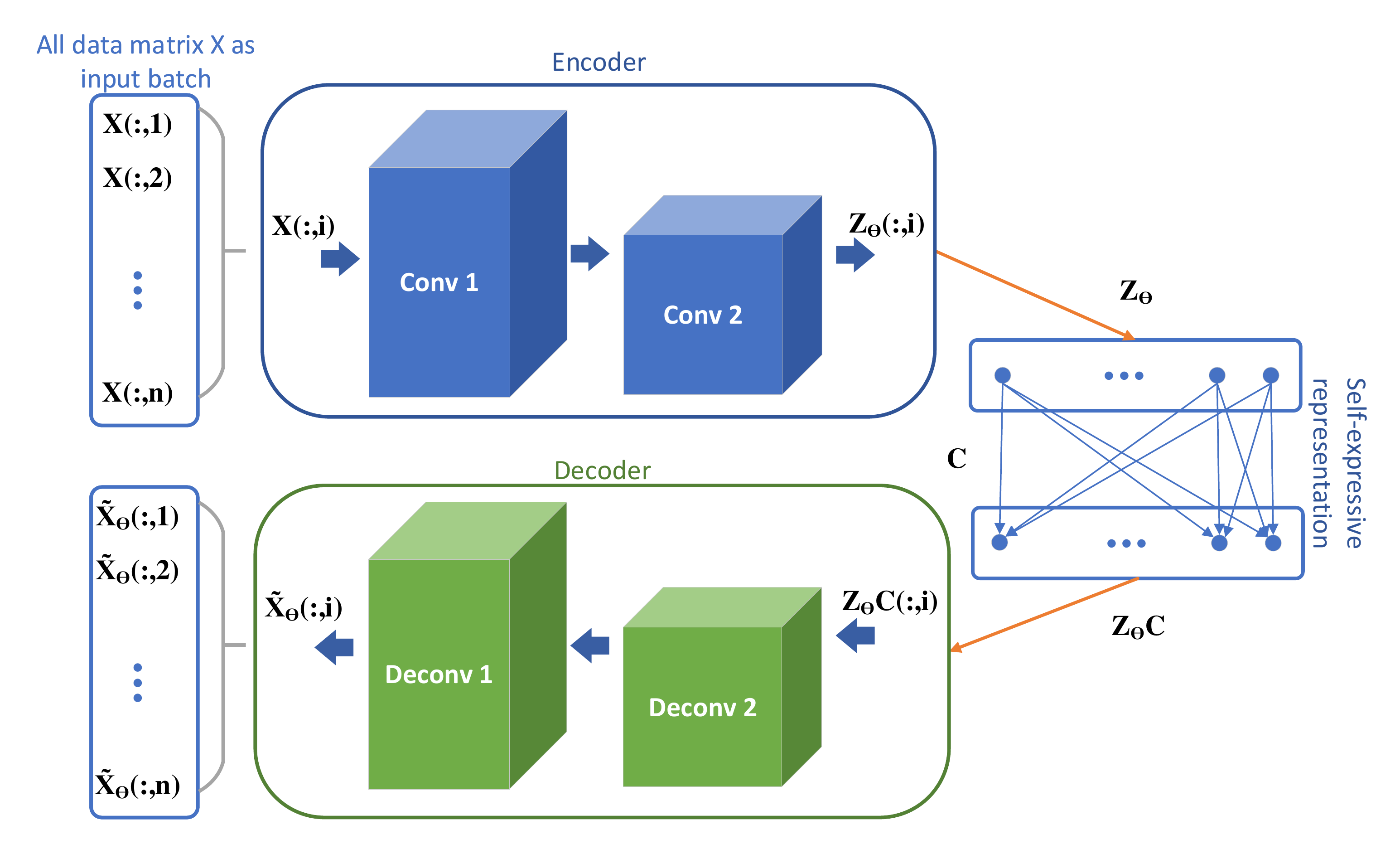}   
		\caption{Illustration of autoencoder based subspace clustering using fully connected linear self-expressive layer in DSC-Net~\cite{ji2017deep}. Two layers of convolutional encoder and two layers of deconvolutional layers are shown for the decoder. All the input data are used as a large single batch for minimizing the loss function and training the autoencoder. 
		}  
		\label{DSC}
	\end{center}
\end{figure}

The idea of explicitly merging self-expressiveness in autoencoder networks for SC was first brought to light in the framework of Deep Subspace Clustering network 
(DSC-Net)~\cite{ji2017deep}. DSC-Net is the pioneer work to identify that the linear combination in the collaborative representation corresponds to a layer of fully connected neurons without non-linear activations and biases. This layer, dubbed the \emph{self-expressive layer}, is added between the encoder and decoder of a standard autoencoder. The parameters (weights) of this layer can be interpreted as the coefficient matrix $C$. The authors promoted the use of convolutional autoencoders as the backbone architecture, reasoning that they can be trained easier compared to the classical fully connected autoencoders as they have less parameters. 
This has been recently proved theoretically in~\cite{li2020convolutional}. 
Let $\Theta$ denote the parameters of the encoder and the  decoder.  
The self-expressiveness in the latent space, that is, $Z_{\Theta} \approx Z_{\Theta}C$, 
is a linear layer located between the encoder and decoder, as illustrated in Figure~\ref{DSC}. In particular, all the input data $X$, as a single large batch, is fed into the encoder to obtain the embedded data $Z_\Theta$, where $Z_\Theta(:,i)$ corresponds to the data point $X(:,i)$. The self-expressive representation of the latent data $Z_\Theta$ is calculated using the subsequent linear layer. The decoder reconstructs each data point in the input batch from the columns of $Z_\Theta C$.
In order to seek a tradeoff between the reconstruction and self-expression errors, DSC-Net solves the following optimization problem: 
\begin{align}
	\min_{\Theta,C} \ &\frac{1}{2} ||X-\tilde{X}_\Theta||_F^2 + \frac{\lambda_1}{2}||Z_{\Theta}-Z_{\Theta}C||_F^2 +\lambda_2 ||C||_p^p, \\
	& \text{such that} \ \diag(C)=0, \nonumber 
\end{align}
where $||\cdot||_p$ is the component-wise $\ell_p$ norm that  regularizes the coefficient matrix $C$ (that is, the weights of the fully connected layer). 
Two different regularizations were considered in \cite{ji2017deep}, namely $p=1,2$, leading to DSC-Net-L1 and DSC-Net-L2 networks, respectively. 
The network is trained in two steps: 
(i)~pretraining which includes initialization of the autoencoder without the self-expressive layer, and 
(ii)~fine-tuning which includes training the whole network including the self-expressive layer. 
Once the network is trained, the weights $C$ of the self-expressive layer are clustered using spectral clustering, as in linear SC.   \\

\paragraph{Recent extensions of self-expressive autoencoders}

DSC-Net initiated many extensions that aim to integrate the representational power of neural networks with self-expressive SC algorithms. 
Among noticeable extensions, there are two approaches that attempt to unify the self-expressive autoencoders with the concept of self-supervision/self-training. 
The Self-Supervised Convolutional Subspace Clustering Network (S$^2$ConvSCN)~\cite{zhang2019self} combines DSC-Net with a classification convolutional module 
(for feature learning) and a spectral clustering module (for self-supervision) into a joint framework.  
This framework uses the (noisy) clustering labels of the spectral clustering module to supervise the training of the feature learning network and the self-expressive coefficient learning. 
The idea of collaborative learning between the classification module and the self-expressive layer was reused in~\cite{zhang2019neural}, but using the concept of confidence learning. In contrast to 
S$^2$ConvSCN, the model in~\cite{zhang2019neural} supervises the training without the use of the computationally expensive spectral clustering. In particular, two affinity matrices are constructed to supervise the training by selecting highly confident sample pairs: (i)~the affinity matrix of a \emph{binary} classifier module based on a convolutional network, which indicates whether two data points belong to the same cluster or not, and (ii)~the affinity matrix corresponding to the weights of the self-expressiveness layer.  Nonetheless, selecting confident samples depends on thresholding parameters which are hard to tune in an unsupervised setting.

Other notable extensions include the following. 
Sparse and low-rank regularized DSC (SLR-DSC)~\cite{zhu2020sparse} regularizes the coefficient matrix using the nuclear and Frobenius norms. An adaptation of DSC-Net for multi-modal data was presented in~\cite{abavisani2018deep}. 
Zhou et al.~\cite{zhou2019latent} added a distribution consistent loss term to keep the consistency between two distributions of the original data and the embedded representation (the output of the encoder). Inspired by human cognition, Jiang et al.~\cite{jiang2018learn} aggregated a self-paced learning framework~\cite{jiang2014self} with the DSC-Net loss function to encourage the network to learn ``easier" samples at first. Easier samples are defined as the ones with a small value of the reconstruction and of the self-expressive representation errors. 
However, similar to many self-paced frameworks, it is very sensitive to thresholding parameters~\cite{graves2017automated}.
Instead of focusing on the learned features from the output of the encoder, Kheirandish et al.~\cite{kheirandishfard2020multi} proposed a multi-level representation of DSC-Net (MLRDSC) which combines low-level information from the initial layers with high-level information from the deeper layers.

\subsubsection{Deep adversarial subspace clustering} \label{adversarial_sc}

So far we have reviewed deep SC approaches that are based on a discriminative supervised neural network (such as those for self-supervised feature learning) and autoencoder architectures. 
There are a few approaches that rely on  
\emph{Generative Adversarial Networks} (GANs)~\cite{goodfellow2014generative}.
GANs are composed of two modules, a \emph{generator} and a \emph{discriminator}. The generator module produces new ``fake'' samples by learning to map from a latent space (often based on random uniform or Gaussian distributions) to the unknown desired distribution of the samples from the given data set. The goal of the generator is to produce samples that follow the true distribution of the data points as closely as possible. On the other hand, the goal of the discriminator is to distinguish the generated fake samples from the ``real'' samples of the data set. These two modules are trained through a minimax game such that the success of each module is the loss of the other one. GANs are often difficult to train and this explains the fewer number of works that combines GANs with SC.

The first SC algorithm that adopted the GAN architecture is Deep Adversarial Subspace Clustering (DASC)~\cite{zhou2018deep}. 
In order to match the goal of SC, DASC modifies the generator and discriminator modules based on well-known linear subspace properties: 
\begin{itemize}
	\item{DASC generator.} 
	The generator has two main components: a self-expressive autoencoder (similar to DSC-Net, with the encoder and decoder network parameters denoted by $\Theta$) and a sampling layer for producing fake data for each estimated subspace. The generator first uses a deep self-expressive auto-encoder to transform the data $X$ into the latent variable $Z_\Theta$ such that they are encouraged to be located in a union of linear subspaces. Thereafter, spectral clustering is applied on the weights from the self-expressive layer ($C$) to obtain the clustering assignment. The generator produces new ``fake" samples $\{\bar{Z}_\Theta^{(i)}\}_{i=1}^c$ by sampling from the estimated subspaces using property of linear subspaces: \emph{linearly combining samples within a 		subspace generates a sample from the same subspace}. To this end, the fake samples are generated by random linear combination of the samples within each cluster (the weights of the combinations are chosen uniformly at random in $[0,1]$). 
	Apart from the fake data generation process, the definition of the ``real" data is also noticeably different from the classic GAN for which the real data is the given data set. In DASC, the real data $\{\tilde{Z}_\Theta^{(i)}\}_{i=1}^c$ are a predefined fraction of the samples within each estimated cluster that have a small projection residual onto the learned subspaces by the discriminator. 
	Mathematically, the generator solves 
	\begin{align} \label{DASC_GEN}
		\min_{C,\Theta}  \ & \underbrace{\frac{1}{c} \sum_{i=1}^c \frac{1}{n_i} \sum_{j=1}^{n_i} \ ||\bar{Z}_\Theta^{(i)}(:,j) - U_iU_i^\top \bar{Z}_\Theta^{(i)}(:,j)||_2^2}_{\text{sum of projection residual of fake data from each subspace}}  \\ 
		& + \underbrace{\lambda_1 ||X - \tilde{X}_\Theta||_F^2}_{\text{reconstruction loss of autoencoder}} + \underbrace{\lambda_2||Z_\Theta-Z_\Theta C||_F^2 + \lambda_3||C||_F^2,}_{\text{LSR based self-expressiveness in embedded space}} \nonumber
	\end{align}
	where $\Theta$ and $\tilde{X}_\Theta$ denote the parameters of the network and the reconstructed data in the autoencoder, respectively, and $n_i$ is the number of fake samples for the $i$th subspace/cluster. The matrices $U_i$ ($i=1,\dots,c$) are the learned bases for each subspace in the discriminator. In fact, the first term in \eqref{DASC_GEN} is the sum of projection residuals of the fake data for all of the estimated subspaces. The generator tries to ``fool" the discriminator by producing (embedded) samples that are very close to the estimated subspaces by the discriminator.
	
	\item{DASC discriminator.} 
	The discriminator is parametrized by the basis vectors of each subspace of the nonlinear transformed data ($Z$). The basis vectors of each subspace/cluster are learned such that the projection residual loss function for the real data is smaller compared to the fake ones. The discriminator solves 
	\begin{align} \label{DASC_DIS}
		\min_{\{U_i\}_{i=1}^c}  \ & \frac{1}{c} \sum_{i=1}^c \frac{1}{n_i} \sum_{j=1}^{n_i} \big(  \underbrace{||\tilde{Z}_\Theta^{(i)}(:,j) - U_iU_i^\top \tilde{Z}_\Theta^{(i)}(:,j)||_2^2}_{\text{ projection residual of \emph{real} data}} + \ \big[ \epsilon - \underbrace{||\bar{Z}_\Theta^{(i)}(:,j) - U_iU_i^\top \bar{Z}_\Theta^{(i)}(:,j)||_2^2}_{\text{projection residual of \emph{fake} data}} \big]_+ \ \big) \\ 
		& \qquad + \underbrace{\lambda_1 \sum_{i=1}^c||U_i^\top U_i-I||_F^2}_{\text{approximately orthonormal bases}} + \underbrace{\lambda_2 \sum_{i\neq j}||U_i^\top U_j||_F^2,}_{\text{separibility of subspaces}}  \nonumber
	\end{align}
	
	where $\big[x\big]_+=\max(x,0)$. The first term in \eqref{DASC_DIS} ensures that the discriminator learns the bases such that they fit the intrinsic subspace of each cluster for the real data. The second term is the adversarial goal of the discriminator compared to the generator. This term ensures that the learned bases of the discriminator produce sufficiently large residuals for the generated fake data. 
\end{itemize} 
Subsequently, the generator and discriminator are trained such that the generator is encouraged to produce fake
data close to the subspaces learned by the discriminator
which leads to higher clustering quality.

However, DASC completely ignores the distribution of the input data $X$ and merely focuses on the latent embedded data $Z$. 
Recently, Yu et al.~\cite{yu2020gan} proposed two extensions to DASC. In the first approach, an additional adversarial learning is utilized to model the distributions of the input data along with the adversarial learning for the corresponding latent representations (similar to DASC). In the second approach, inspired by the self-supervised SC approach in~\cite{zhang2019neural}, the adversarial learning in the latent space is replaced by a self-supervised module to encourage the (autoencoder) network
to learn discriminative embedding (features) for each subspace.

\begin{remark}
	Instead of using GANs to enhance SC, one could also use linear SC to overcome challenges of classical GANs, such as the ``mode-collapse" issue~\cite{bau2019seeing}, that is, the lack of sample variety in the generator's output. 
	For example, Liang et al.~\cite{liang2018sub} suggested to use a clustering module based on linear SC to exploit subspaces within the latent representation of the data. The clustering output is used by the generator to produce samples conditioned on their corresponding subspace to promote diverse sample generation from all latent subspaces. 
\end{remark}

\section{Computational cost of nonlinear SC algorithms} \label{sec:compcost} 

Despite the nice theoretical properties 
of self-expressive representations~\cite{soltanolkotabi2012geometric,elhamifar2013sparse} and their practical efficiency,  
expressing the data points using other data points from the data set can be computationally inefficient. In fact, the computational complexity for computing the coefficient matrix in the major linear SC algorithms is either quadratic or cubic in the number of data points~\cite{pourkamali2020efficient}, and hence using them for medium and large scale data sets is computationally prohibitive. To address the scalability issue, several approaches were proposed which either limit the self-expression over a smaller size dictionary rather than the whole data~\cite{abdolali2019scalable, traganitis2017sketched, chen2020stochastic, you2018scalable} 
or propose specific solvers (such as the greedy OMP~\cite{you2016scalable} or using proximal gradient descent framework~\cite{pourkamali2020efficient}). 

\begin{remark}
	In addition to the high computational complexity of calculating the self-expressive representation (the first step of most nonlinear SC algorithms), the spectral clustering step has in general a computational cost of $O(n^3)$ as well~\cite{huang2019ultra}. However, structures in the affinity matrix such as sparsity can reduce the computational time significantly~\cite{chen2020stochastic}. Moreover, there exists fast approximation algorithms for spectral clustering~\cite{yan2009fast,tremblay2016compressive,huang2019ultra}. Hence, the main focus of scalable extensions of linear SC was to provide faster and more efficient approaches to (approximately) compute the coefficient matrix. 
\end{remark}

The computational burden of self-expressive representations is present in the majority of nonlinear extensions reviewed in this survey as well. Representing data points using the whole data set, whether it is in the ambient or in the (implicit) embedding space, typically has the computational cost of $O(n^3)$ or $O(dn^2)$. Similar to linear SC, ADMM is the standard algorithm used for the majority of the optimization problems. The major bottleneck is computing the $n \times n$ coefficient matrix which often involves solving an $n$-by-$n$ linear system for Kernel based and  locality preserving approaches (in the subcategories of tangent space approximation and avoiding cannot-links). Solving this problem in each iteration of the ADMM framework has a computational complexity of $O(n^3)$, or $O(dn^2)$ using the matrix inversion lemma (also known as Sherman-Morrison-Woodbury identity~\cite{higham2002accuracy}, 
see~\cite[Remark 1]{pourkamali2020efficient} for more details). 
Using the graph Laplacian regularizer, an additional bottleneck of computing a pairwise similarity matrix in the ambient space is added (in $O(dn^2)$ operations), 
and, moreover, the coefficient matrix should be computed by solving a Sylvester equation with the complexity of $O(n^3)$. The high computational cost is also a very evident drawback of the self-expressive based autoencoders in neural network approaches. 
In fact, the number of parameters of the fully-connected self-expressive layer is $\mathcal{O}(n^2)$. Computing the latent representation of the encoder has the time complexity of $O(d_en^2)$ where $d_e$ is the dimension of the encoder output.

In addition to the time complexity, the memory requirement is often significant as well. Storing (and clustering) an $n \times n$ coefficient matrix is usually restrictive, unless the coefficient matrix is sparse. Computing and storing $n \times n$ (dense) Gram matrices increases the space complexity of kernel based approaches further. This is even worse for MKL algorithms which update the kernel matrix at each iteration of an ADMM based algorithm.

\paragraph{Scalable nonlinear SC approaches}
In contrast to linear SC, the scalability issue of nonlinear alternatives has not yet been investigated much.  
To the best of our knowledge, the only scalable algorithm in the category of locality preserving SC is an avoiding cannot-link approach~\cite{han2015locality} which uses $k$ nearby data points to represent each data point, with a computational cost of $O(k^2nd)$ operations. 

With neural networks based SC approaches gaining increasing attention, some approaches tried to tackle the computational bottleneck of DSC-Net, which is due to the self-expressive layer. Zhang et al.~\cite{zhang2018scalable} suggested to replace this layer with a more computationally effective module. Specifically, instead of directly pursuing self-expressive representation in the latent space and constructing an affinity matrix, an iterative linear SC algorithm, namely, the k-subspace clustering (k-SC)~\cite{ho2003clustering} was revisited. The k-SC approach alternates between assigning the data to individual subspaces and estimating the parameters of each subspace. However, as discussed in Section~\ref{intro},
a significant disadvantage of iterative linear SC approaches, including k-SC, is that they require the dimension of each subspace to be known in advance. 
Instead of changing the architecture, Seo et al.~\cite{seo2019deep} proposed an efficient optimization framework for DSC-Net which uses a closed-form solution for deriving the weights of the self-expressive layer using Lagrangian multipliers. However, not only the obtained accuracy is lower than DSC-Net, but also the computational complexity is still quadratic in the number of samples. 

\section{Evaluation and numerical comparison} \label{eval}

In this section, we investigate the properties and performances of major nonlinear SC approaches on synthetic and real world data sets.

\paragraph{Selected Algorithms}   
Among many nonlinear SC algorithms covered in this survey, eight major methods are considered for a detailed comparison and evaluation. The selected algorithms within each category are summarized in Table~\ref{rep_methods} and described as follows: 
\begin{table}[!htbp]
	\caption{Selected representative methods for numerical comparison}\label{rep_methods} \small\addtolength{\tabcolsep}{-1pt}
	\begin{tabularx}{\textwidth}{>{\raggedright}c|X|c|X|c} \hline
		\thead{Category} & \thead{subcategory} & \thead{method} & \thead{characteristic} & \thead{year}  \\ \hline
		& \multirow{2}{*}{graph laplacian regularization} & SMR~\cite{hu2014smooth} & Graph Laplacian regularized self-expressive representation & 2014
		\\ \cline{3-5} {Locality}  
		& & LR$\ell_1$-SSC~\cite{yang2014data} & sparse Graph Laplacian regularization & 2014 \\  \cline{2-5} 
		preserving
		& avoiding cannot-links & KNN-SSC~\cite{zhuang2016locality} & using local self-expressive dictionary & 2016 \\ \cline{2-5}
		& tangent space approximation & SMCE~\cite{elhamifar2011sparse} & tangent approximation using adaptive neighborhood size & 2011 \\ \hline 
		\multirow{2}{*}{Kernel-based} & single kernel & KSSC~\cite{patel2014kernel} & kernelized SSC & 2014 \\ \cline{2-5}
		& multiple kernels & LKG~\cite{kang2019low} & MKL with low-rank consensus kernel & 2019 \\ \hline
		Neural-network & self-expressive latent representation learning & DSC-Net~\cite{ji2017deep} & autoencoder with self-expressive layer & 2017 \\  \cline{2-5} based 
		& deep adversarial subspace clustering & DASC~\cite{zhou2018deep} & combining adversarial learning with LSR & 2018 \\ \hline	
	\end{tabularx}
\end{table}
\begin{itemize}
	\item \textit{Locality preserving based approaches}:
	\begin{itemize}
		
		\item \textbf{Smooth Representation (SMR)}~\cite{hu2014smooth} is in the category of locality preserving SC based on Laplacian regularization. The laplacian matrix in SMR is obtained from a binary K-NN similarity matrix (with parameter $k$).
		
		\item \textbf{Laplacian Regularized $\ell_1$-SSC (LR$\ell_1$-SSC)} \cite{yang2014data}  is similar to SMR but with additional sparse regularization. For a fair comparison, we use the same Laplacian matrix $L$ as SMR.\footnote{
			In \cite{yang2014data}, Gaussian kernels are used for initializing
			the Laplacian matrix $L$. 
			Then, at each iteration of the optimization process, the Laplacian matrix $L$ is updated using the coefficient matrix $C$ from the previous iteration. 
			For a fair comparison, we use the same fixed Laplacian matrix based on binary K-NN, similar to SMR. 
			Moreover, convergence of such a scheme is not guaranteed.  
		}\footnote{
			Optimizing the Laplacian regularized SC problem with additional non-differentiable regularizations (such as the $\ell_1$ or nuclear norm) using ADMM involves solving Sylvester equations~\cite{yankelevskyadmm}. Some approaches in the literature mistakenly derive linear systems instead of Sylvester equations  (this error comes from the derivative of $\tr(CLC^\top)$, authors use $2LC$ instead of $2CL$). Hence, we use our own implementation for the experiments.}

		\item \textbf{k-nearest neighbors based sparse subspace clustering (KNN-SSC)} \cite{zhuang2016locality} 
		is the category of avoiding cannot-links. 
		This approach uses only data points in the k-nearest neighbors of samples as the dictionary for the self-expressive representation. We use the efficient scalable implementation provided in~\cite{tierney2017efficient}.
		
		\item \textbf{Sparse Manifold Clustering and Embedding (SMCE)}~\cite{elhamifar2011sparse} falls in the category of Tangent space approximation. It is one of the early nonlinear extensions of SSC (by the same authors) and the most notable tangent space nonlinear SC approach.
		
	\end{itemize}
	
	\item \textit{Kernel based subspace clustering}: 
	
	\begin{itemize}
		\item \textbf{Kernel Sparse Subspace Clustering (KSSC)} \cite{patel2014kernel} is the pioneer single kernel based nonlinear SC approach based on sparsity regularization. We consider Gaussian (RBF) kernel for the experiments. Note that KSSC with a linear kernel is equivalent to classical linear SSC algorithm~\cite{patel2014kernel}.
		
		\item \textbf{Low-rank Kernel Learning for Graph matrix (LKG)} \cite{kang2019low} is a multiple kernels based nonlinear SC approach.  LKG learns a low-rank kernel from a group of predefined kernels, solving the optimization problem~\eqref{sc_MKL2}; see Section~\ref{multiple_kernel}. We follow the generic setting introduced in \cite{kang2017twin} and consider 12 kernels: 7 Gaussian kernels defined, for $i=1,2,\dots,7$, as  
		\[ 
		K_G^{(i)}(x,y) = 
		\exp^\frac{-||x-y||^2_2}{t_i d_{\max}^2}, 
		\quad t_i \in \{0.01,0.05,0.1,0.5,1,10,50\}
		\] 
		where $d_{\max}$ is the maximal distance between two data points, a linear kernel $K_G^{(8)}(x,y)=x^\top y$, 
		and 4 polynomial kernels of the form 
		$\big(a+x^\top y \big)^b$ for all possible values of $a\in \{0,1\}$ and $b \in \{2,3\}$.  Following the procedure in~\cite{kang2017twin}, all kernel matrices are normalized to have values between 0 and 1. 
		
	\end{itemize}
	\item \textit{Neural networks based subspace clustering}: 
	\begin{itemize}
		\item \textbf{Deep Subspace Clustering Network (DSC-Net)} \cite{ji2017deep} is arguably the most notable neural-network based nonlinear SC algorithm which introduced the self-expressive fully connected layer in the autoencoder architecture. DSC-Net can be used with $\ell_1$ and $\ell_2$ regularizations for the columns of the coefficient matrix, leading to DSC-Net-L1 and DSC-Net-L2 algorithms, respectively. For DSC-Net, we use both the convolutional autoencoder version provided by the authors and a modified \emph{fully connected autoencoder} version (with tanh activation function) implemented by ourselves. The fully connected autoencoder is used for synthetic data sets where the ambient dimension is small.  
		
		\item \textbf{Deep Adversarial Subspace Clustering (DASC)} \cite{zhou2018deep} is the main existing approach in the literature for the combination of self-expressive linear SC with GANs; see Section~\ref{adversarial_sc}.
	\end{itemize}
\end{itemize}

Moreover, three linear SC algorithms are used as baselines: SSC, with both Frobenius norm and $\ell_1$ norm regularization for the error matrix (denoted by SSC-L2 and SSC-L1, respectively), and LRR with $\ell_1$ regularization for the reconstruction error.

The code for SMR\footnote{\url{http://zero-lab-pku.github.io/publication/limingjie/cvpr14_smooth_representation_clustering/}}, SMCE\footnote{\url{http://khoury.neu.edu/home/eelhami/codes.htm}}, KNN-SSC\footnote{\url{http://github.com/sjtrny/kssc}}, DSC-Net\footnote{\url{http://github.com/panji530/Deep-subspace-clustering-networks/}} and DASC\footnote{\url{http://github.com/hyqneuron/dsc_gan}} are available from the author's website. The rest of the algorithms,  implemented by us, are available from \url{https://sites.google.com/site/nicolasgillis/code}. 
The code contains the selected parameters for the algorithms for each data set as well.  The parameters of each algorithm are chosen following the recommendations of the authors.

\paragraph{Quality Measures}  From the many possible metrics to quantify the performances of (subspace) clustering algorithms, two main metrics are commonly used to evaluate the clustering performance: the Accuracy (ACC) and 
the Normalized Mutual Information (NMI)~\cite{strehl2002cluster}. Given the ground-truth labels (clusters) $\ell \in \{1,\dots,c\}^n$ and the obtained labels $\hat{\ell} \in \{1,\dots,c\}^n$ by a clustering algorithm, where $c$ is the number of clusters, these metrics are defined as follows: 
\begin{align*}
	\text{ACC}  = \max_{\pi, \text{ a permutation}}\frac{\sum_{i=1}^{n} \mathbf{1}\big(\ell(i) = 
		\hat{\ell}_\pi(i) \big) }{n},  \quad \text{ and } \quad 
	\text{NMI}  = \frac{I(\ell;\hat{\ell})}{\sqrt{H(\ell)H(\hat{\ell})}},
\end{align*}
where $\mathbf{1}(\cdot)$ is the indicator function which returns one if the input condition is correct, 
$\hat{\ell}_\pi$ is the labelling obtained after permuting the $c$ labels using $\pi$, 
$I( \cdot ; \cdot )$ is the mutual information metric, 
and $H(\cdot)$ is the entropy function. The values of ACC and NMI belong to the interval $[0,1]$; 
a higher value indicates a better clustering performance.

\subsection{Synthetic data} \label{synthetic_num}

The performances of the nonlinear SC approaches are compared on both linear and nonlinear synthetic data sets. We believe that analyzing the properties of nonlinear approaches on carefully designed linear data sets is helpful in understanding the characteristic of these approaches. In fact, linear subspaces are special cases of nonlinear ones, and hence a nonlinear approach should work in linear scenarios. 
For a fair comparison of the algorithms, {no  post-processing} is performed on the coefficient matrices. The obtained coefficient matrices are only symmetrized and then spectral clustering is applied.

\paragraph{Parameters}
For locality preserving approaches, namely, SMR, KNN-SSC and LR$\ell_1$-SSC, the parameter $k$ which indicates the number of nearest neighbors in the similarity matrix construction is set to 10. For the linear synthetic datasets, we did not notice sensitivity to this parameter as long as it is not below 10 (but as this value increases the computational time increases as well). The sensitivity to the parameter $k$ is more noticeable for the nonlinear datasets. We obtained similar results for $k \in [10,15]$, while larger values for $k$ leads to increasingly wrong connections. 
For KSSC, we tuned the parameter $\sigma$ of the RBF kernel from the set \{0.01,0.05,0.1,0.5,1,2,5,10,20\} for the best overall accuracy and set it as 1 for linear synthetic datasets (we obtained similar result for $\sigma = \{0.5,1,2\}$), and as 0.05 
for the nonlinear datasets. The nonlinear datasets are more sensitive to the value of $\sigma$ which is due to the complex nonlinear structures in these datasets. In general KSSC shows significant sensitivity to the parameters which is a common challenge of kernel based approaches.
Moreover, the proper value of $\sigma$ depends on the regularization parameter which balances the trade-off between sparsity and self-expressiveness. Sparser solutions are more sensitive to lower values of $\sigma$ which might lead to over-segmentation. In contrast,  denser solutions are more sensitive to  larger values of $\sigma$ which are prone to dense wrong connections.
For DSC-Net, we consider three hidden layers for the encoder with \{10,8,4\} units for the linear synthetic data sets (we obtained similar results as long as no layer has less than 4 neurons), and \{8,4,2\} units for the nonlinear synthetic data sets. 
Increasing or decreasing the number of hidden layers did not affect the results significantly. 
The encoder in DASC has two hidden layers with \{8,4\} and \{2,2\} units for the synthetic linear and nonlinear data sets, respectively.
For a detailed report on the selected parameters for each approach, see Appendix~\ref{param}.

\subsubsection{Linear subspace clustering} \label{sec:linearresults}

Two principal arrangements of subspaces are considered for the generation of linear synthetic data sets: independent and disjoint. Intuitively, and theoretically~\cite{soltanolkotabi2012geometric}, 
the separation between the subspaces plays a critical role in recovery of the subspaces. Hence, 
for the two types of synthetic data sets, we consider a semi-random data generation approach where the bases of the subspaces are carefully determined with a parameter that controls the affinity between the subspaces. The data generation model, which is inspired by the models in~\cite{soltanolkotabi2012geometric,abdolali2019scalable}, is explained in details in the Appendix~\ref{data_gen}.  
In this section, we compare the performance of nonlinear approaches for both synthetic independent and disjoint subspaces.

\paragraph{First experiment: two independent subspaces}  
Two linearly {independent}  subspaces  with intrinsic dimension two are first considered. 
We set $N=900$ (with 450 points per subspace), $d_i=2 \text{ for }i=1,2$, 
and $d=20$ (the ambient dimension is 20). The angle between the two subspaces is controlled by the parameter $\theta \in \{5,10,20,30,45,60,90\}$. 
For each value of $\theta$, 10 sets of data points are randomly generated to analyze the performance of the tested algorithms. 
Table~\ref{tab_independent} reports the average and standard deviation of the accuracy depending on the  different values of $\theta$. 
\begin{center}
	\begin{table*}[!htbp]
		\begin{center}
			\caption{The average and standard deviation of the accuracy, in percent, of SC approaches on the two independent linear subspaces separated by an angle $\theta$.
			} 
			\label{tab_independent}  
			\small\addtolength{\tabcolsep}{-1pt}
			\begin{tabular}{|c|c||ccccccc|}
				\hline
				\multicolumn{1}{|c}{}& \multicolumn{1}{c}{ $\theta =$ } & 5 & 10 & 20 & 30 & 45 & 60 & 90 \\ \hline \hline 
				\multirow{2}{*}{SMCE} & mean &  51.17 & 50.56 & 98.45 & 99.72 & 99.94 & 99.94 & 99.95  \\ 
				& std & 1.14 & 0.52 & 0.50 & 0.25 & 0.07 & 0.05 & 0.05 \\\hline
				\multirow{2}{*}{SMR} & mean & 100 & 100 & 100 & 100 & 100 & 100 & 100  \\ 
				& std & 0 & 0 & 0 & 0 & 0 & 0 & 0\\\hline
				\multirow{2}{*}{KSSC(RBF)} & mean & 51.25 & 51.85 & 99.84 & 99.96 & 99.97 & 99.97 & 100  \\ 
				& std & 0.71 & 0.99 & 0.13 & 0.05 & 0.04 & 0.04 & 0 \\\hline
				\multirow{2}{*}{LR$\ell_1$-SSC} & mean & 99.88 & 99.97 & 100 & 100 & 100 & 100 & 99.98  \\ 
				& std & 0.09 & 0.04 & 0 & 0 & 0 & 0 & 0.03 \\\hline
				\multirow{2}{*}{LKG} & mean & 51.47 & 51.42 & 51.13 & 50.72 & 99.18 & 99.63 & 99.67 \\ 
				& std & 0.72 & 1.34 & 0.81 & 0.73 & 0.65 & 0.26 & 0.26 \\\hline
				\multirow{2}{*}{KNN-SSC} & mean & 51.60 & 51.44 & 98.26 & 99.35 & 99.85 & 99.86 & 99.82 \\ 
				& std & 0.97 & 0.81 & 0.63 & 0.38 & 0.16 & 0.08 & 0.17 \\\hline
				\multirow{2}{*}{DSC-Net-L2} & mean & 51.41 & 60.65 & 94.14 & 97.56 & 99.44 & 99.46 & 99.33  \\ 
				& std & 0.96 & 14.09 & 5.37 & 1.98 & 0.45 & 0.52 & 0.49 \\\hline
				\multirow{2}{*}{DSC-Net-L1} & mean & 51.33 & 50.84 & 83.82 & 96.11 & 98.97 & 99.51 & 99.57  \\ 
				& std & 1.06 & 0.91 & 8.68 & 6.03 & 0.76 & 0.25 & 0.34 \\\hline
				\multirow{2}{*}{DASC} & mean & 51.70 & 52.22 & 94.15 & 98.95 & 99.93 & 99.85 & 99.91 \\
				& std & 1.19 & 1.40 & 5.62 & 2.57 & 0.09 & 0.16 & 0.12 \\ \hhline{|==#=======|}
				\multirow{2}{*}{SSC-L2} & mean & 99.92 & 99.91 & 99.92 & 99.96 & 99.96 & 99.95 & 99.94 \\
				& std & 0.07 & 0.08 & 0.07 & 0.03 & 0.03 & 0.04 & 0.05\\ \hline
				\multirow{2}{*}{LRR} & mean & 100 & 100 & 100 & 100 & 100 & 100 & 100 \\
				& std & 0 & 0 & 0 & 0 & 0 & 0 & 0\\ \hline
			\end{tabular}
		\end{center}
	\end{table*}
\end{center} 
We observe the following: 
\begin{itemize}
	\item For a sufficiently small affinity between subspaces (more precisely, for $\theta \geq 45$), all of the approaches provide an almost perfect clustering  (namely, the accuracy is above 98.97\% in all cases).	
	
	\item  As guaranteed by the theory~\cite{patel2014kernel,elhamifar2013sparse,hu2014smooth}, SMR, SSC-L2 and LRR perform almost perfectly for all cases of independent subspaces, and their performance is not affected by the affinity between the subspaces. The performance of SSC-L1 was identical to SSC-L2. This is expected as the data set is not contaminated by noise so that the data fitting term does not affect the result. 
	
	\item Interestingly, LR$\ell_1$-SSC, which integrates  Laplacian regularization via the $\ell_1$ norm, also leads to subspace preserving coefficients for all values of $\theta$  although this is not (yet) theoretically supported. 
	
	\item KNN-SSC, which prevents connections between faraway data points, performs poorly for close subspaces (namely, $\theta = 5,10$). 
	However, we noticed that by increasing the number of nearest numbers to more than 100, KNN-SSC leads  to an almost perfect clustering result. 
	
	\item SMCE and KSSC with RBF kernel are producing subspace preserving coefficients for $\theta$ sufficiently large. 
	This is expected, because SMCE relies on the proximity of data points for the tangent space approximation, and KSSC with RBF kernel is dependent on the nearby samples which is controlled by the parameter $\sigma$. Hence, the closer the subspaces get, the higher the chance of wrong clusterings.

	\item The accuracy of DSC-Net increases with $\theta$.  Interestingly, DSC-Net based on $\ell_1$ regularization perform worse than with $\ell_2$. 
	The authors in \cite{ji2017deep} argued that this happens in practice because $\ell_1$ is non-differentiable at zero. However, we do believe this is related to optimizing the network weights (including the coefficient matrix) using the default subgradient descent method of neural networks.  
	In practice, they usually produce non-sparse solutions compared to other optimization methods such as proximal methods  ~\cite{bach2011optimization}. Generating non-sparse solutions is due to slow convergence. Although spectral clustering is expected to be robust to weak wrong connections, there is no guarantee that it is robust to the intermediate non-sparse solutions; in particular if the solution contains many small non-zero entries with no dominant large entries.

	\item LKG has the worst performance, providing high accuracy only for $\theta \geq 45$. 
	This shows that multiple kernel learning might lead to worse performance compared to the single kernel based alternatives. 
\end{itemize}

\paragraph{Second experiment: three disjoint subspaces}   
In the second noiseless experiment, samples are generated from three linearly {disjoint} subspaces. In particular, we set $N=900$ (with 300 data points per subspace), $d_i=2 \text{ for }i=1,2,3$, $d=20$ and $\theta \in \{5,10,20,30,45,60\}$ (note that we do not consider $\theta=90^{\circ}$ because it leads to identical subspaces for the first and second basis); see Appendix~\ref{data_gen} for more details on the data generation model. 
The average and standard deviation of accuracy over 10 trials for 10 randomly generated subspaces are reported in Table~\ref{tab_disoint}.

\begin{center}
	\begin{table*}[!htbp]
		\begin{center}
			\caption{The average and standard deviation of the accuracy, in percent, of SC algorithms on three disjoint linear subspaces whose  affinity is measured by the angle $\theta$. 			
				\label{tab_disoint}} 
			\label{tab_disjoint}  
			\small\addtolength{\tabcolsep}{-1pt}
			\begin{tabular}{|c|c||cccccc|}
				\hline
				\multicolumn{1}{|c}{}& \multicolumn{1}{c}{$\theta = $} & 5 & 10 & 20 & 30 & 45 & 60  \\ \hline \hline 
				\multirow{2}{*}{SMCE} & mean & 35.53 & 35.54 & 63.81 & 99.25 & 99.73 & 99.92  \\
				& std & 0.56 & 0.74 & 9.73 & 0.35 & 0.17 & 0.07 \\ \hline
				\multirow{2}{*}{SMR} &mean & 67.61 & 67.22 & 66.97 & 66.93 & 67.70 & 35.70  \\ 
				& std & 0.54 & 2.96 & 2.81 & 3.10 & 1.11 & 1.02 \\\hline
				\multirow{2}{*}{KSSC(RBF)} & mean & 35.42 & 35.74 & 54.82 & 99.47 & 99.78 & 99.95   \\ 
				& std & 0.57 & 0.93 & 10.18 & 0.28 & 0.20 & 0.07 \\\hline
				\multirow{2}{*}{LR$\ell_1$-SSC} & mean & 62.93 & 80.72 & 84.76 & 92.94 & 99.96 & 99.98  \\
				& std & 11.65 & 11.06 & 13.49 & 11.51 & 0.05 & 0 \\ \hline
				\multirow{2}{*}{LKG} & mean & 35.72 & 35.30 & 35.86 & 37.23 & 97.50 & 99.03   \\ 
				& std & 0.94 & 0.70 & 1.02 & 1.52 & 0.90 & 0.68 \\\hline
				\multirow{2}{*}{KNN-SSC} & mean & 35.74 & 35.86 & 42.13 & 98.85 & 99.72 & 99.76   \\ 
				& std & 0.71 & 1.15 & 7.02 & 0.59 & 0.13 & 0.09 \\\hline
				\multirow{2}{*}{DSC-Net-L2} & mean & 39.50&54.58 & 62.88 & 64.73 & 60.93 & 38.35   \\ 
				& std & 6.60 & 6.05 & 3.62 & 1.85 & 4.65 & 2.25 \\\hline
				\multirow{2}{*}{DSC-Net-L1} & mean & 35.18 & 40.46 & 57.93 & 61.93 & 97.91 & 98.55   \\ 
				& std & 0.72 & 5.53 & 6.02 & 7.07 & 1.34 & 0.82 \\\hline
				\multirow{2}{*}{DASC} & mean & 35.94 & 59.42 & 65.45 & 65.61& 67.44& 37.03  \\ 
				& std & 0.87 & 5.06  & 4.28  & 3.91 & 1.20 & 1.45 \\ \hhline{|==#======|}
				\multirow{2}{*}{SSC-L2} & mean & 60.95 & 83.15 &  82.77 & 92.84 & 99.76 & 99.84  \\ 
				& std & 11.12 & 10.40 & 13.94 & 11.41 & 0.18 & 0.13 \\\hline
				LRR & mean & 67.63 & 67.22 & 66.97 & 66.92 & 67.70 & 35.75 \\ 
				& std & 0.55 & 2.96 & 2.81 & 3.09 & 1.11 & 1.07 \\\hline
			\end{tabular}
		\end{center}
	\end{table*}
\end{center}

We observe the following:
\begin{itemize}
	
	\item Compared to the independent case, the accuracy of all nonlinear SC approaches decreases significantly for $\theta$ sufficiently small (high affinity between subspaces).  
	However, sparsity based approaches, namely, SMCE, KSSC, LR$\ell_1$-SSC, KNN-SSC and DSC-Net-L1 have the best performances, especially for $\theta$ sufficiently large. This is inline with theoretical justifications for sparse based linear SC algorithms, that is, SSC~\cite{soltanolkotabi2012geometric} and $\ell_0$-SSC~\cite{yang2016ell}. 
	
	\item As opposed to the first experiment with independent subspaces, the performance of SMR is among the worse for  disjoint subspaces. 
	This is expected because SMR only considers the \emph{grouping effect} with no other regularization. 
	Similar performance is observed for DSC-Net-L2 and DASC which are based on $\ell_2$ regularization. This highlights the vital role of sparsity in dealing with disjoint subspaces.

	\item Interestingly, KSSC with RBF kernel, KNN-SSC and SMCE perform slightly better than SSC-L2 and LR$\ell_1$-SSC for $\theta=30^{\circ}$. 
	This suggests that nonlinearity might lead to better representation learning when subspaces are close (but with sufficient separability) as shown empirically in \cite{patel2014kernel} for KSSC.
	
\end{itemize}

\begin{figure*}[!htbp]
	\begin{minipage}[b]{0.3\linewidth}
		\centering
		\centerline{\includegraphics[width=6.5cm]{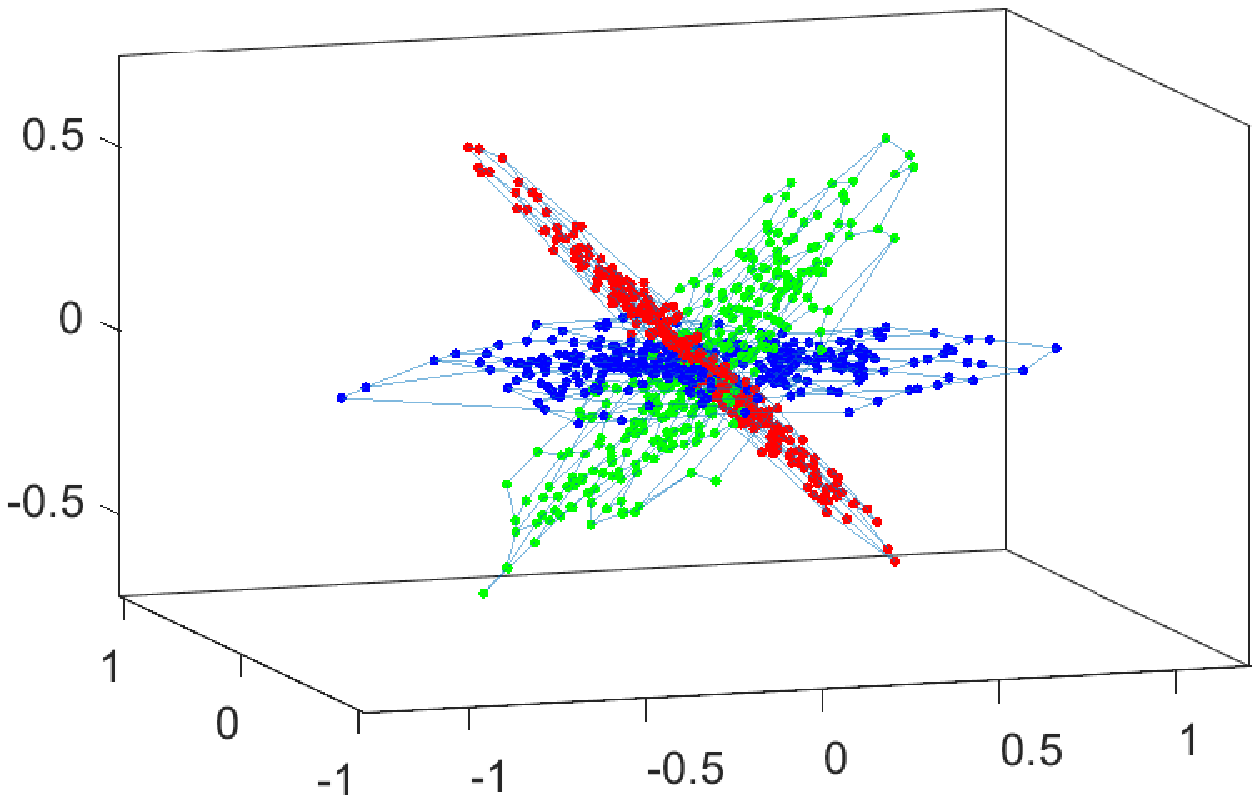}}
		\centerline{(a) SMCE (99.78\%)}\medskip
	\end{minipage}
	\hfill
	\begin{minipage}[b]{0.3\linewidth}
		\centering
		\centerline{\includegraphics[width=6.5cm]{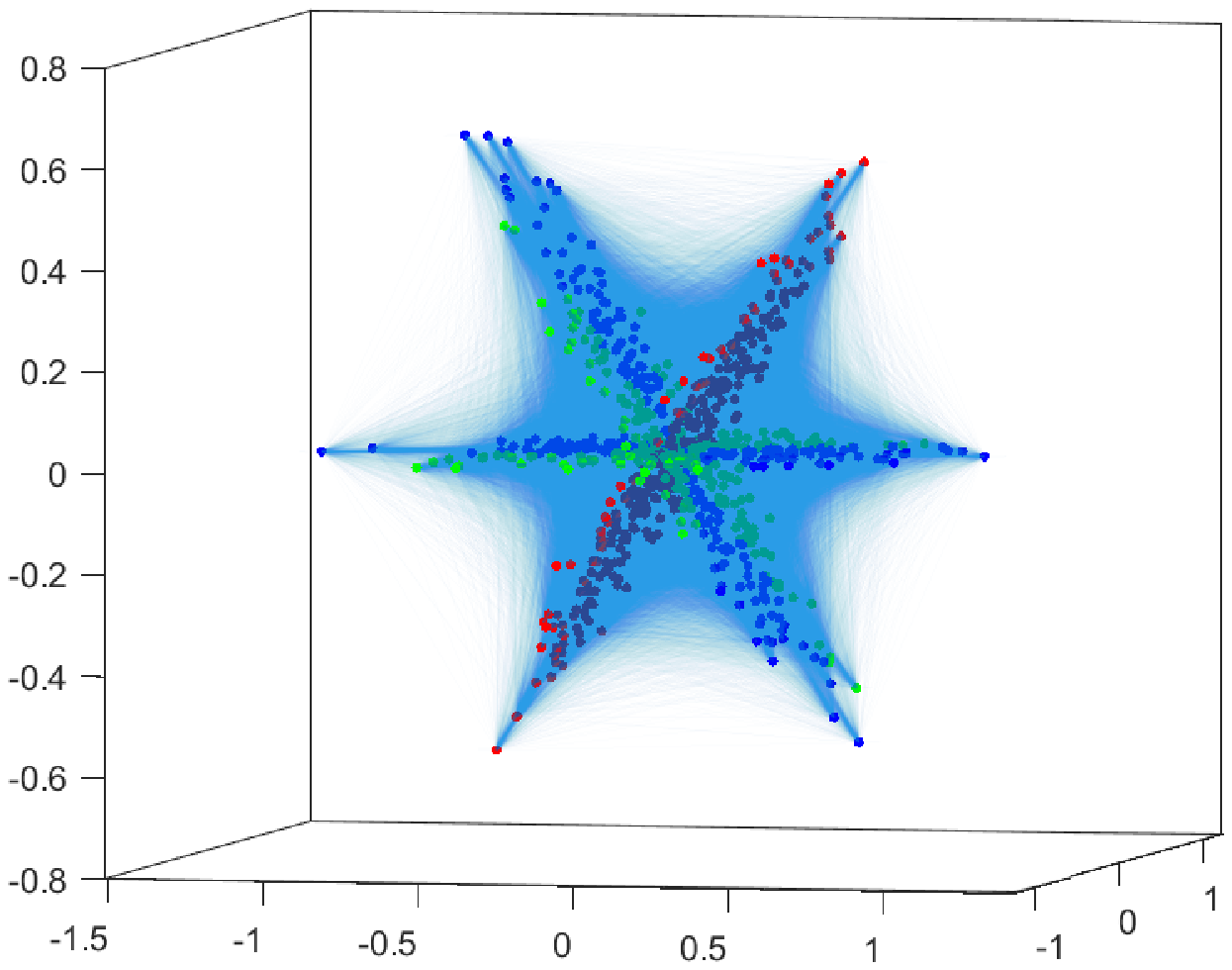}}
		\centerline{(b) SMR (66.67\%)}\medskip
	\end{minipage}
	\hfill
	\begin{minipage}[b]{0.3\linewidth}
		\centering
		\centerline{\includegraphics[width=6.5cm]{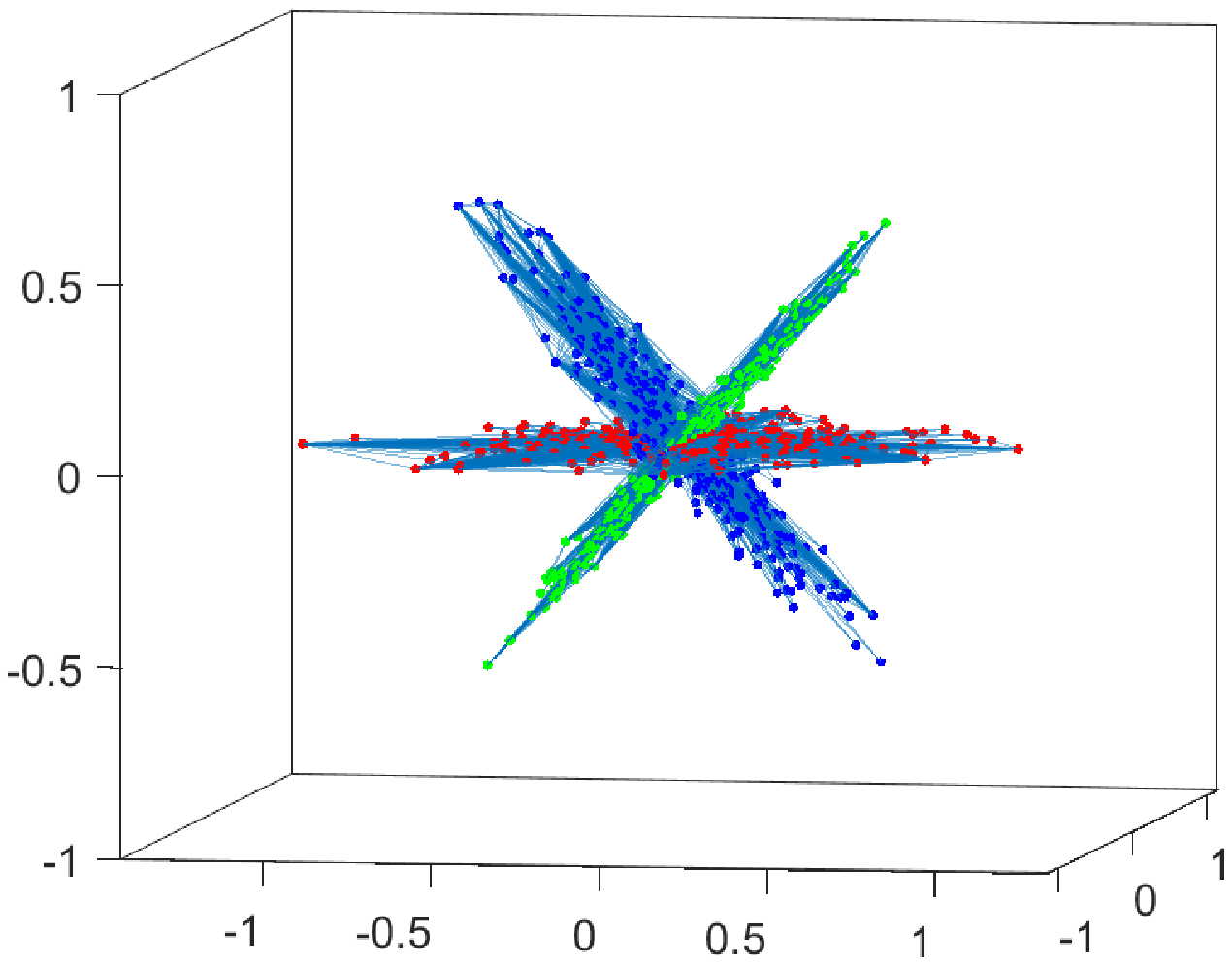}}
		\centerline{(c) LR$\ell_1$-SSC (99.89\%)}\medskip
	\end{minipage}
	\hfill
	\begin{minipage}[b]{0.3\linewidth}
		\centering
		\centerline{\includegraphics[width=6.5cm]{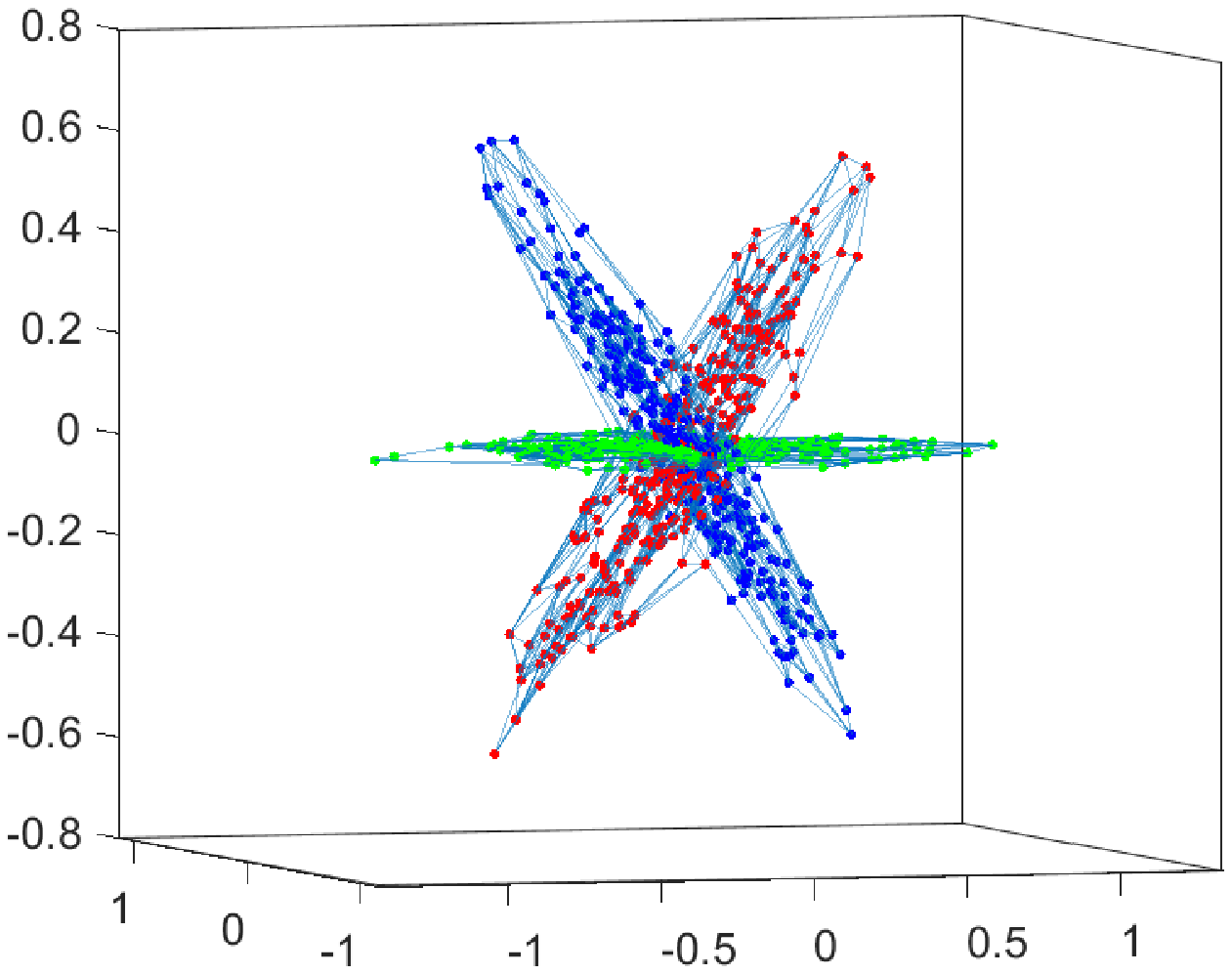}}
		\centerline{(d) KNN-SSC (99.78\%)}\medskip
	\end{minipage}
	\hfill
	\begin{minipage}[b]{0.3\linewidth}
		\centering
		\centerline{\includegraphics[width=6.5cm]{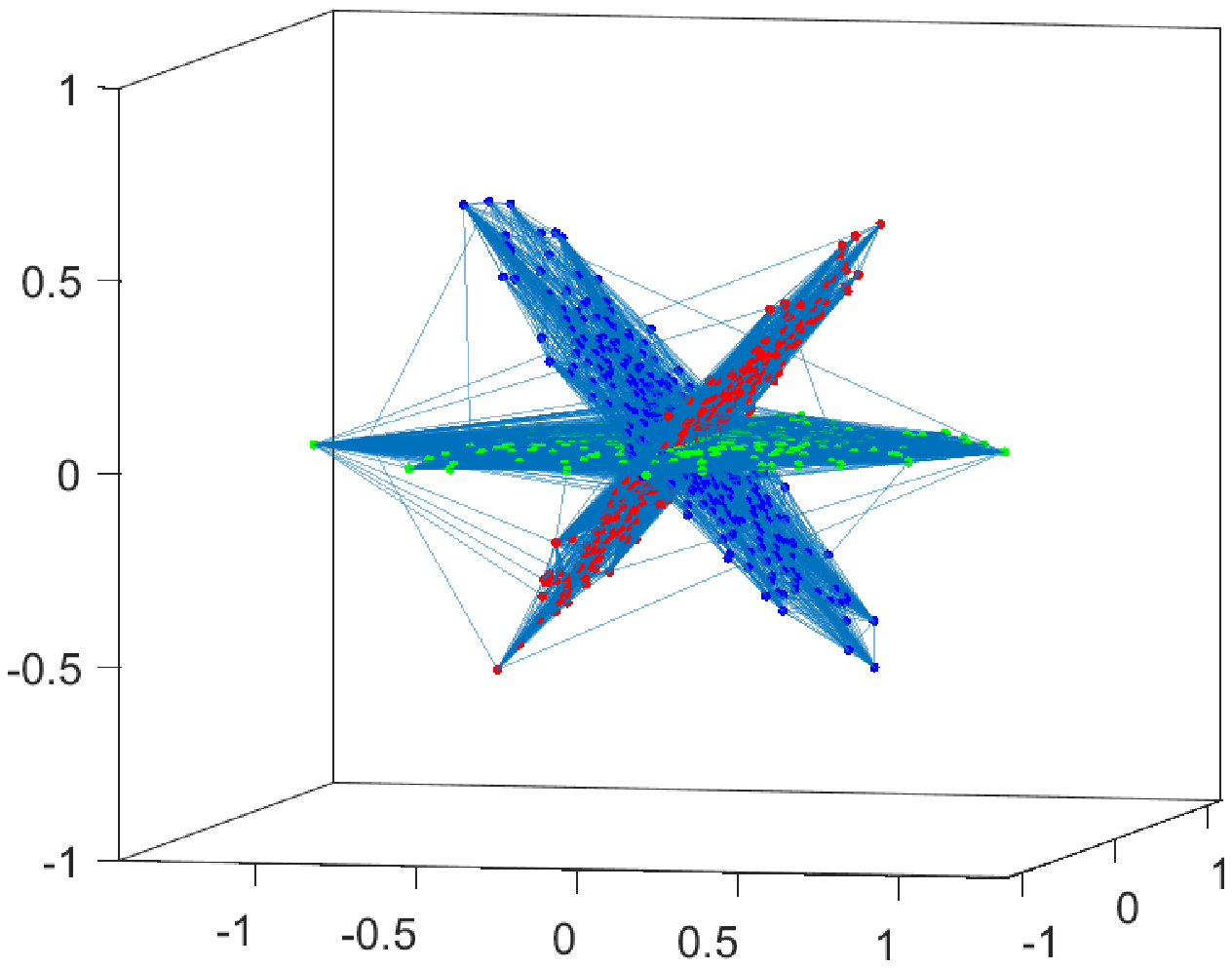}}
		\centerline{(e) KSSC (RBF kernel) (100\%)}\medskip
	\end{minipage}
	\hfill
	\begin{minipage}[b]{0.3\linewidth}
		\centering
		\centerline{\includegraphics[width=6.5cm]{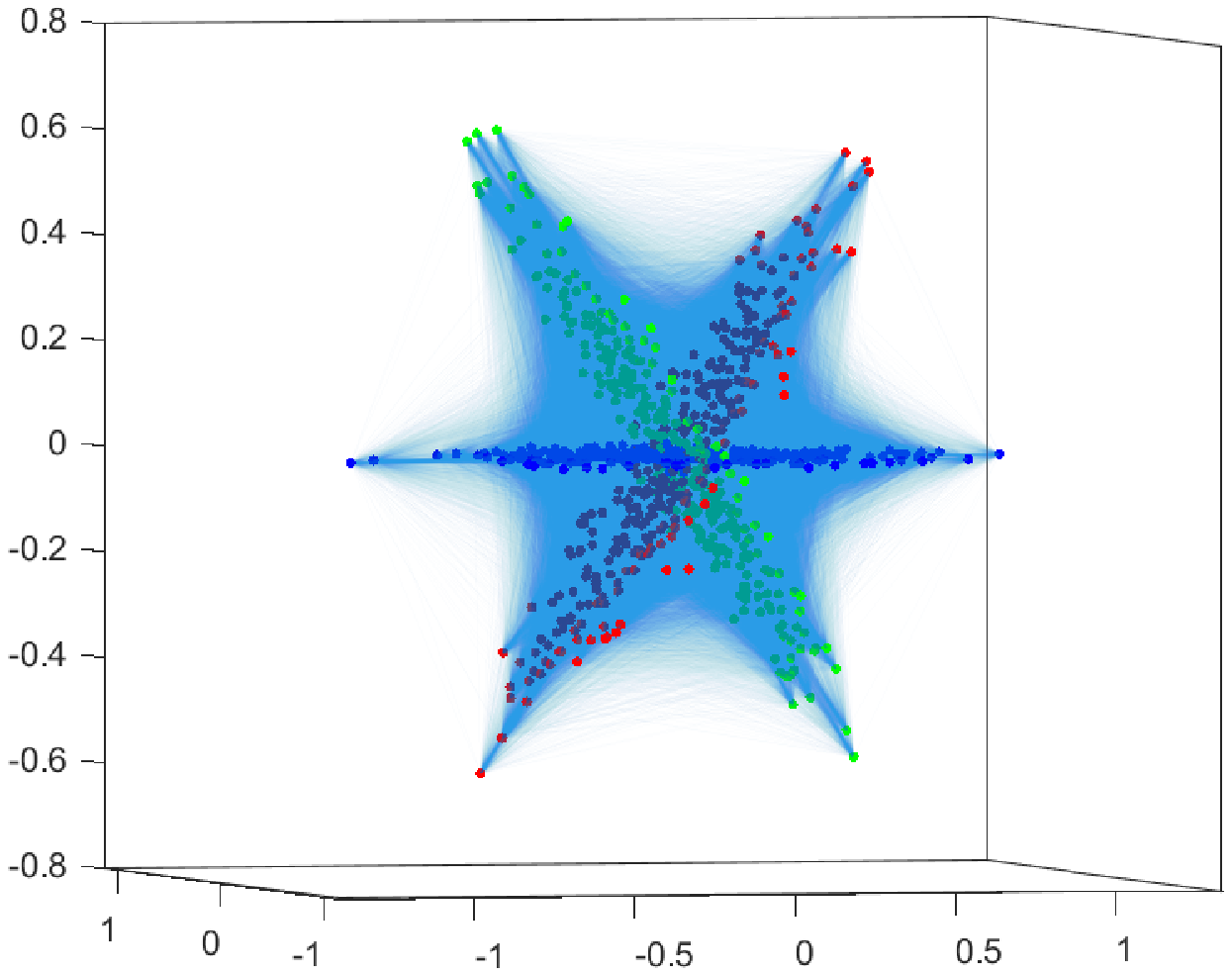}}
		\centerline{(f) LKG (99.22\%)}\medskip
	\end{minipage}
	\hfill
	\begin{minipage}[b]{0.3\linewidth}
		\centering
		\centerline{\includegraphics[width=6.5cm]{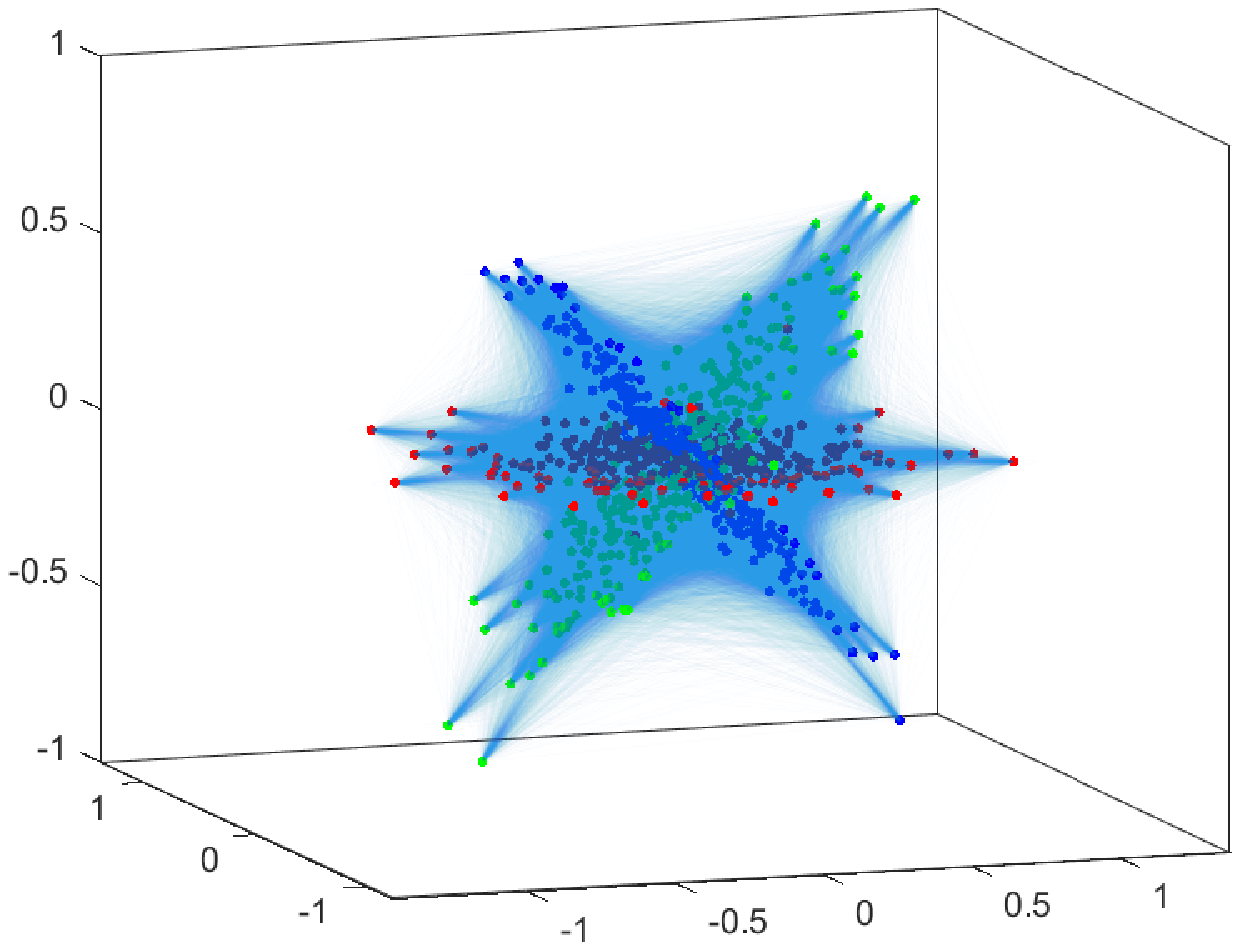}}
		\centerline{(g) DSC-Net $\ell_1$ (99.56\%)}\medskip
	\end{minipage}
	\hfill
	\begin{minipage}[b]{0.3\linewidth}
		\centering
		\centerline{\includegraphics[width=6.5cm]{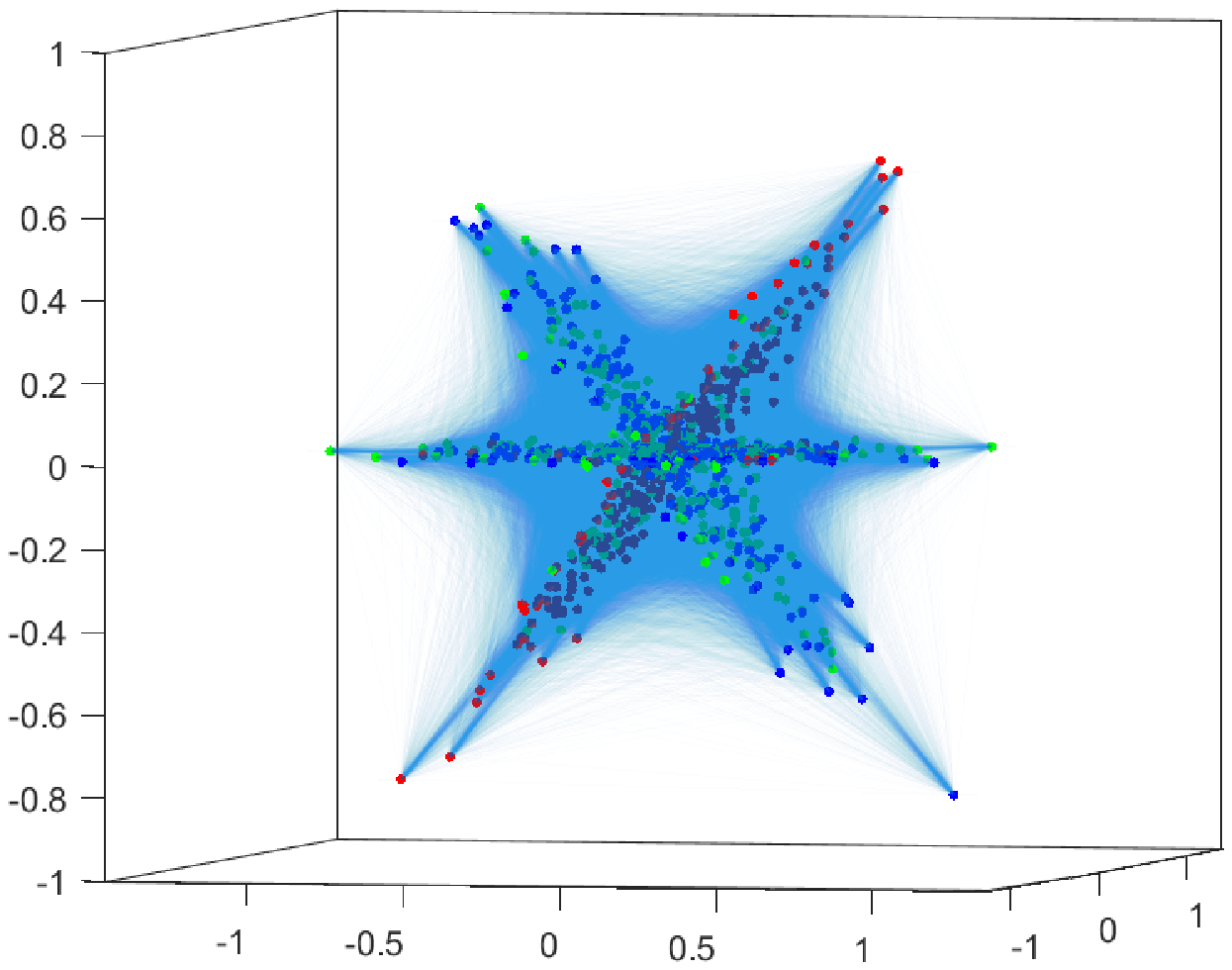}}
		\centerline{(h) DSC-Net $\ell_2$ (60.89\%)}\medskip
	\end{minipage}
	\hfill
	\begin{minipage}[b]{0.3\linewidth}
		\centering
		\centerline{\includegraphics[width=6.5cm]{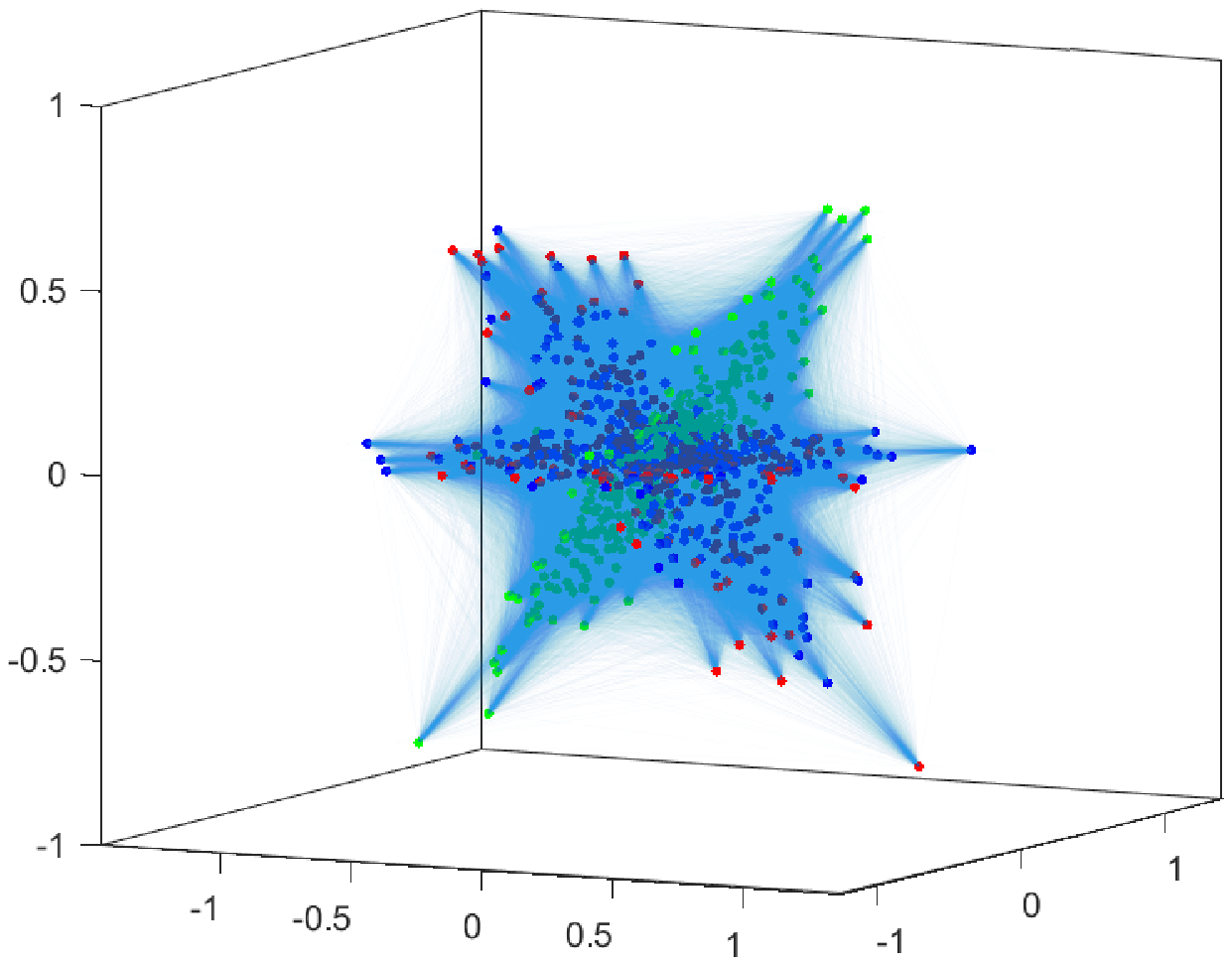}}
		\centerline{(i) DASC (66.70\%)}\medskip
	\end{minipage}
	\caption{The connectivity graphs of nonlinear SC algorithms on three disjoint subspaces with $\theta=45^{\circ}$. The accuracy of the displayed instances is indicated in brackets. The average accuracy over the 10 randomly generated instances can be found in Table~\ref{tab_disoint}.
	}
	\label{disjoint}
\end{figure*}

Figure~\ref{disjoint} displays the connectivity graphs of the three disjoint subspaces for $\theta=45^{\circ}$, for the 9 tested algorithms. 
The data points for three clusters are shown in red, blue and green. The first three dimensions are considered for these plots. Note the dense and many wrong connections for SMR, DSC-Net-L2 and DASC. 
There are many wrong weak connections in the graph of DSC-Net-L1 too, however, the stronger connections within subspaces overcome the weaker wrong ones, leading to an almost correct clustering result (namely, 97.91\%; 
see Table~\ref{tab_disjoint}). 
Non-sparse solution of DSC-Net-L1 is expected as subgradient descent based methods usually do not lead to sparse solutions in practice~\cite{bach2011optimization}.

\paragraph{Conclusion for the linear subspace clustering experiments} 

The majority of nonlinear SC approaches have difficulties in segmenting the subspaces that are close to each other. This is specifically evident for the cases with $\theta=5,10$. 
This is expected since the majority of non-linear approaches have the implicit assumption that the data points from different clusters are not spatially close to each other. This behavior is expected to be present in clustering nonlinear subspaces as well, as we will see in the next experiments.

\subsubsection{Nonlinear subspace clustering}

\begin{figure*}[!htbp]
	\begin{minipage}[b]{0.24\linewidth}
		\centering
		\centerline{\includegraphics[width=5cm]{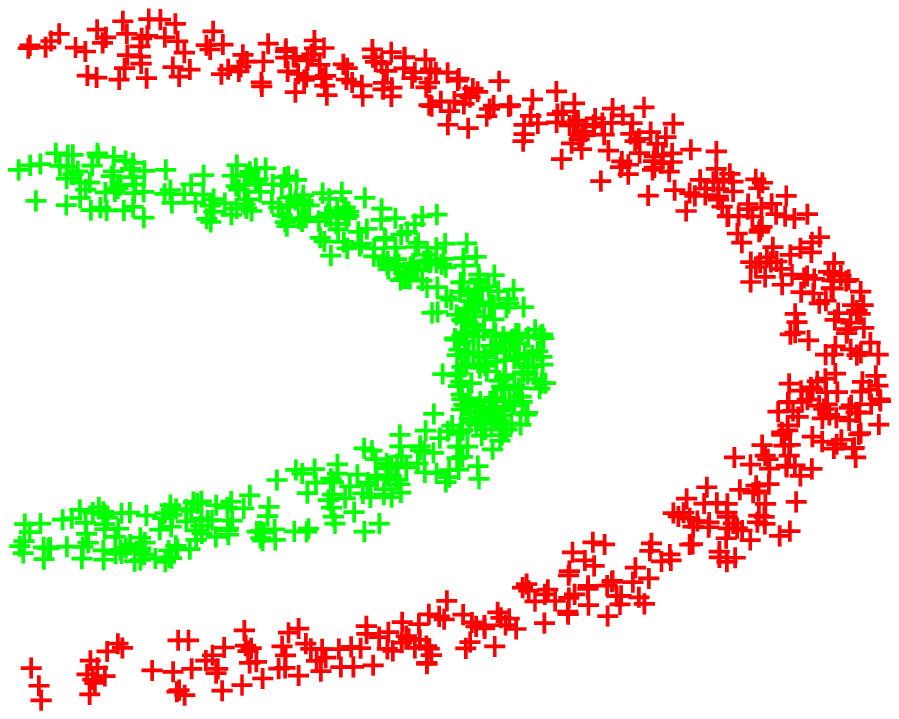}}
		\centerline{(a) ground-truth}\medskip
	\end{minipage}
	\hfill
	\begin{minipage}[b]{0.24\linewidth}
		\centering
		\centerline{\includegraphics[width=5cm]{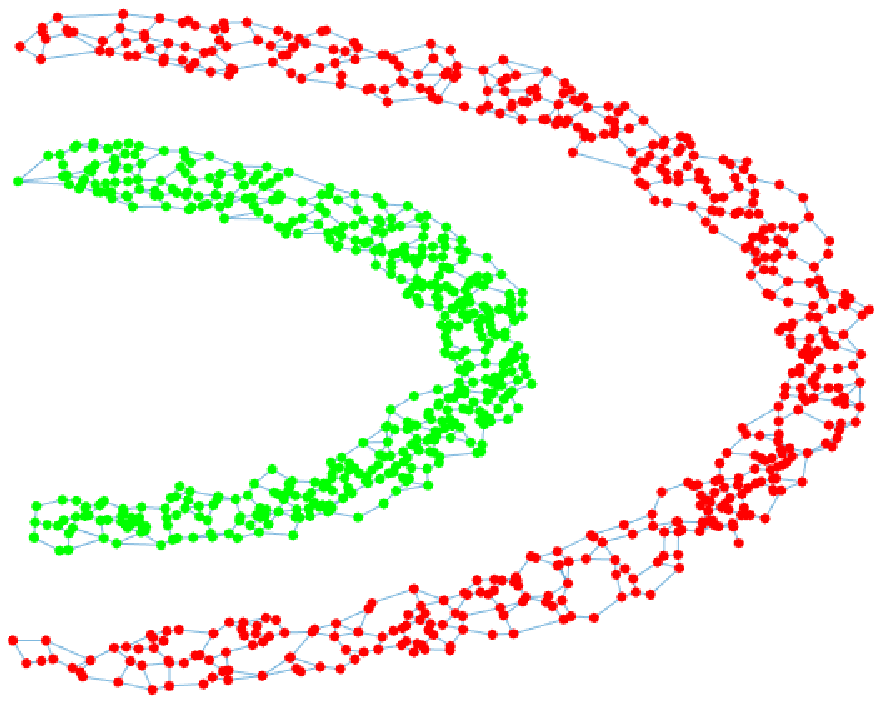}}
		\centerline{(b) SMCE (100\%)}\medskip
	\end{minipage}
	\hfill
	\begin{minipage}[b]{0.24\linewidth}
		\centering
		\centerline{\includegraphics[width=5cm]{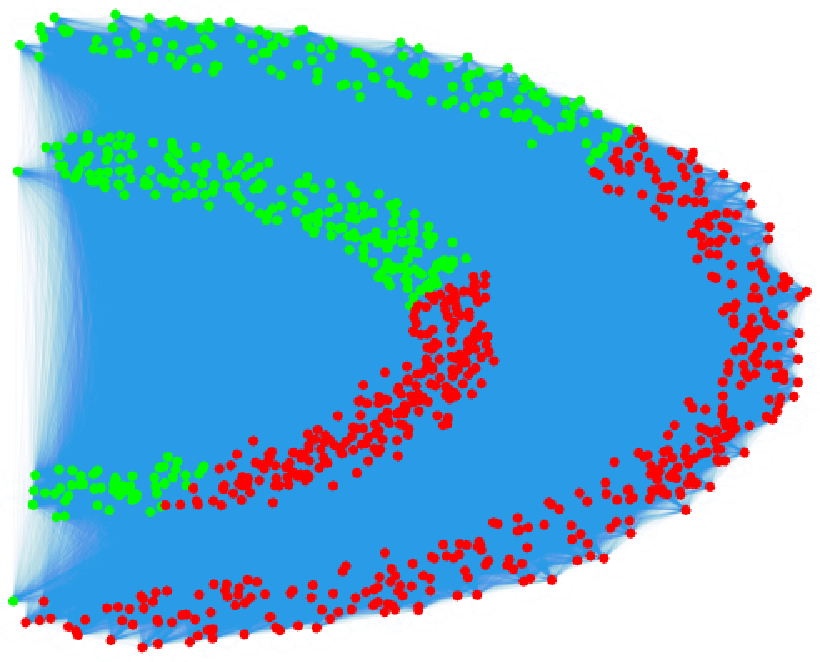}}
		\centerline{(c) SMR (61.80\%)}\medskip
	\end{minipage}
	\hfill
	\begin{minipage}[b]{0.24\linewidth}
		\centering
		\centerline{\includegraphics[width=5cm]{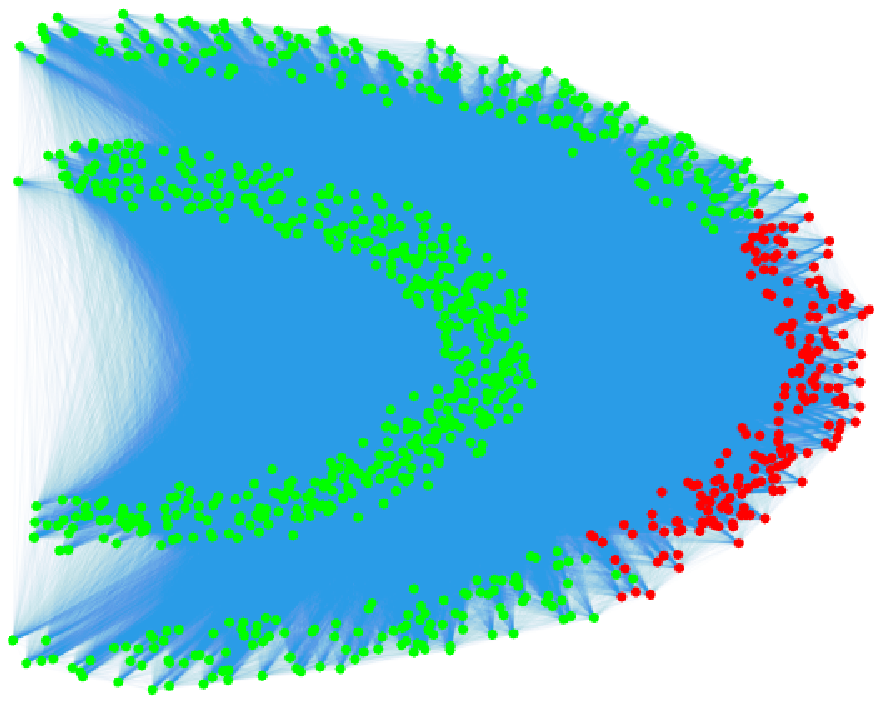}}
		\centerline{(d) LR$\ell_1$-SSC (68.20\%)}\medskip
	\end{minipage}
	\hfill
	\begin{minipage}[b]{0.3\linewidth}
		\centering
		\centerline{\includegraphics[width=5cm]{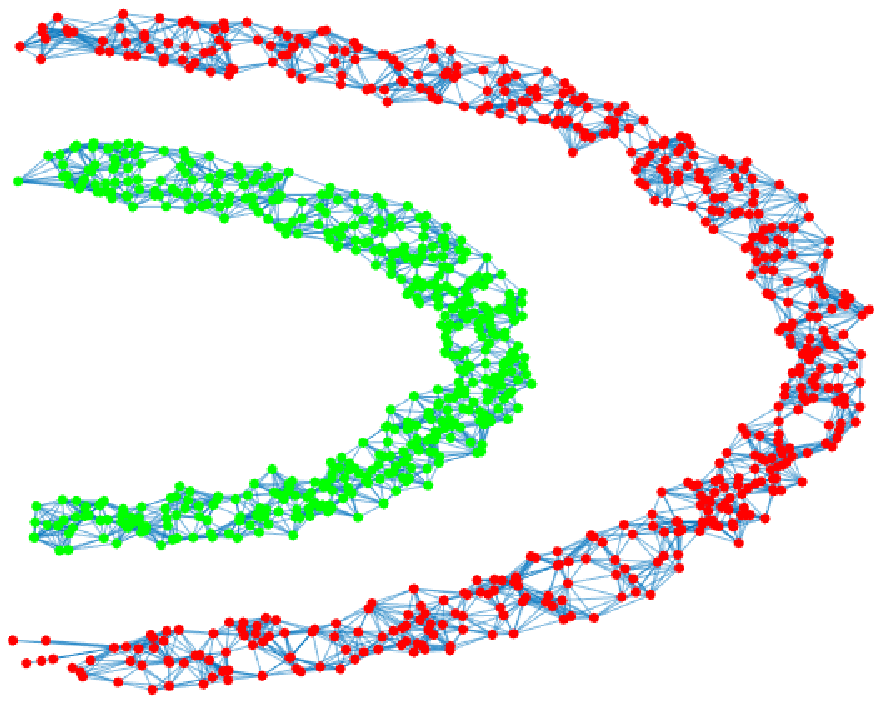}}
		\centerline{(e) KNN-SSC (100\%)}\medskip
	\end{minipage}
	\hfill
	\begin{minipage}[b]{0.3\linewidth}
		\centering
		\centerline{\includegraphics[width=5cm]{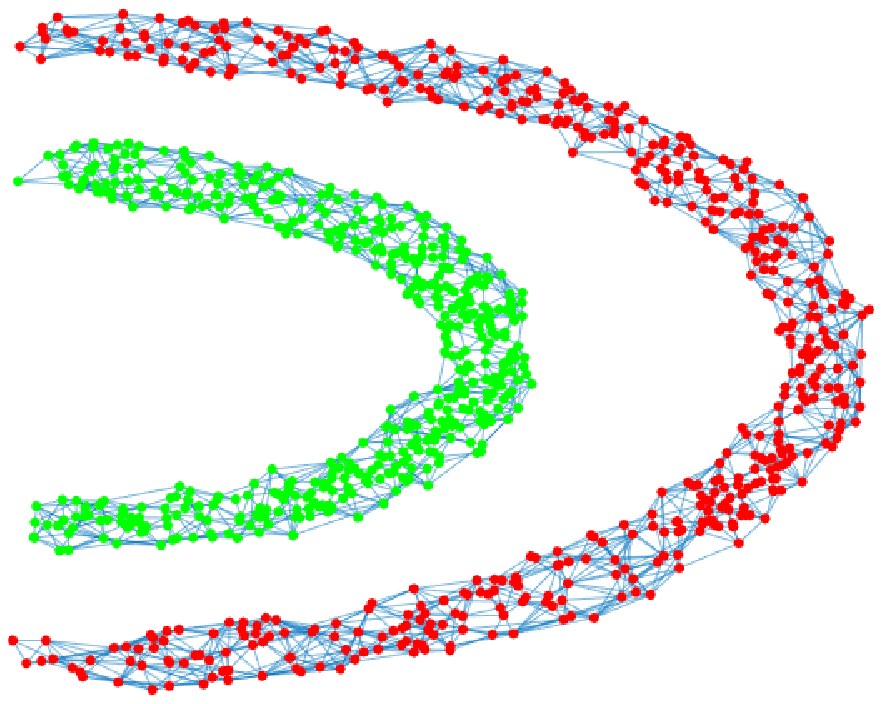}}
		\centerline{(f) KSSC (RBF) (100\%)}\medskip
	\end{minipage}
	\hfill
	\begin{minipage}[b]{0.3\linewidth}
		\centering
		\centerline{\includegraphics[width=5cm]{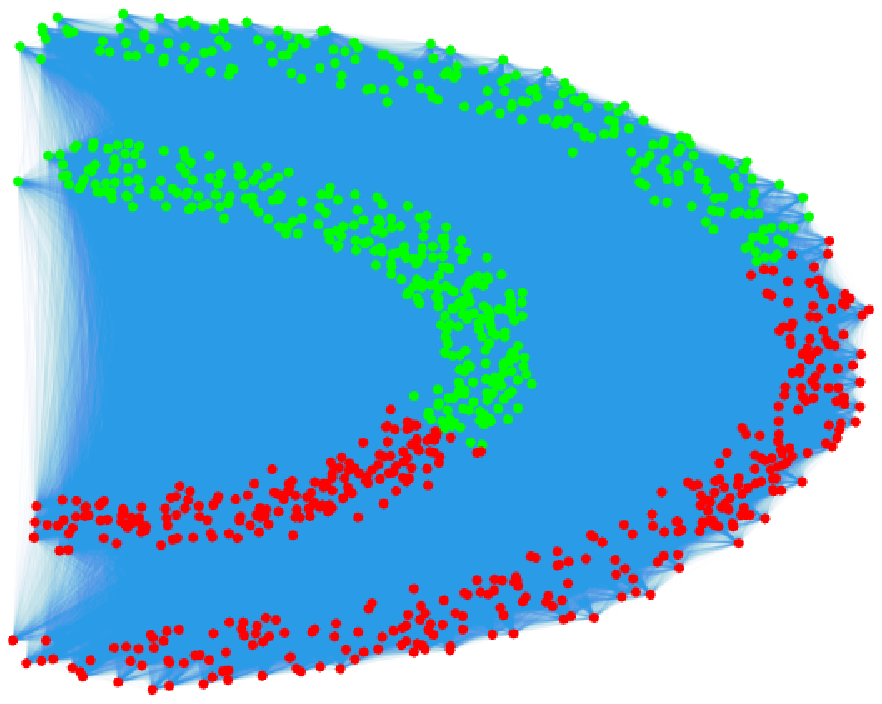}}
		\centerline{(g) LKG (61.50\%)}\medskip
	\end{minipage}
	\hfill
	\caption{
		Comparing the performance of nonlinear SC approaches on \ngi{a} half kernels synthetic data set. The accuracy of the displayed instances is indicated in brackets. The average accuracy over the 10 randomly generated instances can be found in Table~\ref{nonlinear_avg}.
	} 
	\label{halfkernel}
\end{figure*}

\begin{figure*}[!htbp]
	\begin{minipage}[b]{0.24\linewidth}
		\centering
		\centerline{\includegraphics[width=4.5cm]{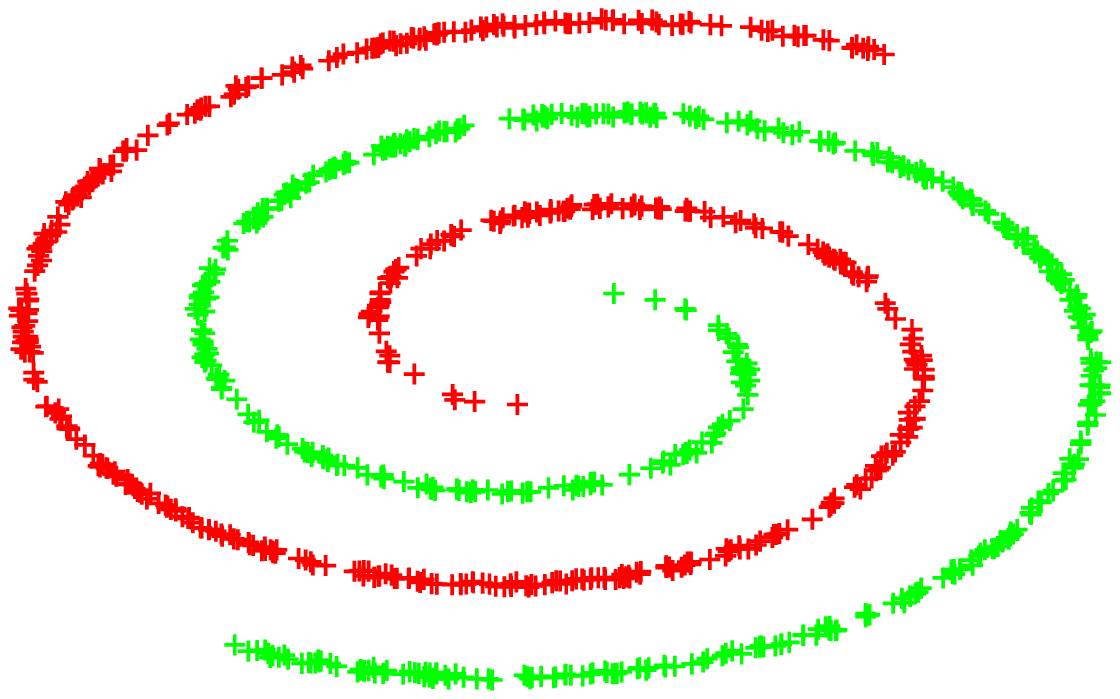}}
		\centerline{(a) ground-truth}\medskip
	\end{minipage}
	\hfill
	\begin{minipage}[b]{0.24\linewidth}
		\centering
		\centerline{\includegraphics[width=4.5cm]{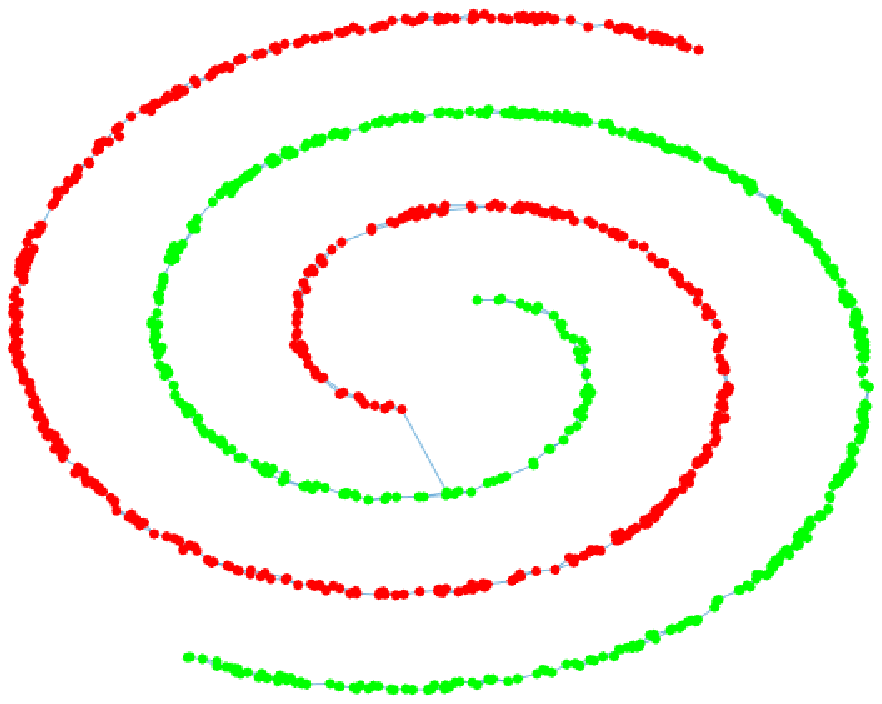}}
		\centerline{(b) SMCE (100.00\%)}\medskip
	\end{minipage}
	\hfill
	\begin{minipage}[b]{0.24\linewidth}
		\centering
		\centerline{\includegraphics[width=4.5cm]{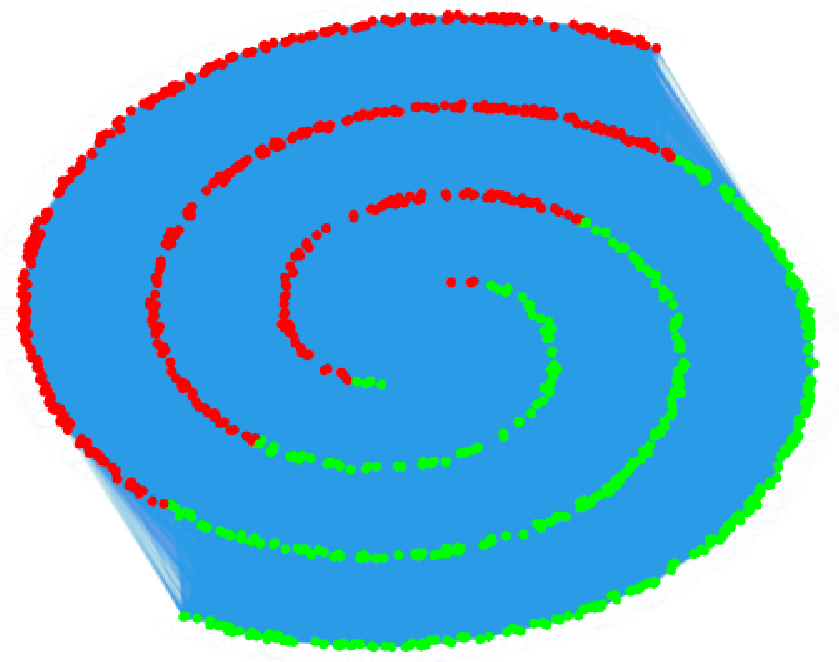}}
		\centerline{(c) SMR (63.90\%)}\medskip
	\end{minipage}
	\hfill
	\begin{minipage}[b]{0.24\linewidth}
		\centering
		\centerline{\includegraphics[width=4.5cm]{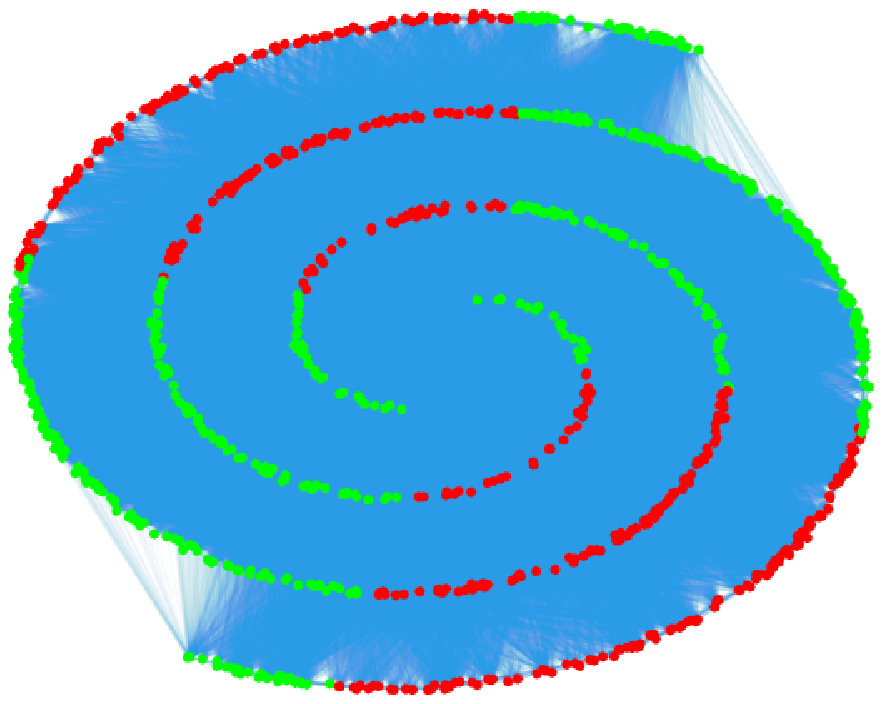}}
		\centerline{(d) LR$\ell_1$-SSC (51.00\%)}\medskip
	\end{minipage}
	\hfill
	\begin{minipage}[b]{0.3\linewidth}
		\centering
		\centerline{\includegraphics[width=5cm]{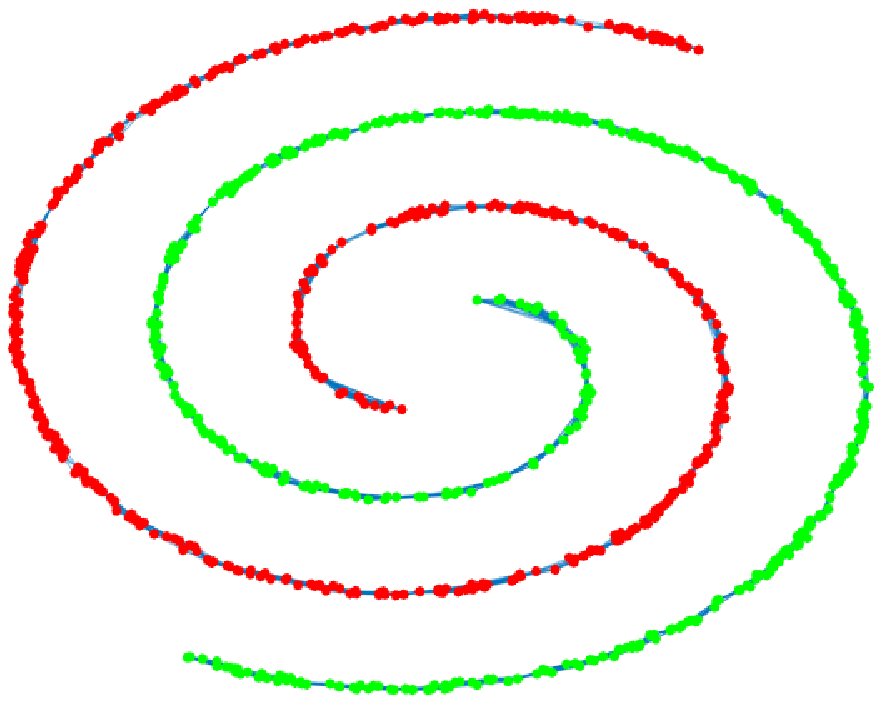}}
		\centerline{(e) KNN-SSC (100\%)}\medskip
	\end{minipage}
	\hfill
	\begin{minipage}[b]{0.3\linewidth}
		\centering
		\centerline{\includegraphics[width=5cm]{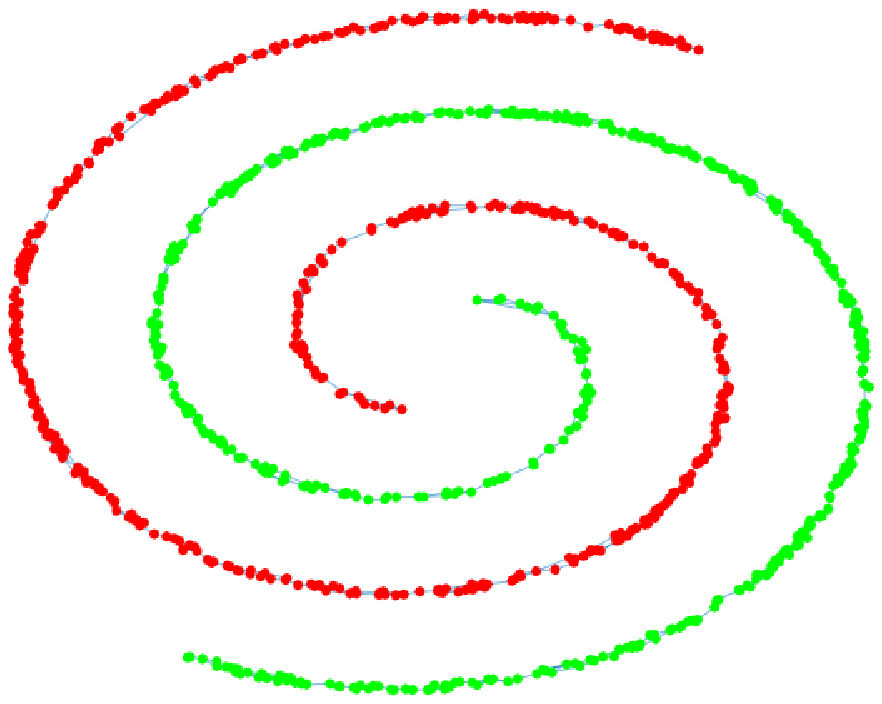}}
		\centerline{(f) KSSC (RBF) (100\%)}\medskip
	\end{minipage}
	\hfill
	\begin{minipage}[b]{0.3\linewidth}
		\centering
		\centerline{\includegraphics[width=5cm]{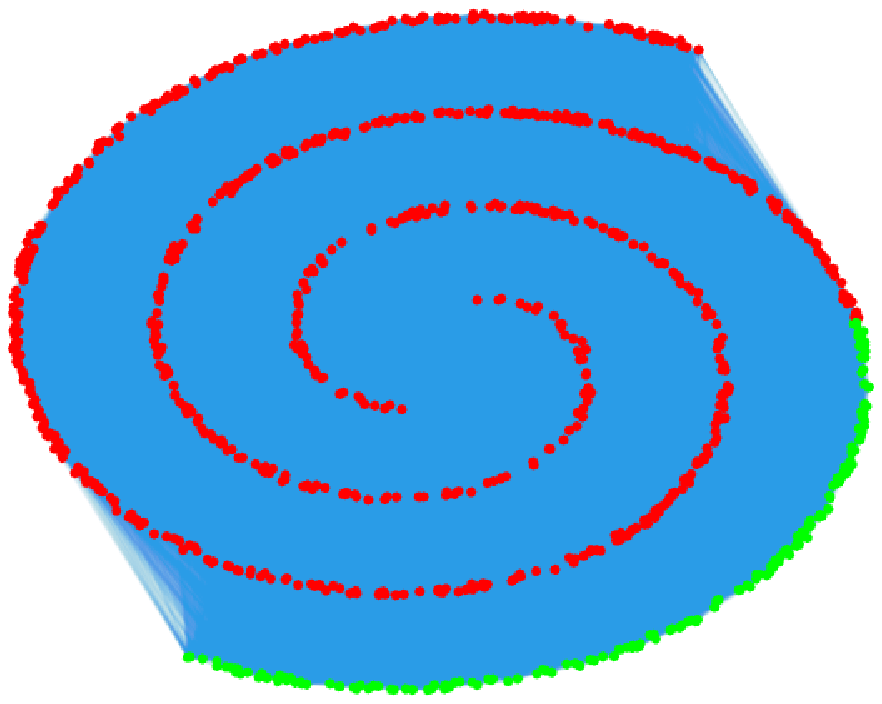}}
		\centerline{(g) LKG (68.50\%)}\medskip
	\end{minipage}
	\hfill
	\caption{Comparing the performance of nonlinear SC approaches on a two spirals synthetic data set. The accuracy of the displayed instances is indicated in brackets. The average accuracy over the 10 randomly generated instances can be found in Table~\ref{nonlinear_avg}.
	} 
	\label{spirals}
\end{figure*}

\begin{figure*}[!htbp]
	\begin{minipage}[b]{0.24\linewidth}
		\centering
		\centerline{\includegraphics[width=4.5cm]{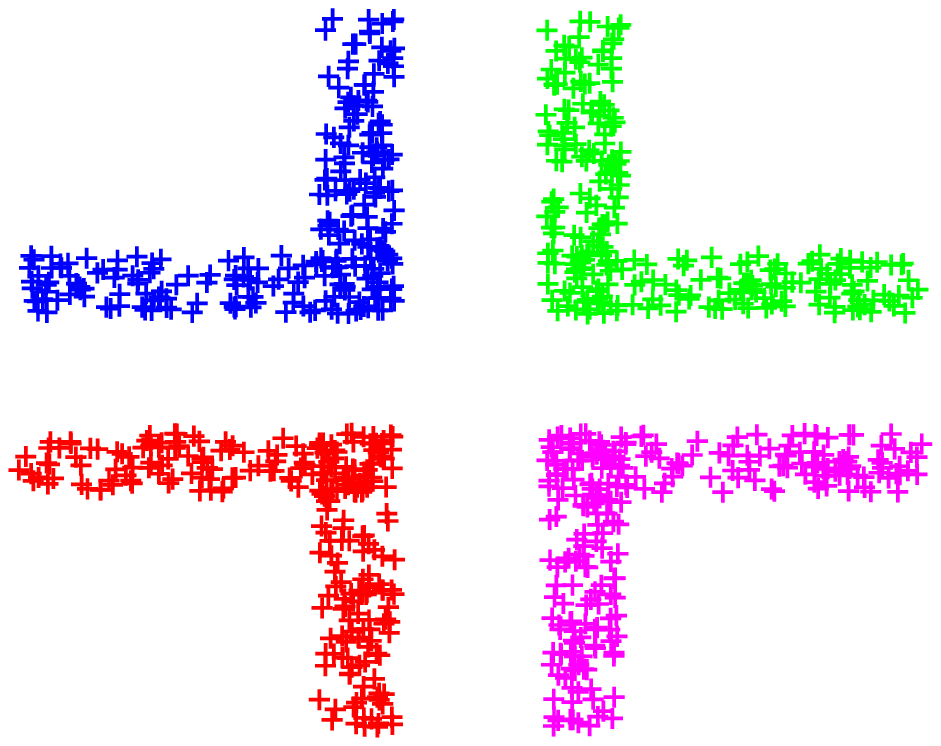}}
		\centerline{(a) ground-truth}\medskip
	\end{minipage}
	\hfill
	\begin{minipage}[b]{0.24\linewidth}
		\centering
		\centerline{\includegraphics[width=4.5cm]{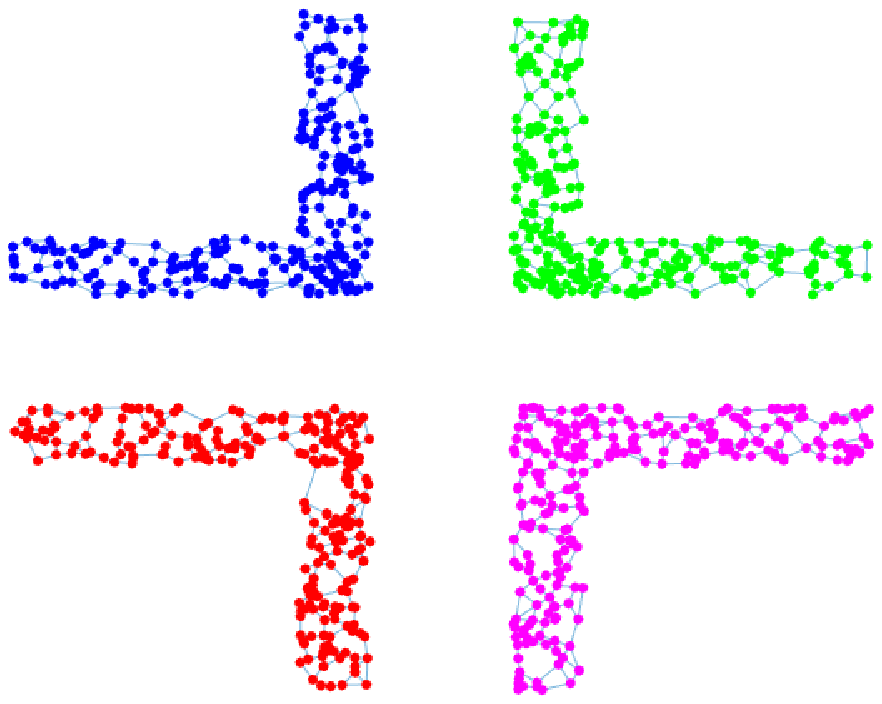}}
		\centerline{(b) SMCE (100\%)}\medskip
	\end{minipage}
	\hfill
	\begin{minipage}[b]{0.24\linewidth}
		\centering
		\centerline{\includegraphics[width=4.5cm]{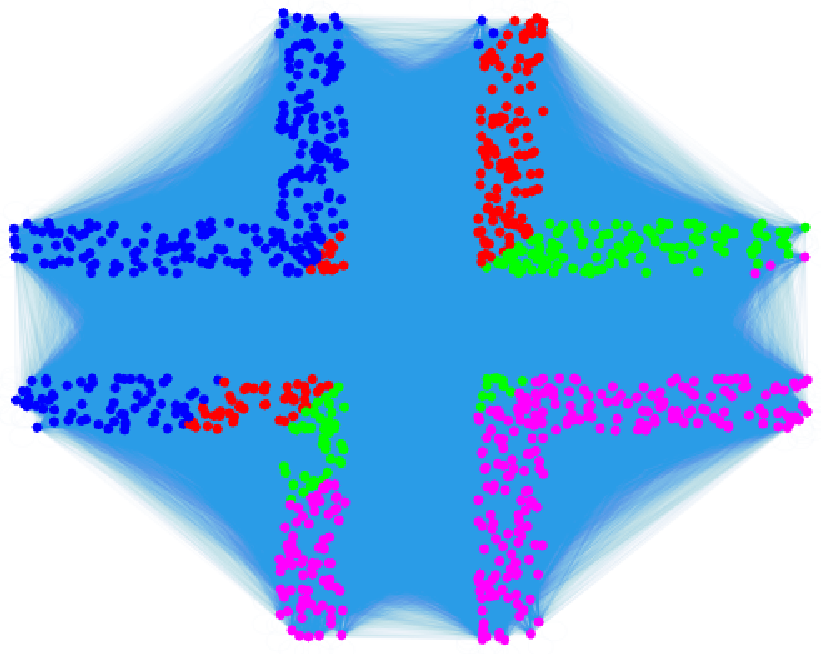}}
		\centerline{(c) SMR (65.10\%)}\medskip
	\end{minipage}
	\hfill
	\begin{minipage}[b]{0.24\linewidth}
		\centering
		\centerline{\includegraphics[width=4.5cm]{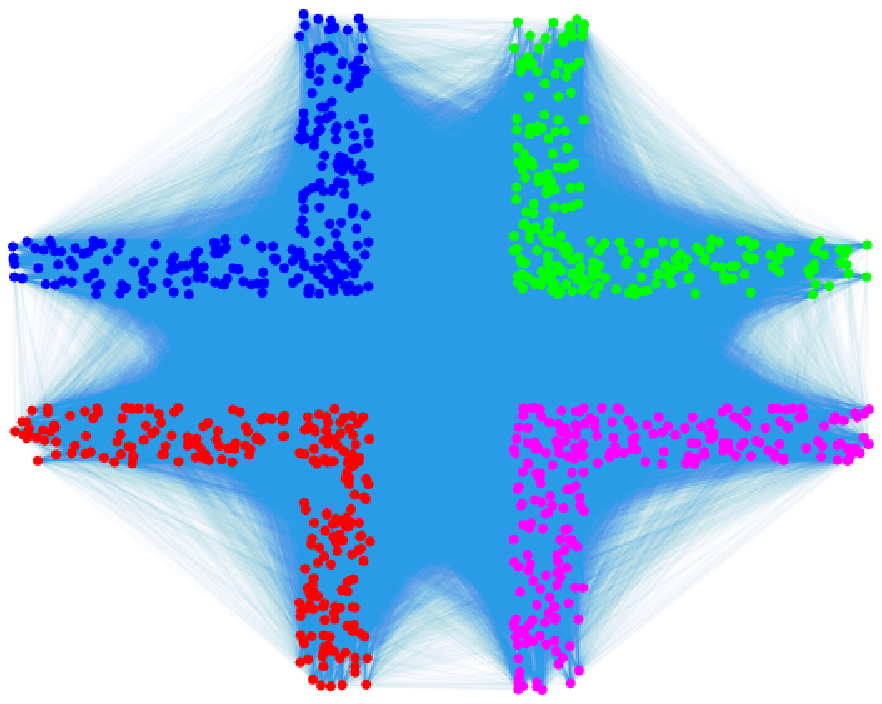}}
		\centerline{(d) LR$\ell_1$-SSC (100\%)}\medskip
	\end{minipage}
	\hfill
	\begin{minipage}[b]{0.3\linewidth}
		\centering
		\centerline{\includegraphics[width=5cm]{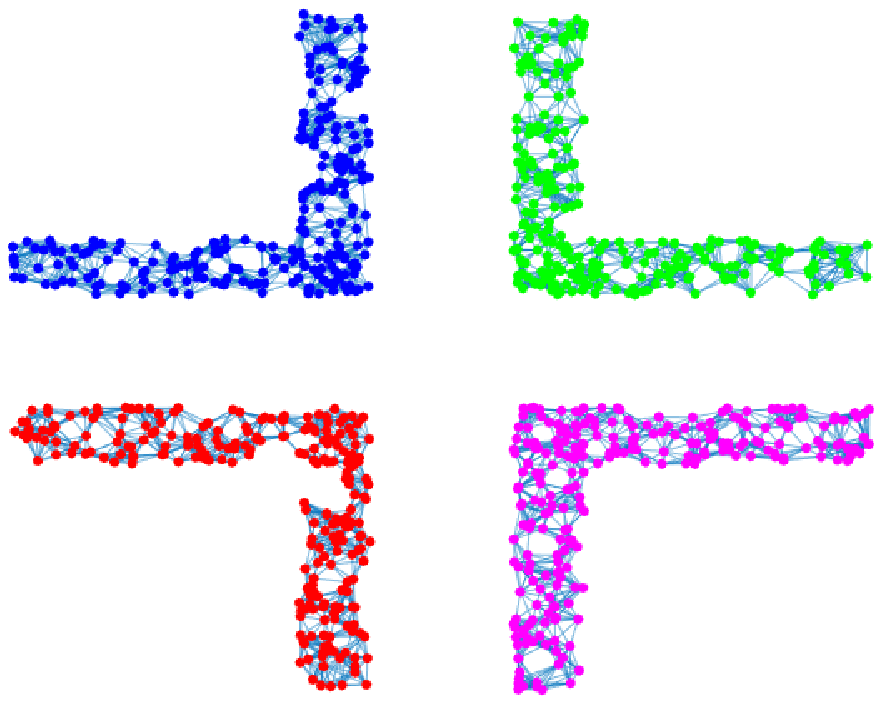}}
		\centerline{(e) KNN-SSC (100\%)}\medskip
	\end{minipage}
	\hfill
	\begin{minipage}[b]{0.3\linewidth}
		\centering
		\centerline{\includegraphics[width=5cm]{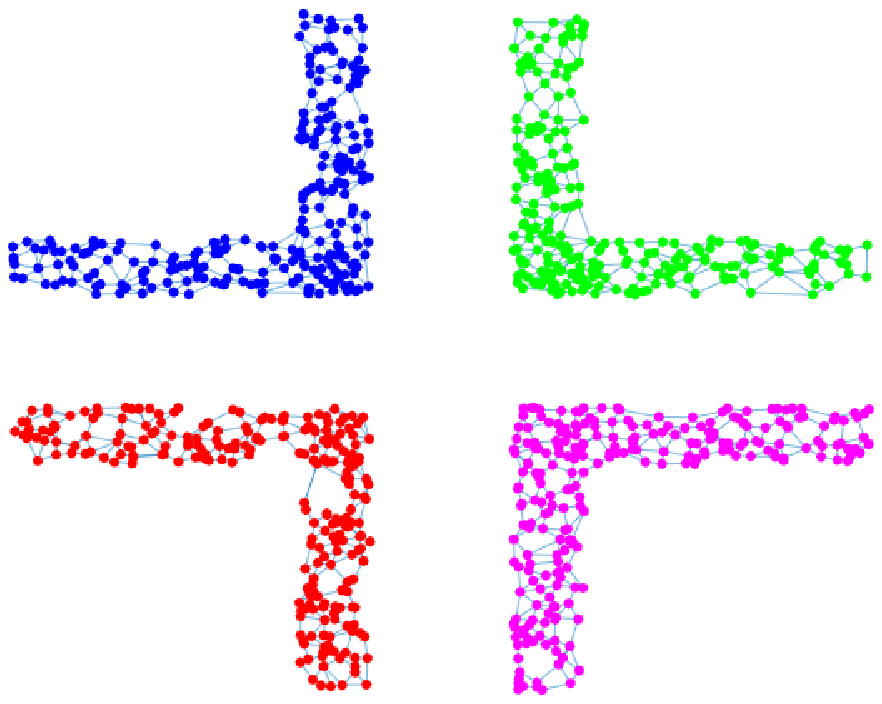}}
		\centerline{(f) KSSC (RBF) (100\%)}\medskip
	\end{minipage}
	\hfill
	\begin{minipage}[b]{0.3\linewidth}
		\centering
		\centerline{\includegraphics[width=5cm]{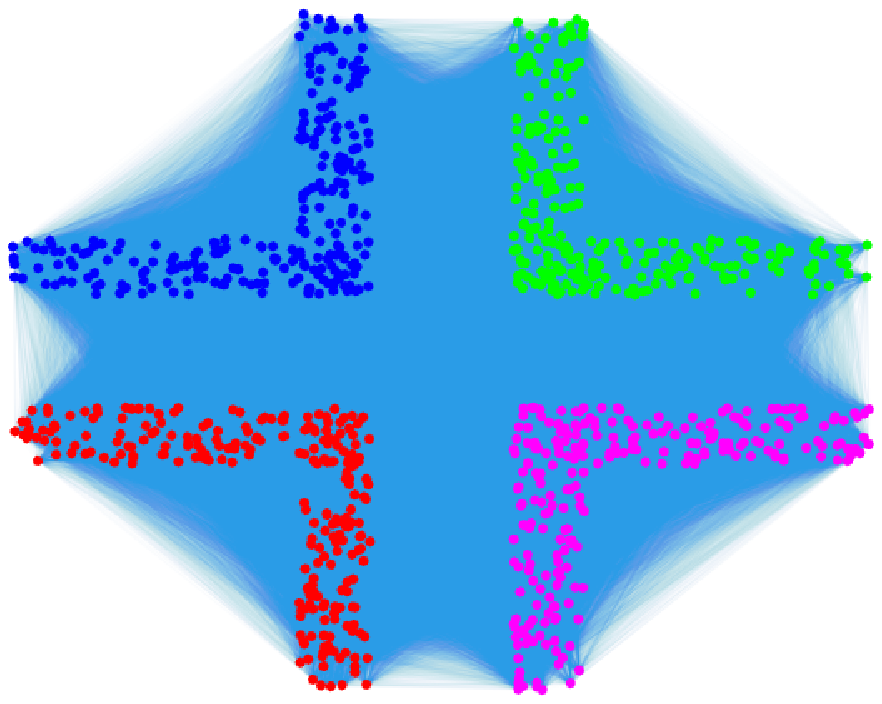}}
		\centerline{(g) LKG (100\%)}\medskip
	\end{minipage}
	\hfill
	\caption{Comparing the performance of nonlinear SC approaches on a four corners synthetic data set. The accuracy of the displayed instances is indicated in brackets. The average accuracy over the 10 randomly generated instances can be found in Table~\ref{nonlinear_avg}.
	} 
	\label{corners}
\end{figure*}

\begin{figure*}[!htbp]
	\begin{minipage}[b]{0.24\linewidth}
		\centering
		\centerline{\includegraphics[width=4.75cm]{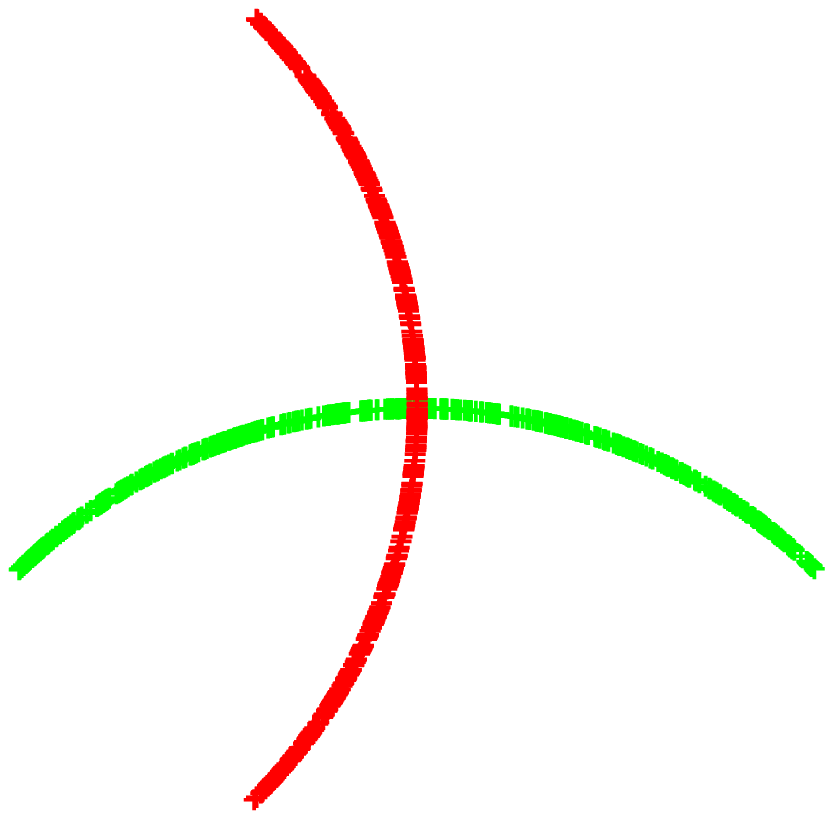}}
		\centerline{(a) ground-truth}\medskip
	\end{minipage}
	\hfill
	\begin{minipage}[b]{0.24\linewidth}
		\centering
		\centerline{\includegraphics[width=4.75cm]{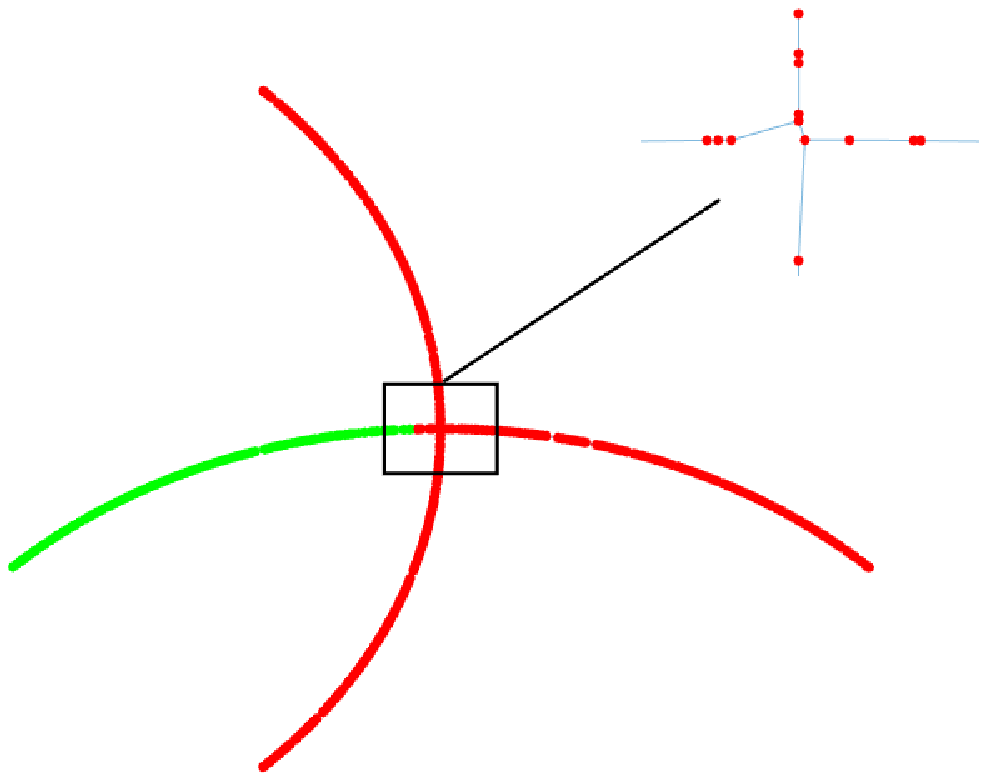}}
		\centerline{(b) SMCE (74.40\%)}\medskip
	\end{minipage}
	\hfill
	\begin{minipage}[b]{0.24\linewidth}
		\centering
		\centerline{\includegraphics[width=4.75cm]{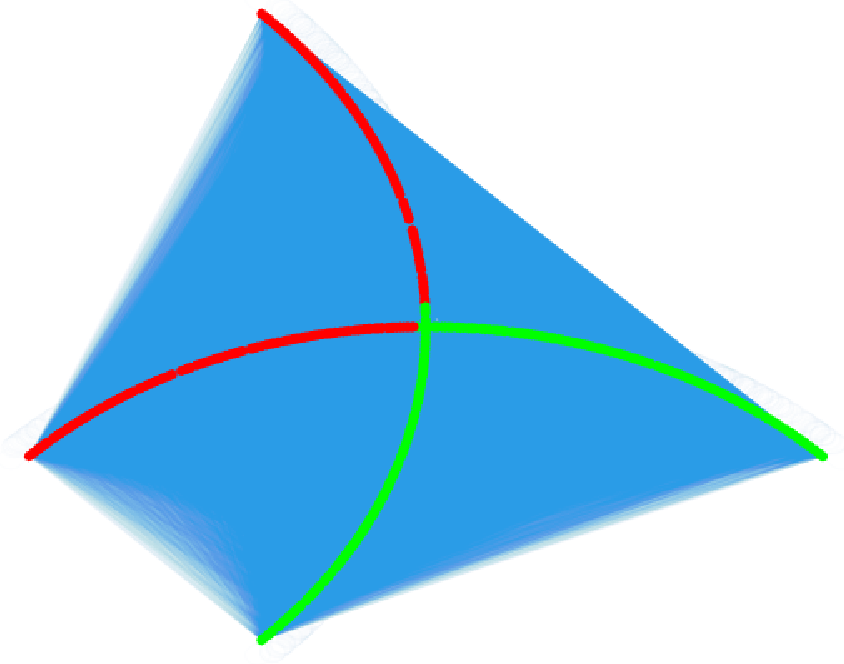}}
		\centerline{(c) SMR (50.30\%)}\medskip
	\end{minipage}
	\hfill
	\begin{minipage}[b]{0.24\linewidth}
		\centering
		\centerline{\includegraphics[width=4.75cm]{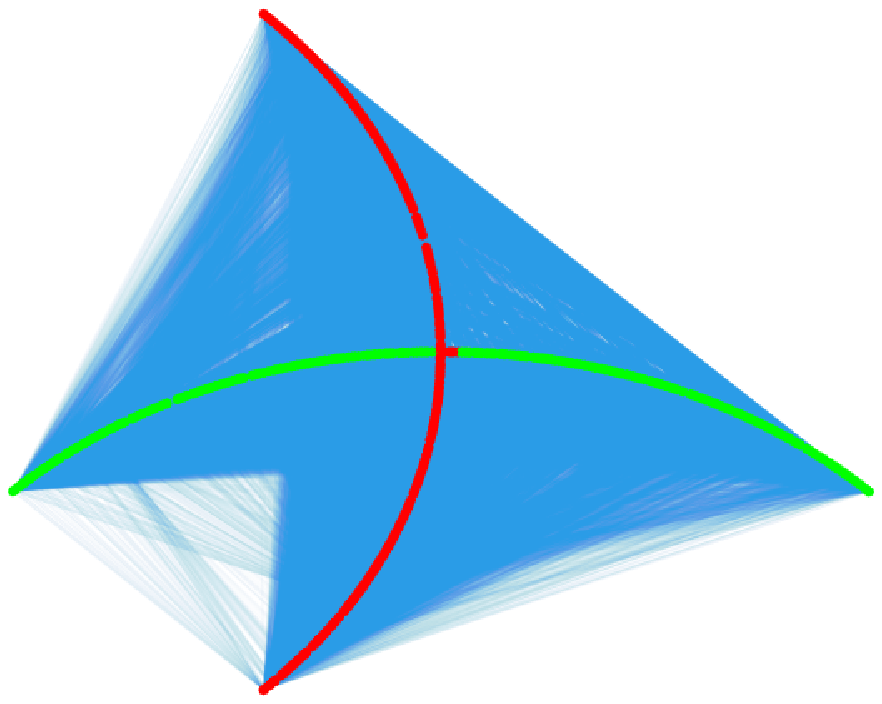}}
		\centerline{(d) LR$\ell_1$-SSC (98.80\%)}\medskip
	\end{minipage}
	\hfill
	\begin{minipage}[b]{0.3\linewidth}
		\centering
		\centerline{\includegraphics[width=5cm]{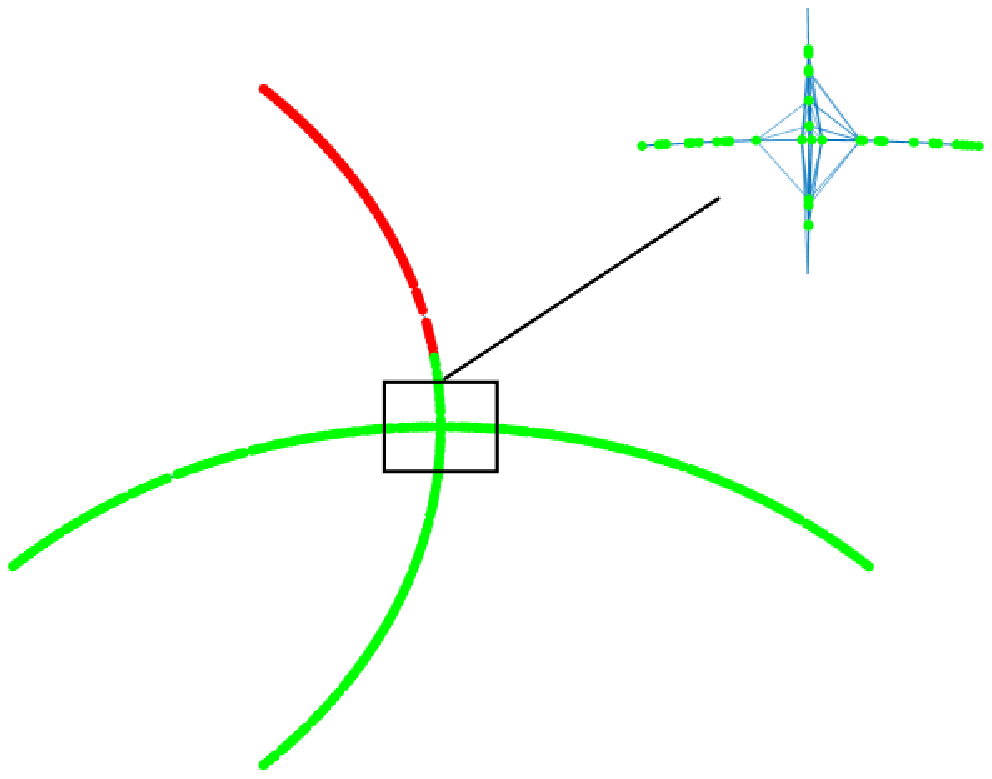}}
		\centerline{(e) KNN-SSC (70.30\%)}\medskip
	\end{minipage}
	\hfill
	\begin{minipage}[b]{0.3\linewidth}
		\centering
		\centerline{\includegraphics[width=5cm]{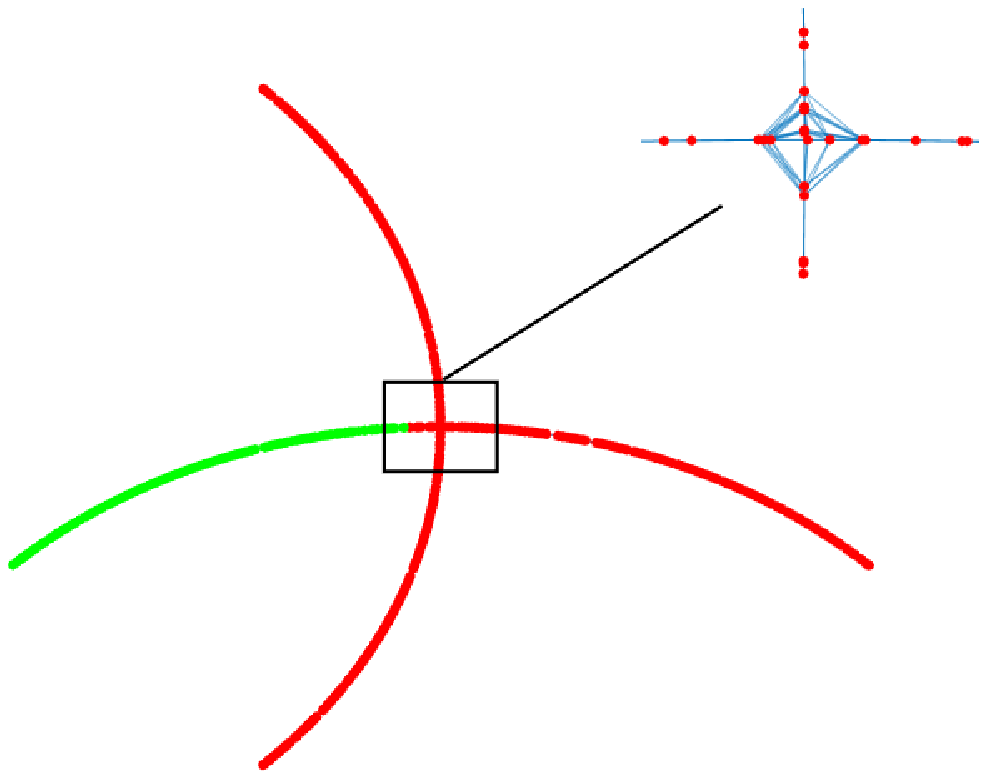}}
		\centerline{(f) KSSC (RBF) (74.10\%)}\medskip
	\end{minipage}
	\hfill
	\begin{minipage}[b]{0.3\linewidth}
		\centering
		\centerline{\includegraphics[width=5cm]{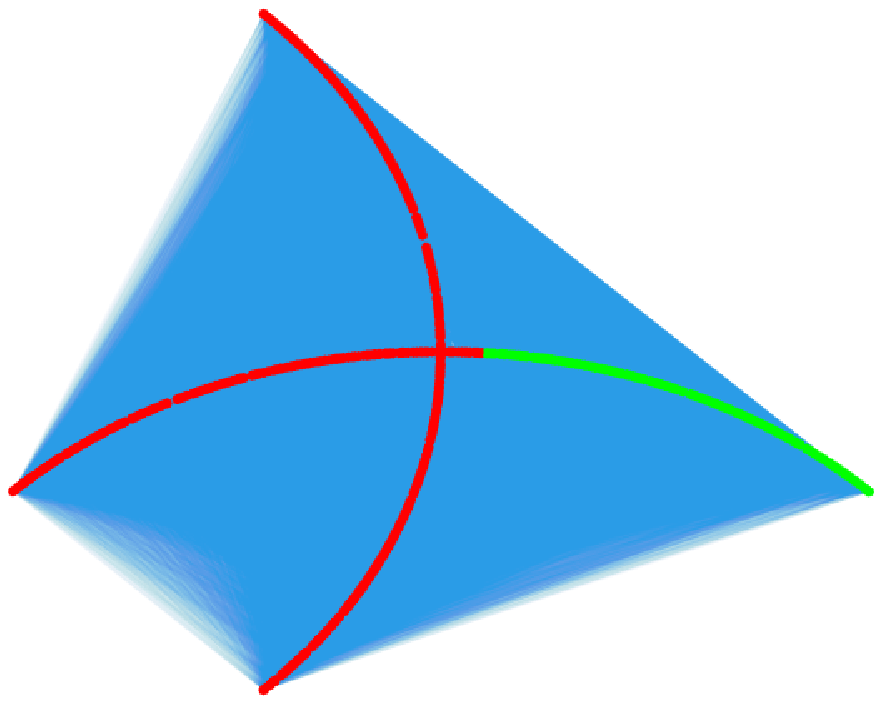}}
		\centerline{(g) LKG (74.10\%)}\medskip
	\end{minipage}
	\hfill
	\caption{Comparing the performance of nonlinear SC approaches on a two arcs synthetic data set. 
		The accuracy of the displayed instances is indicated in brackets. The average accuracy over the 10 randomly generated instances can be found in Table~\ref{nonlinear_avg}.
	} 
	\label{arcs}
\end{figure*}

In this section, we analyze the nonlinear SC approaches on 4 well-known widely used nonlinear synthetic data sets in 2 dimensions: two half-kernels, two spirals, four corners and two arcs. 
For each data set, we sample 1000 total data points from the corresponding manifolds\footnote{We used the code from https://www.mathworks.com/matlabcentral/fileexchange/41459-6-functions-for-generating-artificial-datasets.}. 
The data is normalized to have values between 0 and 1.
The obtained connectivity graphs and the corresponding segmentation are shown in 
Figures~\ref{halfkernel}, 
\ref{spirals}, 
\ref{corners} 
and~\ref{arcs}. 
For each data set, the ground-truth clustering is shown and is followed by the obtained connectivity graphs of the nonlinear approaches. The average and standard deviation of clustering accuracy for 10 different sampling of the manifolds are reported in Table~\ref{nonlinear_avg}.

\begin{center}
	\begin{table*}[!htbp]
		\begin{center}
			\caption{The average and standard deviation of the accuracy, in percent, of SC approaches over 10 trials for nonlinear synthetic data sets. The best accuracy is indicated in bold, the second best is underlined}
			\label{nonlinear_avg}   
			\small\addtolength{\tabcolsep}{-1pt}
			\resizebox{\textwidth}{!}{
				\begin{tabular}{|cc||cccccccc||cc|}
					\hline
					& \multicolumn{1}{c||}{} & SMCE & SMR & KSSC & LRL$\ell_1$-SSC & LKG & KNN-SSC & DSC-Net & DASC & SSC-L2 & LRR \\
					\hline 
					\multirow{2}{*}{Half-kernels} & \multicolumn{1}{|c||}{mean} & \textbf{100} & 63.06 & \textbf{100} & 68.88 & 62.60 & \underline{99.99} & 58.23 & 64.38 & 70.87 & 62.00 \\ 
					& \multicolumn{1}{|c||}{std} & 0 & 2.05 & 0 & 1.10 & 2.96 & 0.03 & 4.08 & 8.58 & 2.08 & 2.20 \\ \hline
					\multirow{2}{*}{Two Spirals} & \multicolumn{1}{|c||}{mean} & 98.68  & 64.64 & \textbf{100} & 51.05 & 68.02 & \underline{98.99} & 55.58 & 60.66  & 63.98 & 64.84 \\ 
					& \multicolumn{1}{|c||}{std} & 1.69 & 1.54 & 0 & 0.61 & 6.23 & 2.28 & 3.16 & 4.53 & 2.16 & 2.29 \\ \hline
					\multirow{2}{*}{Four Corners} & \multicolumn{1}{|c||}{mean} & \textbf{100} & 65.19 & \textbf{100} & \underline{99.98} & 99.43 & \textbf{100} & 48.08 & 58.48 & 58.23 & 62.97\\ 
					& \multicolumn{1}{|c||}{std} & 0 & 1.54 & 0 & 0.06 & 1.80 & 0 & 6.18 & 5.51 & 5.82 & 2.51\\\hline
					\multirow{2}{*}{Two Arcs} & \multicolumn{1}{|c||}{mean} & 72.93 & 51.92 & 73.13 & \textbf{98.36} & 70.42 & \underline{73.18} & 58.07  & 66.37 & 55.99 & 51.56 
					\\ 
					& \multicolumn{1}{|c||}{std} & 1.57 & 1.51 & 1.35 & 0.91 & 6.39 & 2.06 & 6.23 & 7.72 & 1.71 & 1.27\\ \hline
			\end{tabular}}
		\end{center}
	\end{table*}
\end{center}
We observe that:
\begin{itemize}
	
	\item Linear SC algorithms, SSC-L2 and LRR do not perform well on nonlinear data sets. Similar result were obtained for SSC-L1, and hence we did not include them in Table~\ref{nonlinear_avg}.
	
	\item SMCE, KSSC with RBF kernel and KNN-SSC successfully produced sparse but well-connected graph on the half-kernels and 4 corners data sets.
	
	\item For the two spirals data set, the performances of SMCE and KNN-SSC are not consistent over the 10 different samplings of the manifolds. In particular, the inner ending points of the two spirals might not have subspace preserving representation and the performance might depend on the sampling process and the spectral clustering step.
	
	\item Both SMCE and KSSC fail to properly segment the two arcs data set. This is due to the intersection of the two nonlinear manifolds. Hence, we can conclude that \emph{SMCE and KSSC might not be successful in segmenting intersecting manifolds}. 
	Note that the rest of the connectivities are correct and sparse, the failure comes from the confusion at the points close to the intersection. 
	
	\item SMR is not successful in clustering nonlinear manifolds. The coefficient matrix corresponding to SMR has many wrong connections. 
	This suggests that, on top of the grouping property and the locality preserving regularizations, sparsity also plays a critical role. 
	
	\item LR$\ell_1$-SSC is the only successful algorithm in clustering two arcs (although it produces many wrong connections). This algorithm is also successful in clustering the four corners data set. But as the nonlinearity increases (such as two spirals or half kernels), the clustering performance decreases significantly. 
	
	\item The two arcs data set (see Figure~\ref{arcs}) is an example of two intersecting manifolds; they intersect at (0.29,0.71). We have observed that if the two arcs intersect at (0,0) instead, then the clustering accuracy of KNN-SSC and SMR reaches 100\%. This observation does not hold for other algorithms but this suggests that additional affinity constraints on the coefficient matrix might lead to better results in some cases.

	\item Neural network based approaches, that is, DSC-Net and DASC, perform very poorly. The reason is that their embedding acted as an almost identity mapping and merely rotated the input datta. In other words, they failed to produce a meaningful embedding in the encoder output. This is consistent with the results reported in~\cite{haeffele2020critique}. 
\end{itemize}

\subsection{Real-world data sets}

In this section, we compare the performances of SC algorithms on three widely used real-world data sets:

\begin{itemize}
	
	\item \textbf{Extended Yale B \cite{georghiades1998illumination,lee2005acquiring}:} The Extended Yale B data set contains 2,414 images of 38 individuals, taken under different illumination conditions. There are 64 frontal-face images, each of the size $192 \times 168$ pixels, per person. The images were downsampled to $48 \times 42$ pixels, similar to the general setting provided in \cite{elhamifar2013sparse}. Different number of clusters are considered to analyze the performance with respect to the number of subspaces. 
	
	\item \textbf{Coil-20 \cite{nene1996columbia}:} Columbia Object Image Library contains 72 images from 20 objects. The images were taken from objects in different poses. The images are of size $128 \times 128$ pixels and following~\cite{ji2017deep}, we downsampled them to $32 \times 32$ pixels. 
	
	\item \textbf{MNIST test data set:} The MNIST data set contains 70,000 images of 10 handwritten digits where each image is of size 28-by-28 pixels. In particular, this data set includes training and test sets of 60,000 and 10,000 images, respectively. Due to high computational complexity of the majority of the nonlinear approaches (see Section~\ref{sec:compcost}), 
	we only use the MNIST test data set for the evaluation, which is a common practice for evaluating computationally expensive clustering approaches~\cite{yang2016joint, yang2019deep}.  
\end{itemize}

The parameters of each approach are set to the values introduced by the code of the original papers (if the code is available). For the approaches that are implemented by us, the parameters are tuned within the ranges proposed by the corresponding paper for the best result for each data set; see Appendix~\ref{param} for selected parameters of each approach.  
The results (namely, the accuracy and the NMI) 
on Extended Yale B for different number of subjects, COIL-20, and MNIST are summarized in Tables~\ref{yaleb}, \ref{coil20}, and \ref{mnist}, respectively. 
The value of M
indicates that the corresponding algorithm ran out of 16 GB memory, and the numbers indicated with * correspond to the algorithms that did not converge within 6 hours. 

We observe the following:  
\begin{itemize}
	
	\item On the Extended Yale B data set, SSC-L1 has the overall best performance. As the number of subjects increases, the neural network based approaches, including DSC-Net-L2 and DASC, have the overall second best performances. 
	LKG has the  worst performance, even worse than KSSC with a single kernel. 
	
	\item The high performance of SSC-L1 on Extended Yale B highlights the importance of modeling noise and corruptions while clustering. Based on the experiments in~\cite{elhamifar2013sparse} (see Section 7.2.1), after removing the sparse noise by applying RPCA~\cite{candes2011robust} on images from each cluster separately, the clustering accuracy of SSC increases to 100\%. This suggests that Extended Yale B can be well modeled by linear subspaces after the noise is eliminated.
	
	\item On the COIL-20 data set, SMCE performs the best followed by KSSC. The neural-network based approaches perform the worse. Note that, the high accuracy reported on their corresponding papers heavily relies on the post-processing step rather than the obtained coefficient matrix by the model~\cite{haeffele2020critique}.
	
	\item On the MNIST data set, KNN-SSC performs the best, followed closely by SSC-L2 and SMCE. 
	KSSC performs better than neural-network based approaches. The Laplacian regularized LR$\ell_1$-SSC did not converge within 6 hours of timelimit. This is due to the fact that the iterative optimization process based on ADMM involves solving a Sylvester equation in each iteration which is very time consuming compared to solving a linear system in the original SSC algorithm. The same observation holds for LRR as well because of the computation of an SVD in each iteration of ADMM. The multiple-kernel based approach of LKG ran out of 16 GB memory which highlights the high computational cost of kernel approaches that rely on computing multiple pair-wise dense Gram matrices. 
\end{itemize}

\begin{center}
	\begin{table*}[!htbp]
		\begin{center}
			\caption{Comparison of nonlinear SC algorithms on the Extended Yale B data set. The best value is highlighted in bold, the second best is underlined.} 
			\label{yaleb}  
			\small\addtolength{\tabcolsep}{-1pt}
			\begin{tabular}{|c||ccccccccccc|}
				\hline
				Metric & SMCE & SMR & LR$\ell_1$-SSC & KNN-SSC & KSSC & LKG & DSC-L2 & DASC & SSC-L2 & SSC-L1 & LRR \\ \hline
				\multicolumn{1}{c}{2 subjects} &&&&&&&&&&& \multicolumn{1}{c}{}\\ \hline
				ACC &\textbf{100} &95.31 & \underline{99.22} &\textbf{100} &88.28 &50.78	&92.97	& 93.75 & \underline{99.22} & \textbf{100} & 83.59 \\
				NMI &\textbf{100} &73.06 & \underline{94.18} &\textbf{100} &54.12 &0.01	&64.77	& 66.27 & \underline{94.18} & \textbf{100} & 48.39 \\ \hline
				\multicolumn{1}{c}{3 subjects} &&&&&&&&&&&  \multicolumn{1}{c}{}\\ \hline
				ACC &55.21 &59.89 &79.69 &58.85 &61.45	&34.37	&60.94	& 86.45 & \underline{88.02} & \textbf{99.47} & 59.37 \\
				NMI &35.21 &39.06 &58.33 &35.98 &44.52	&0.02	&47.22	& 67.04 & \underline{69.33} & \textbf{97.55} & 38.26 \\ \hline
				\multicolumn{1}{c}{5 subjects} &&&&&&&&&&&  \multicolumn{1}{c}{}\\ \hline
				ACC &60.00 &63.12 &84.37 &49.68 &64.06	&24.37	&65.93	& \underline{89.06} & \underline{89.06} & \textbf{99.37} & 50.93 \\
				NMI &54.79 &53.04 &76.73 &38.06 &56.87	&1.26	&58.49	& 77.99 & \underline{80.72} & \textbf{97.99} & 41.37 \\ \hline
				\multicolumn{1}{c}{10 subjects} &&&&&&&&&&&  \multicolumn{1}{c}{}\\ \hline
				ACC &51.56& 62.34 &64.22 &57.19 &44.06	&21.25	& 69.68	& \underline{73.75} & 54.06 & \textbf{91.25} & 61.87 \\
				NMI &50.98& 63.61 &64.38 &57.63 &43.51	&14.64	& 67.89	& \underline{70.66} & 53.74 & \textbf{87.97} & 62.91 \\ \hline
				\multicolumn{1}{c}{20 subjects} &&&&&&&&&&&  \multicolumn{1}{c}{}\\ \hline
				ACC &57.66& 71.48 &46.56 &58.05 &47.65	&18.43	&71.79	& \underline{71.87} & 66.17 & \textbf{87.03} & 70.31 \\
				NMI &60.61& 74.35 &51.42 &67.02 &52.08	&20.25	&74.05	& \underline{75.15} & 69.30 & \textbf{85.99}  & 72.58 \\ \hline
				\multicolumn{1}{c}{38 subjects} &&&&&&&&&&&  \multicolumn{1}{c}{}\\ \hline
				ACC &55.02 & \underline{73.56} &46.01 &52.09 &50.29	&18.54	&69.32	& \textbf{74.38} & 65.21 & 73.06 & 69.86\\
				NMI &62.76 & \underline{77.74} &52.62 &63.46 &59.56	&27.41	&74.41	& \textbf{77.90} & 69.92 & 77.20 & 73.89 \\ \hline
			\end{tabular} 
		\end{center}
	\end{table*}
\end{center}

\begin{center}
	\begin{table*}[!htbp]
		\begin{center}
			\caption{Comparison of nonlinear SC algorithms on the COIL20 data set. The best value is highlighted in bold, the second best is underlined.} 
			\label{coil20}  
			\small\addtolength{\tabcolsep}{-1pt}
			\begin{tabular}{|c||ccccccccccc|}
				\hline
				Metric & SMCE & SMR & LR$\ell_1$-SSC & KNN-SSC & KSSC & LKG & DSC-L2 & DASC & SSC-L2 & SSC-L1 & LRR \\ \hline
				ACC &\textbf{90.76}	&64.03	&82.29	&79.44	& \underline{85.06}	&73.40	&59.58& 57.77 & 75.62 & 74.86 & 60.83\\
				NMI &\textbf{96.71}	&73.57	&93.21	&91.12	& \underline{95.70}	&82.72	&74.11& 67.32 & 88.55 & 90.30 & 74.50\\ \hline
			\end{tabular} 
		\end{center}
	\end{table*}
\end{center}

\begin{center}
	\begin{table*}[!htbp]
		\begin{center}
			\caption{Comparison of nonlinear SC algorithms on the MNIST data set. The best value is highlighted in bold, the second best is underlined. The symbol M
				indicates that the algorithm ran out of 16 GB memory.} 
			\label{mnist}  
			\small\addtolength{\tabcolsep}{-1pt}
			\begin{tabular}{|c||ccccccccccc|}
				\hline
				Metric & SMCE & SMR & LR$\ell_1$-SSC & KNN-SSC & KSSC & LKG & DSC-L2 & DASC & SSC-L2 & SSC-L1 & LRR \\ \hline
				ACC &68.77	&54.64	& 55.08*	&\textbf{70.48}	&64.32	& M	&55.46	&46.28 & \underline{70.12} & 59.03 & 43.92*\\
				NMI &73.06	&54.97	& 62.26*	& \underline{73.98}	&70.99	& M	&56.46	&44.51 & \textbf{74.89} & 71.06 &35.50*\\ \hline
			\end{tabular} 
		\end{center}
	\end{table*}
\end{center}

\section{Discussion, challenges and future directions of research} \label{future}

Following the extensive numerical comparisons, in this section, we discuss nonlinear SC algorithms from different perspectives with an emphasis on the existing challenges and possible future research directions.

\subsection{Discussion on the performance of nonlinear SC algorithms} \label{discussion}

The main advantages and disadvantages of each category of nonlinear SC algorithms are summarized in Table~\ref{categorization}. Based on the experimental results, no nonlinear approach is completely superior to the other nonlinear approaches. 
\begin{table}[!htbp]
	\caption{Advantages and disadvantages of representative nonlinear SC  approaches}\label{categorization} \small\addtolength{\tabcolsep}{-1pt}
	\begin{tabularx}{\textwidth}{>{\raggedright}c|X|X} \hline
		\thead{Category} & \thead{Advantages} & \thead{Disadvantages} \\ \hline
		Locality preserving & 
		\begin{itemize}
			\item Convenient adaptation of classic linear SC
			\item Interpretable and intuitive
			\item supporting both independent and disjoint subspaces 
		\end{itemize}
		&
		\begin{itemize}
			\item Dependent on the estimation of locality parameter 
			\item Fixed locality estimation for all data points 
		\end{itemize}	
		\\ \hline
		Kernel Based & 
		\begin{itemize}
			\item Based on classical mathematical theory to implicitly transform data into higher dimensions
			\item Easily applicable to the majority of linear formulations via the kernel trick
			\item Supporting both independent and disjoint subspaces
		\end{itemize}
		& 
		\begin{itemize}
			\item No guarantee that kernels lead to implicit feature space suitable for linear subspace clustering 
			\item Difficulty in choosing the right kernel(s)
			\item Difficulty in understanding and interpreting results
			\item Memory inefficiency due to Gram matrix
		\end{itemize}
		\\ \hline
		Neural-network Based &
		\begin{itemize}
			\item Learning the nonlinear transformation based on the data
			\item High capacity for learning complex data representations
		\end{itemize}
		&
		\begin{itemize}
			\item Ill-posed and might lead to trivial embeddings
			\item No theoretical guarantees 
			\item Memory inefficiency due to self-expressiveness representation in latent space
			\item Difficult optimization and no general rule for selecting critical parameters such as the number of epochs used for training 
			\item Failure to cluster disjoint subspaces 
		\end{itemize}
		\\ \hline
	\end{tabularx}
\end{table}

Locality preserving approaches, which are based on the assumption that the manifolds are smooth and well sampled, are intuitive while they can be easily designed by adapting the linear models; see Section~\ref{local}. 
However, many of the approaches in this category are sensitive to intersecting subspaces, and they depend on locality estimation parameters (such as $k$ in KNN based similarity matrix construction approaches).

Kernel-based nonlinear approaches can be designed as direct extensions of linear models as well (depending on the loss function, which is usually based on the Frobenius norm).  However, there is no guarantee that the corresponding feature space is more suitable for linear models. Moreover, with no prior knowledge, choosing the right kernel and tuning its parameters is highly nontrivial. We noticed that even though learning weighted combination of multiple kernels might be expected to reduce the sensitivity to the kernel parameters,
defining a \emph{proper criterion} to adaptively choose the weights of each kernel is not straightforward for all data sets and applications. 

With the growing interest in neural network and their recent success in learning complex data representations in many fields, deep SC methods have received considerable attention in the past few years. These approaches tend to \emph{learn} the kernel embedding function instead of using the existing predefined kernels. 
However, Haeffele et al.~\cite{haeffele2020critique} pointed out potential theoretical concerns for neural network based nonlinear SC approaches that rely on autoencoder regularization. It is theoretically argued that the underlying model is ill-posed and encouraging the latent representation to have a union of subspaces structure through the additional self-expressive loss term (or layer) is insufficient and leads to degenerate and trivial embeddings of the data in many cases.
The potential degenerate embeddings is also noticed in deep k-means algorithm under the name of ``scaling-down phenomenon"~\cite{fard2020deep} where it is possible to make the clustering loss (for SC, the self-expressive loss) arbitrarily small without changing the reconstruction loss of the network. This problem can be avoided by several strategies such as regularizing the network weights or the norm of the embedded data. 
However, even using proper regularizations,  
the joint learning of embedding and self-expressive representation by autoencoder regularized nonlinear SC approaches can still lead to trivial data geometry in the embedding space. These issues are pointed out from a theoretical perspective, under the assumption that the auto-encoder is highly expressive.  Intuitively, based on this assumption, the network can generate many possible embeddings of the data in the latent space, $Z_\Theta$, while the decoder is still capable of reconstructing the data accurately. 

However, in practice and for both synthetic and real-world data sets, the performance of current neural-network based nonlinear SC approaches is often inferior to other nonlinear approaches. In fact, DSC-Net as the representative approach, strongly benefits from an ad-hoc post-processing step to improve the quality of the obtained coefficient matrix (see the second introduced post-processing step in Section~\ref{sec:post_proc}). Our numerical experiments showed that DSC-Net does not necessarily lead to subspace preserving representations in many cases including the data drawn from independent subspaces. Hence, we believe that neural network based nonlinear SC through enforced self-expressive representation in latent space does not impose significant constraints
on the geometric arrangement of the embedded points and appears to be not sufficient nor necessarily successful to recover the union of latent linear subspace structure.

\subsection{Challenges and Future directions of research}

Nonlinear subspace/manifold clustering is a challenging problem for which many approaches have been proposed. However, many challenges remain, including the following: 

\begin{itemize}
	
	\item Sensitivity to parameters: 
	SC is an unsupervised problem with no available labelling information. 
	The majority of algorithms discussed in this paper are found to be sensitive to their parameters. 
	However, with no prior knowledge, setting these parameters is highly nontrivial and tricky. This is especially true for neural network based approaches that are very sensitive to the number of epochs for the pre-training and fine-tuning steps. 
	In our experimental results, we fine-tuned the number of the epochs; however, we noticed that the performance can degrade significantly as the number of epochs increases.  Multiple kernel learning approaches are also sensitive to their parameters; in particular to the lack of ``good" criterion for choosing the weight of each Kernel. 
	Developing algorithms with less sensitivity to their parameters (and ideally parameter free) is a general challenging research direction for unsupervised learning tasks.
	
	\item Scalabe algorithms: The majority of nonlinear SC based on self-expressiveness are computationally expensive; see Section~\ref{sec:compcost}. 
	Locality preserving algorithms are usually based on estimating a local neighborhood graph which has at least the complexity of $O(n^2)$ and if the neighboorhood graph is not sparse, the memory requirement would be huge too. Computing the Gram matrix in kernel based approaches is another example of computational inefficiency. Except a few works that address scalability issue in deep self-expressive based SC approaches, e.g., \cite{seo2019deep,zhang2018scalable}, 
	nonlinear SC algorithms	 are not practical for data sets with more than 10,000 data points. Computing and storing the $n\times n$ coefficient matrix $C$ is specifically an important bottleneck in these approaches.
	
	\item Theoretical guarantees: In contrast to linear SC algorithms whose theoretical aspects are well studied and understood, nonlinear alternatives are mostly intuition driven. Except for the few works in locality preserving nonlinear approaches (such as~\cite{hu2014smooth}), there is no clear theoretical analysis on conditions for accurate subspace recovery for nonlinear subspaces. The lack of theoretical analysis is also evident in neural network based nonlinear SC approaches. Understanding the \emph{black box} of neural networks is in general an ongoing research direction and combining these highly expressive models with SC based on mere intuition might  result in trivial geometric embeddings~\cite{haeffele2020critique} (see also the discussion in Section~\ref{discussion}). 
	Hence, understanding and analyzing nonlinear approaches from the theoretical standpoint can help avoiding ill-posed models that might be intuitively appealing at first.   
	
	\item Non-image data extensions: Almost all nonlinear SC approaches are evaluated on image data sets such as images of faces, objects and handwritten digits. Investigating nonlinear structures in other data formats is another interesting research direction. There are selective approaches for subspace data on Grassmann manifolds such as \cite{wang2017localized,wang2016kernelized} or for symmetric positive definite (SPD) matrices such as \cite{yin2016kernel,hechmi2019multi}, 
	but the major focus in the past years has been on image data.
	
	\item Robustness: The majority of nonlinear SC approaches rely on the Frobenius norm to measure the data fitting error,  and hence are sensitive to gross corruptions such as occlusions, and the presence of outliers. Improving the robustness in such scenarios is definitely an important practical aspect.

	\item Clustering data with intersecting manifolds: The results on the synthetic data sets highlighted the difficulty for nonlinear SC approaches to cluster manifolds that are close to each other or intersect.  
	The closer the manifolds are, the more the locality preserving approaches are sensitive to the locality controlling parameter, and similarly the kernel based approaches to the kernel parameters. Moreover, learning discriminating features to disentangle the multiple manifolds structure in the embedding space is harder for neural networks for data from closer manifolds. Hence, a general challenging research direction is to develop approaches to capture internal multiple manifold structures when they are spatially close in the ambient space. 
\end{itemize}

\section{Conclusion} \label{conc}

In this paper, we presented a comprehensive overview of nonlinear subspace clustering (nonlinear SC) approaches, our main focus being on algorithms based on self-expressiveness.
In Section~\ref{nsc}, we provided a taxonomy for classifying nonlinear SC approaches into three broad categories: locality preserving, kernel based, and neural network based approaches. The approaches within each category were further divided into detailed subcategories and were thoroughly reviewed and summarized. 
In Section~\ref{sec:compcost}, 
we briefly discussed the computational cost of these approaches.  
In Section~\ref{eval}, the representative approaches within each (sub)-category were extensively compared on synthetic and real-world data sets. Based on the obtained results, Section~\ref{future} discussed the advantages and disadvantages of the different algorithms, and also elaborated on the current challenges for future research.

\small 

\bibliographystyle{spmpsci} 
\bibliography{sscSurvey}

\newpage 
\appendix

\section{Synthetic data generation} \label{data_gen}
In this section, we provide the detailed data generation process for independent and disjoint linear subspaces used in Section~\ref{sec:linearresults}.

We introduce a semi-random model for generating the linear synthetic data. In this model, the bases of the subspaces are fixed, 
to control the angle between them, but the data points are generated at random from each of the subspaces. The data generation model for independent and disjoint subspaces are as follows:
\begin{itemize}
	\item \textbf{Independent subspaces:} 
	In order to generate two independent subspaces with intrinsic dimension $m$ in a $d$-dimensional space ($m < d$) with controlled affinity between subspaces, the basis of the subspaces are initially constructed in $2m$-dimensional space as follows:
	\begin{align*}
		\hat{U_1} = \binom{I_{m}}{0_{m}}, 
		\hat{U_2} = \binom{\cos(\theta)  I_{m}}{\sin(\theta)  I_{m}}, 
	\end{align*}
	where $I_{m}$ and $0_{m}$ are the identity and zero $m \times m$ matrices and $\theta \in [0, \frac{\pi}{2}]$. The affinity between the two subspaces is explicitly controlled by the parameter of $\theta$. By decreasing the value of $\theta$ from $\frac{\pi}{2}$ to 0, the affinity between the subspaces decreases and the subspace clustering task gets more challenging~\cite{soltanolkotabi2012geometric}. 
	
	Let $N$ be the total number of data points on all subspaces, with half of the data points on each subspace. The data points within each subspace in $2m$-dimensional space are randomly produced by setting linear mixture weights at random using the Gaussian distribution and multiplying them by the initial bases (in Matlab $\hat{X}_i=\hat{U}_i*randn(d,N/2)$ for $i=1,2$). Let $\hat{X}_1$ and $\hat{X}_2$ denote the $2m$-dimensional data points generated for each subspace. The $2m$-dimensional data points are transferred to the final $d$-dimensional space by multiplying them by orthogonal columns of random matrix $P \in \mathbb{R}^{d \times 2m}$ (in Matlab, $P$ is generated by $orth(randn(d,2m))$):
	\[ X_i = P \times \hat{X_i}, \ \text{for }i=1,2.
	\]
	Note that, by increasing the dimension using the orthogonal matrix $P$, we keep the affinity between subspaces in the initial $2m$-dimensional space which is controlled by the parameter $\theta$. It is a common practice to normalize the data points to have unit $\ell_2$ norm, however, since normalization ruins the structure for \emph{affine} and \emph{nonlinear} subspaces, we do not normalize the data.
	\item \textbf{Disjoint subspaces:} For disjoint synthetic subspaces, we generate $N$ samples from three subspaces (with $N/3$ samples for each subspace) similar to the independent case. In particular, three initial subspaces bases, with intrinsic dimension of $m$, are constructed in $2m$-dimensional space as:
	\begin{align*}
		\hat{U_1} = \binom{I_{m}}{0_{m}}, 
		\hat{U_2} = \binom{\cos(\theta)  I_{m}}{\sin(\theta)  I_{m}}, 
		\hat{U_3} = \binom{\cos(\theta)  I_{m}}{-\sin(\theta)  I_{m}}. 
	\end{align*}
	Identical to the previous independent subspaces generation, the initial $2m$-dimensional data points within each subspace, that is, $\{\hat{X}_i\}_{i=1}^3$ are obtained by multiplying $\{\hat{U}_i\}_{i=1}^3$ by linear combinations produced from Gaussian distribution with zero mean and standard deviation of 1. The dimension of the data points are lifted to the final $d$-dimensional space by multiplying an orthogonal randomly generated matrix $P$.
\end{itemize}

\section{Parameter setting} \label{param}

In this section, the selected parameters for each approach is listed in details in Table~\ref{tabpar} for SMCE, SMR, LR$\ell_1$-SSC, KNN-SSC, KSSC, LKG, SSC-L2, SSC-L1 and LRR; in Table~\ref{tabpar_dsc} for DSC-Net and in Table~\ref{tabpar_dasc} for DASC. 

The adaptive parameter selection in~\cite{elhamifar2013sparse} is followed for setting the regularization parameter $\lambda$ which controls the importance of the self-expressiveness term, $||X-XC||_F^2$ (or $||\Phi(X)-\Phi(X)C||_F^2$ in kernel based approaches). To this end, the following data-driven strategy is used: $\lambda = \alpha \max_i \frac{1}{\max_{j\neq i} |X(:,j)^\top X(:,i)|}$. Hence, the larger the parameter $\alpha$ is, the more important the self-expressiveness term is considered in the optimization. This parameter setting is used for LR$\ell_1$-SSC, KNN-SSC, KSSC, SSC-L1 and SSC-L2.

\newpage 
\thispagestyle{empty} 
\begin{sidewaystable}
	\begin{center}
		\begin{center}
			\caption{Parameters of the compared approaches.}
			\label{tabpar}  
			\small\addtolength{\tabcolsep}{-1pt}
			\begin{tabularx}{\textwidth}{c||XXXXX}
				\hline
				\multicolumn{1}{c||}{Approach} & \multicolumn{5}{c}{Parameters} \\
				& Linear & Nonlinear & Yale B & Coil 20 & MNIST\\
				\hline
				SMCE & $\lambda = 10$  & $\lambda = 20$ & $\lambda = 10$ & $\lambda = 20$ & $\lambda = 5$ \\
				SMR 
				& 
				$\lambda = 1$, \mbox{$k=10$} & $\lambda = 0.1$, \mbox{$k=10$} & $\lambda = 1e3$, \mbox{$k=100$} & $\lambda = 1e3$, \mbox{$k=50$} & $\lambda = 1e5$, \mbox{$k=20$}\\
				LR$\ell_1$-SSC 
				& $\alpha=20, \lambda_2=1$, \mbox{$k=10$} & 
				$\alpha=20, \lambda_2=10$, \mbox{$k=10$} 
				& $\alpha=10, \lambda_2=200$, \mbox{$k=100$}  
				& $\alpha=10, \lambda_2=20$, \mbox{$k=20$} 
				& $\alpha=10, \lambda_2=20$, \mbox{$k=20$} \\
				KNN-SSC & $\alpha= 10, k=10$ & $\alpha= 10, k=10$ & $\lambda= 0.2, k=100$ & $\lambda= 0.2, k=20$ & $\lambda=0.033, k=100$\\ 
				KSSC & $\sigma=1, \alpha = 20$ & $\sigma=0.05, \alpha =5$ & $\sigma=10, \alpha = 10$ & $\sigma=5, \alpha=10$ & $\sigma=5, \alpha=10$ \\
				LKG & $\lambda_1=0.1, \lambda_2=10$, \mbox{$\lambda_3=1$} & 
				$\lambda_1=0.1, \lambda_2=10$, \mbox{$\lambda_3=1$} & 
				$\lambda_1=0.1, \lambda_2=10$, \mbox{$\lambda_3=0.5$} & 
				$\lambda_1=0.1, \lambda_2=10$, \mbox{$\lambda_3=0.5$} & 
				$\lambda_1=0.1, \lambda_2=10$,\mbox{$ \lambda_3=0.5$}\\
				SSC-L2 & $\alpha=10$ & $\alpha=20$ & $\alpha=5$ & $\alpha=5$ & $\alpha=5$\\
				SSC-L1 & $\alpha=5$ & $\alpha=10$ & $\alpha=20$ & $\alpha=20$ & $\alpha=20$\\
				LRR & $\lambda=0.1$ & $\lambda=0.1$ & $\lambda=0.009$ & $\lambda=0.009$ & $\lambda=0.0092$ \\
				\hline
			\end{tabularx}
		\end{center}
	\end{center}
	
	\begin{center}
		\caption{Parameters of DSC-Net.}
		\label{tabpar_dsc}  
		\small\addtolength{\tabcolsep}{-1pt}
		\begin{tabularx}{\textwidth}{c||X}
			\hline
			\multicolumn{1}{c||}{data set} & \multicolumn{1}{c}{Parameters} \\ \hline
			Linear & 3-layered fully connected autoencoder: \{10,8,4\} units for encoder, $\lambda_1=1$, $\lambda_2= 10^{c/10-3}$, \# of epochs: $50+25c$\\
			Nonlinear & 3-layered fully connected autoencoder: \{8,4,2\} units for encoder, $\lambda_1=1$, $\lambda_2= 10^{c/10-3}$, \# of epochs: $50+25c$ \\
			Yale B & 3-layered convolutional autoencoder:
			$\begin{array}{ll}
				encoder-1 \ \& \ decoder-3: 5\times 5 kernel, 10 \ channels\\
				encoder-2 \ \& \ decoder-2: 3\times 3 kernel, 20 \ channels\\
				encoder-3 \ \& \ decoder-1: 3\times 3 kernel, 30 \ channels
			\end{array}$ , $\lambda_1=1$, $\lambda_2= 10^{c/10-3}$, \# of epochs: $50+25c$ \\
			Coil 20 & 1-layered convolutional autoencoder: $\begin{array}{ll}
				encoder-1 \ \& \ decoder-1: 3\times 3 kernel, 15 \ channels
			\end{array}$ , $\lambda_1=1$, $\lambda_2= 150$, \# of epochs: 30 \\
			MNIST & 3-layered convolutional autoencoder: $\begin{array}{ll}
				encoder-1 \ \& \ decoder-3: 5\times 5 kernel, 10 \ channels\\
				encoder-2 \ \& \ decoder-2: 3\times 3 kernel, 20 \ channels\\
				encoder-3 \ \& \ decoder-1: 3\times 3 kernel, 30 \ channels
			\end{array}$ , $\lambda_1=1$, $\lambda_2= 1$, \# of epochs: 1000\\
			\hline
			
		\end{tabularx}
	\end{center}
	
	\begin{center}
		\caption{Parameters of DASC.}
		\label{tabpar_dasc}  
		\small\addtolength{\tabcolsep}{-1pt}
		\begin{tabularx}{\textwidth}{c||X}
			\hline
			\multicolumn{1}{c||}{data set} & \multicolumn{1}{c}{Parameters} \\ \hline
			Linear & 2-layered fully connected autoencoder: \{8,4\} units for encoder, $\lambda_1=0.5$, $\lambda_2= 0.1$, $\lambda_3=1$, \# of pretrain epochs = 200, \# of epochs: 300\\
			Nonlinear & 2-layered fully connected autoencoder: \{2,2\} units for encoder, $\lambda_1=0.5$, $\lambda_2= 0.1$, $\lambda_3=1$, \# of pretrain epochs = 200, \# of epochs: 300\\
			Yale B & 3-layered convolutional autoencoder:
			$\begin{array}{ll}
				encoder-1 \ \& \ decoder-3: 5\times 5 kernel, 10 \ channels\\
				encoder-2 \ \& \ decoder-2: 3\times 3 kernel, 20 \ channels\\
				encoder-3 \ \& \ decoder-1: 3\times 3 kernel, 30 \ channels
			\end{array}$ , $\lambda_1=0.5$, $\lambda_2= 0.1$, $\lambda_3=1$, \# of pretrain epochs = 1000, \# of epochs: 1500\\
			Coil 20 & 1-layered convolutional autoencoder:
			$\begin{array}{ll}
				encoder-1 \ \& \ decoder-1: 3\times 3 kernel, 15 \ channels
			\end{array}$  , $\lambda_1=0.5$, $\lambda_2= 0.1$, $\lambda_3=1$, \# of pretrain epochs = 1000, \# of epochs: 1200\\
			MNIST & 3-layered convolutional autoencoder:
			$\begin{array}{ll}
				encoder-1 \ \& \ decoder-3: 5\times 5 kernel, 10 \ channels\\
				encoder-2 \ \& \ decoder-2: 3\times 3 kernel, 20 \ channels\\
				encoder-3 \ \& \ decoder-1: 3\times 3 kernel, 30 \ channels
			\end{array}$  , $\lambda_1=0.5$, $\lambda_2= 0.1$, $\lambda_3=1$, \# of pretrain epochs = 1000, \# of epochs: 2000\\
			\hline
		\end{tabularx}
	\end{center}
\end{sidewaystable}

\end{document}